%% file: main.tex
\documentclass[12pt]{report} 
\usepackage{subfiles}

\usepackage[english]{babel}
\usepackage{comment}
\usepackage{csquotes}
\PassOptionsToPackage{table}{xcolor}
\usepackage{enumitem}
\usepackage{manyfoot}
\usepackage{amsmath,amsfonts,stmaryrd,amssymb,bbm, mathtools, amsthm}

\usepackage[backend=biber, style=ieee, natbib=true]{biblatex} 
\DeclareNewFootnote{H}[Alph]

\usepackage{amsmath}
\usepackage{graphicx}

\usepackage{nicematrix}
\usepackage{caption}
\usepackage{subcaption}
\usepackage{multirow}
\usepackage{float}
\usepackage[nottoc,notlot,notlof]{tocbibind}
\usepackage[super]{nth}
\usetikzlibrary{positioning, shapes, decorations.pathreplacing, decorations.pathmorphing, calligraphy, shapes.symbols, calc}
\usepackage{array, booktabs, makecell}
\usepackage{tabularray}
\usepackage{mathtools}
\usepackage[acronym]{glossaries}
\input{accronyms.tex}

\mathtoolsset{showonlyrefs}

\newcounter{overseteqcounter}

\title{A Quantifier-Reversal Approximation Paradigm for Recurrent Neural Networks}
\author{Clemens Hutter\thanks{Swiss National Bank, Börsenstrasse 15, 8001 Zürich, Switzerland}\hspace{4pt}\thanks{ETH Zürich, Chair for Mathematical Information Science,
Sternwartstrasse 7, 8092 Zürich, Switzerland} \and Valentin Abadie\footnotemark[2] \and Helmut Bölcskei\footnotemark[2] }
\begin{document}
\maketitle
\begin{abstract}
Classical neural network approximation results take the form: for every function $f$
and every error tolerance $\epsilon>0$, one constructs a neural network whose
architecture and weights depend on~$\epsilon$.  This paper introduces a fundamentally
different approximation paradigm that reverses this quantifier order.  For each target
function~$f$, we construct a \emph{single} recurrent neural network (RNN) with fixed
topology and fixed weights that approximates~$f$ to within any prescribed
tolerance $\epsilon>0$ when run for sufficiently many time steps.

The key mechanism enabling this quantifier reversal is temporal computation combined
with weight sharing: rather than increasing network depth, the approximation error
is reduced solely by running the RNN longer.  This yields exponentially decaying
approximation error as a function of runtime while requiring storage of only a small,
fixed set of weights.  Such architectures are appealing for hardware implementations
where memory is scarce and runtime is comparatively inexpensive.

To initiate the systematic development of this novel approximation paradigm, we focus
on univariate polynomials.  Our RNN constructions emulate the structural calculus
underlying deep feed-forward ReLU network approximation theory—parallelization, linear
combinations, affine transformations, and, most importantly, a clocked mechanism that
realizes function composition within a single recurrent architecture.  
The resulting RNNs have size independent of the error tolerance~$\epsilon$ and
hidden-state dimension linear in the degree of the polynomial.

\end{abstract}



{

\subfile{src/polyrnn/rnn_approx_polynom.tex}

}

\printbibliography

\end{document}

%% file: accronyms.tex
\glsdisablehyper
\newacronym[plural=RNNs, longplural=recurrent neural networks]{rnn}{RNN}{recurrent neural network}
\newacronym{lfm}{LFM}{Lipschitz fading-memory}
\newacronym{relu}{ReLU}{rectified linear unit}
\newacronym{elfm}{ELFM}{exponentially Lipschitz fading-memory}
\newacronym{plfm}{PLFM}{polynomially Lipschitz fading-memory}

\newacronym{pou}{p.o.u.}{\emph{partition of unity}}

%% file: src/polyrnn/rnn_approx_polynom.tex
\input{polyrnn_notation.tex}
\section{Introduction}

The starting point of this work is the classical universal approximation theorem for
feedforward neural networks \cite{cybenko89,funahashi89,hornik89}, which asserts that every continuous function on a compact domain can be approximated 
arbitrarily well by a shallow neural network with sigmoidal activation function.
Subsequent quantitative results relating the smoothness of the target function and the prescribed approximation error to the size of 
the approximating network were later obtained in \cite{Barron_1993,Barron_1994}. 

In the past two decades the rectified linear unit (ReLU) has become the dominant activation
function in theory and practice. Beginning with \cite{yarotskyErrorBoundsApproximations2017},
quantitative approximation results for deep ReLU networks have been developed
\cite{telgarskyNeuralNetworksRational2017,
schmidthieberKolmogorovArnoldRepresentation2021, siegel2023optimal},
culminating in \cite{deepAT2019}, which shows that deep ReLU networks approximate a wide
range of function classes in metric-entropy–optimal manner.


The quantitative approximation results in the literature typically take the following form: for a given function $f$ and a given 
approximation error $\epsilon > 0$, there exists a neural network $\nn$ that approximates $f$ to within error $\epsilon$,
formalized as
\begin{equation}\label{eq:paradigm1}
	\forall f: \forall \epsilon: \exists\, \nn \textrm{ such that $\nn$ approximates $f$ to within error $\epsilon$.}
\end{equation}
Thus the network architecture and weights depend on the chosen value of $\epsilon$.
If a smaller error is later required, a new (typically larger) network must be instantiated.
For instance, \cite[Proposition~III.5]{deepAT2019} shows that for every $\epsilon>0$,
there exists a deep ReLU network of $\epsilon$-independent width and depth $\O(\log{(\epsilon^{-1})})$ that approximates polynomials $f$ to within error $\epsilon$. 

In the present paper, we propose a new approximation paradigm that avoids this (unstructured) dependence of the network on the approximation error $\epsilon$. Specifically, we aim to reverse the order of quantifiers $\forall \epsilon, \exists\, \nn$ to get a statement of the form 
\begin{equation}\label{eq:paradigm2}
	\forall f: \exists\, \nn, \textrm{such that } \forall \epsilon: \textrm{ $\nn$ approximates $f$ to within error $\epsilon$.}
\end{equation}
Concretely, this will be achieved by repeatedly self-composing a single neural network of fixed
topology and fixed weights until the desired accuracy is reached. We show that the resulting
approximation error can be made arbitrarily small and, in fact, decays exponentially in the
number of compositions. The nature of this construction lends itself to a temporal formulation, specifically in terms of recurrent neural networks (RNNs).


\begin{definition}[RNN \cite{elman90,Goodfellow2016}]\label{def:elman_rnn}
	We denote by $\relu : \R \to \R, \relu(x) :=  max(0, x)$ the \gls{relu} function which acts component-wise, i.e., $\relu(x_1, \dots, x_m) := (\relu(x_1), \dots, \relu(x_m))$.
	An RNN with input dimension $\inputDim\in\N$, output dimension $\outputDim\in\N$, and hidden state size $\stateDim\in\N$ is parametrized by matrices $A_h \in \R^{\stateDim \times \stateDim}$, $A_x \in \R^{\stateDim \times \inputDim}$, $A_o \in \R^{\outputDim \times \stateDim}$, and vectors $b_h \in \R^{\stateDim}$, $b_o \in \R^{\outputDim}$.
	These quantities, collectively called the weights of the RNN,
	specify the {hidden state operator} $\hidOp: \seqSpace{\inputDim} \rightarrow \seqSpace{\stateDim}$ mapping an input sequence $(x[t])_{t \in \N_0}$ recursively to the sequence of hidden states $(h[t])_{t \in \N_0}$ according to
	\begin{align}
		h[-1] & := 0 \in \R^{\stateDim}               \\
		(\hidOp x)[t]= h[t]  & = \relu(A_h h[t-1] + A_x x[t] + b_h),
	\end{align}
	and the {output mapping} $\outputOp: \R^{\stateDim} \rightarrow \R^{\outputDim}$,
	\[
		\outputOp(h) :=  A_o h + b_o.
	\]
	The associated RNN is the operator $\rnnOp{}: \seqSpace{d} \rightarrow \seqSpace{d'}$ given by
	\[
		\rnnOp{} := \outputOp \hidOp,
	\]
	with 
	\[
		(\rnnOpp{} x)[t] =   (\outputOp \hidOp x)[t] = \outputOp\left((\hidOp x)[t]\right), \quad t\in\No.
	\]
	We use the notation $\rnnSizeFunctionIn{\rnnOp{}}=d$, $\rnnSizeFunctionOut{\rnnOp{}}=\rnnSizeFunctionOut{\outputOp}=d'$, $\rnnSizeFunctionHid{\rnnOp{}} = \rnnSizeFunctionHid{\hidOp}=m$ to describe the size of the RNN.
\end{definition}

%
%
To conform with \eqref{eq:paradigm2}, we wish, for a given function $f$, to find an RNN that achieves any (arbitrarily small) approximation error $\epsilon$ provided we let it run sufficiently long. To this end, if we desire an approximation of the function $f$ at the point $x \in \R$, we take the input sequence of the RNN as $\xtilde[t]=x \ind{t=0}, t \in \N_0$. Here, $\ind{\cdot}$ denotes the truth function which takes on the value $1$ if the statement inside $\{\cdot\}$ is true and equals $0$ otherwise. 
%
To formalize this, we introduce the following operator. 

\begin{definition}\label{def:spreadInput}
	The mapping $\spreadInputOp: \R^d \rightarrow \seqSpace{d}$ is defined according to
	\[
		(\spreadInputOp x)[t] = x \ind{t=0}, \qquad \delforall t \in \N_0.
	\]
\end{definition}

The corresponding output sequence of the RNN then produces increasingly accurate approximations of $f(x)$ as time $t$ evolves. 
We can operationalize 
the approximation paradigm \eqref{eq:paradigm2} in terms of RNNs as
\begin{align}
	\begin{split}
	\label{eq:paradigm3}
	\forall f: \exists \, \rnnIntro: \forall \epsilon: \exists t_0: 
	\sup_{t\geq t_0} \sup_{x}  |(\rnnIntro \spreadInputOp x)[t] - f(x)| < \epsilon.
	\end{split} 
\end{align}
%
%
Every approximation theorem fitting this paradigm must have the size, topology, and weights of the approximating RNN $\rnnIntro$ be independent of the approximation error $\epsilon$, simply by virtue of the quantifier order in \eqref{eq:paradigm3}. 
Only the runtime required to achieve the desired approximation error $\epsilon$ will depend on $\epsilon$. This approximation paradigm exhibits interesting practical properties as storing the fixed (in general small) RNN on digital devices requires little memory and the approximation error can be controlled simply by adjusting the runtime of the RNN.

In the presentation of the proposed new approximation paradigm, we have deliberately kept \eqref{eq:paradigm1}-\eqref{eq:paradigm3} slightly vague in order  to convey the central idea of reversing the quantifier order unencumbered by burdensome technicalities. We believe that neural network approximation results following the paradigm \eqref{eq:paradigm3} are possible for a variety of function classes. 
The present paper aims at initiating such a neural network approximation theory program by considering the approximation of univariate polynomials. 
Note that, while the network is independent of the desired approximation error $\varepsilon$, it does depend on the function $f$ to be approximated. In particular, as Theorem~\ref{thm:main} below shows, the size of the resulting RNN depends on the degree of the polynomial.
The techniques and specific neural network constructions we develop could prove useful more generally. 
For example, we think that our results can be extended to multivariate polynomials with relative ease. 
%
%
%

We next state the central result of this paper, which says that the class of univariate polynomials can be approximated by RNNs according to the paradigm \eqref{eq:paradigm3}. Moreover, and perhaps surprisingly, the approximation error decreases exponentially in the runtime of the RNN.

\begin{theorem}\label{thm:main}
	Let $N\in\N$, $a_0, \dots, a_N \in \R$, and $D \geq 1$. There exists an RNN $\rnnOpp_a$ such that for $t \geq 16 \log(N)$, 
	\begin{equation}
		\sup_{x\in [-D, D]}\left|(\rnnOpp_a \spreadInputOp x)[t] - \sum_{i=0}^{N} a_i x^i\right| \leq
		\norm{a}_1 \constMultiplicative 4^{- \constInPower t},
	\end{equation}
	with $\constMultiplicative = 16N D^{2N}$ and $\constInPower = \frac{1}{4\ceil{\log(N)}}$.
\end{theorem}

A more detailed version of this result is presented in Theorem \ref{tm:main_theorem_with_proof}. The remainder of the paper builds up the technical material needed in the proof of Theorem \ref{tm:main_theorem_with_proof}. We close this section with some notation conventions.
\paragraph{Notation.} $\N$ and $\No$ denote the set of natural numbers excluding and, respectively, including zero.
$\seqSpace{d}$ stands for the space of sequences of vectors in $\R^d$ indexed by time in $\No$, that is for $x[\cdot]\in\seqSpace{d}$, we have $x[t] \in \R^d$, for $t\in\No$. For $x \in \R^d$, we let $\inorm{x} = \max_{i=1, \dots, d} |x_i|$ and write $x \vecleq c$, for some $c\posConst{}$, to indicate that $x_i \leq c$, for all $i\in\{1, \dots , d\}$. $\Imat{d}\in \R^{d\times d}$ is the $d$-dimensional identity matrix and $\Ivec{d}\in\R^d$ stands for the $d$-dimensional vector with all entries equal to one. $\log$ denotes the logarithm to base 2. Constants are throughout understood to be in $\R^+$. 
For $a,b \in \No$, set $\{a,\ldots,b\} := \No \cap [a,b]$. In particular, $\{a,\ldots,b\} = \varnothing$ whenever $b < a$.


\subsection{Relationship to deep neural network approximation theory}

As an aside, it is instructive to compare RNNs with deep feed-forward neural networks. 
This subsection is not needed for the remainder of the paper, but provides additional context.
We begin by recalling the definition of a deep feed-forward neural network.

\begin{definition}[Deep Neural Network] \label{def:deepNN}   \cite[Definition II.1]{deepAT2019}            
  Let $L, \hidLayerDim_0, \hidLayerDim_1, \dots, \hidLayerDim_L \in \mathbb{N}$ with $L \geq 2$.  
  A mapping $\Phi: \mathbb{R}^{\hidLayerDim_0}\rightarrow \mathbb{R}^{\hidLayerDim_L}$ defined by
  \[
  \Phi = 
  \begin{cases}
    W_2 \circ \rho \circ W_1, & L=2, \\
    W_L \circ \rho \circ W_{L-1} \circ \rho \circ  \cdots \circ  \rho \circ W_1, & L \geq 3,
  \end{cases}
  \]
  with affine maps $W_\ell: \mathbb{R}^{\hidLayerDim_{\ell-1}} \rightarrow \mathbb{R}^{\hidLayerDim_\ell}$, 
  $\ell\in \{ 1,\dots,L\}$, and activation function $\rho$, is called a feed-forward neural network.
  The $W_\ell$ are given by $W_\ell(x) = A_\ell x + b_\ell$, with 
  $A_\ell \in \R^{\hidLayerDim_\ell \times \hidLayerDim_{\ell-1}}$ and $b_\ell \in \R^{\hidLayerDim_\ell}$.
\end{definition}

Now let $f$ be a fixed polynomial and let $\epsilon>0$ be the desired approximation error. 
By Theorem~\ref{thm:main}, there exists an RNN $\rnnOpp_f$ such that
\begin{equation}
  \sup_{x\in [-D, D]}
  \big|(\rnnOpp_f \spreadInputOp x)[T] - f(x)\big|
  \leq \epsilon,
\end{equation}
for $T := \ceil{c \log(\epsilon^{-1})}$ with a constant $c>0$ depending only on~$f$.

For this fixed $T$, the map $x \mapsto (\rnnOpp_f \spreadInputOp x)[T]$ can be expressed as a 
deep feed-forward network with $T+2$ layers. We now make this equivalence explicit. 
Recall that the RNN input is given by
$(\spreadInputOp x)[0]=x$ and $(\spreadInputOp x)[t]=0$ for $t\ge1$, 
and that the hidden state is initialized as $h[-1]=0$.
The RNN updates therefore take the form
\[
h[0] = \rho(\Ax{} x + \bh{}), 
\qquad 
h[t] = \rho(\Ah{} h[t-1] + \bh{}) \ (t\ge1),
\qquad
(\rnnOpp_f \spreadInputOp x)[T] = \Ao{} h[T] + \bo{}.
\]
Define the affine maps
\[
W_1(x) := \Ax{} x + \bh{}, \qquad
W_{\mathrm{share}}(h) := \Ah{} h + \bh{}, \qquad
W_{T+2}(h) := \Ao{} h + \bo{},
\]
and set $W_\ell := W_{\mathrm{share}},$ for $\ell = 2,\dots,T+1$.
Iterating the recurrence yields
\[
h[T] 
  = \rho \circ W_{T+1} 
    \circ \cdots \circ \rho \circ W_1(x),
\]
and hence
\[
(\rnnOpp_f \spreadInputOp x)[T]
  = W_{T+2}\!\left(
      \rho \circ W_{T+1} \circ \rho \circ \cdots \circ \rho \circ W_1(x)
    \right).
\]
Thus unfolding $\rnnOpp_f$ over $T$ time steps yields the deep network
\begin{equation}\label{eq:unfolded_rnn}
  \Phi 
  = W_{T+2} \circ \rho \circ W_{T+1} \circ \rho \circ \cdots \circ \rho \circ W_1,
\end{equation}
with $T+2$ layers and shared weights in layers $2$ through $T+1$.

We may now compare this with \cite[Proposition~III.5]{deepAT2019}, which shows that 
for every $\epsilon>0$ there exists a deep ReLU network $\Phi'_f$ of 
$\epsilon$-independent (constant) width and $\mathcal{O}(\log(\epsilon^{-1}))$ layers 
that approximates $f$ to within error $\epsilon$.  
Since unfolding $\rnnOpp_f$ for $T=\mathcal{O}(\log(\epsilon^{-1}))$ time steps likewise 
produces a network of constant width (as shown later in the paper) and depth $\mathcal{O}(\log(\epsilon^{-1}))$,  
we obtain the same connectivity--error trade-off directly from Theorem~\ref{thm:main}.  
The key difference is that the construction in \cite{deepAT2019} requires redesigning the entire 
network $\Phi'_f$ for each new value of $\epsilon$, whereas our RNN-based method uses a single 
fixed architecture: to achieve a smaller error, one simply runs the network longer, repeatedly 
applying the shared layer $W_{\mathrm{share}}$.

\paragraph{Conceptual remark.}
Classical deep neural network approximation theory, particularly the structural 
calculus developed in \cite{deepAT2019}, relies on a 
small set of fundamental operations: composition of networks, parallelization, linear 
combinations, pre-and post-composition with affine maps, and depth extension via identity mappings.

A central theme of the present paper is that each of these operations admits an RNN analogue, 
while strictly adhering to the weight sharing paradigm described above.  
As demonstrated by the unfolding argument, RNNs emulate deep feed-forward networks
by reusing the same weights over time.

Section~3 develops this novel RNN calculus explicitly: parallelization (Lemma~\ref{lm:parallel_net}), 
affine pre- and post-composition (Lemmata~\ref{lm:linear_map_rnn} and \ref{lm:linear_map_rnn_output}), 
clocked concatenation for function composition 
(Theorem~\ref{thm:rnn4func_concat}), and the structured assembly of multiple concatenations 
(Lemma~\ref{lem:multiconcat_rnn_tree} and Corollary~\ref{cor:multiconcat_rnn}).  
These constructions show that the expressive power of deep feed-forward ReLU networks can equivalently be 
attained temporally within a single recurrent architecture through weight sharing, with 
clocked concatenation providing the principal mechanism for implementing compositional 
depth.

\section{Approximating the squaring function and multiplication}

The constructions developed in this section correspond to the basic operations used
in classical deep feed-forward ReLU network approximation theory, in particular squaring and
multiplication.  In the classical setting, these operations are assembled through a
structural calculus—compositions, parallelization, and affine transformations—to
produce approximations of more complex functions such as polynomials.  Our RNN-based
approach reproduces this calculus, but within a recurrent architecture. We design
RNNs that approximate squaring and multiplication and then combine them into polynomial
approximations using the mechanisms developed in Section~\ref{sec:concatenate-RNNs}, most notably the
clocked concatenation and weight sharing ideas inherent to the RNN framework.

Before delving into the details of our RNN constructions, we point out a simplification of the RNN definition. 
\begin{remark}\label{rm:operator_rnn_description}
	As we shall only be concerned with inputs of the form $(\spreadInputOp x)[t] = x \ind{t=0}$, for $x\in\R^d$, Definition \ref{def:elman_rnn} simplifies to
	\begin{align}
		(\hidOp \spreadInputOp x)[0] & = \relu( A_x x + b_h)                                            &               &                   \\
		(\hidOp \spreadInputOp x)[t] & = \relu( A_h \left( (\hidOp \spreadInputOp x)[t-1]\right) + b_h), & t\in \N,
	\end{align}
	and
	\begin{equation}
		(\rnnOp{} \spreadInputOp x)[t] = (\outputOp \hidOp \spreadInputOp x)[t] = \outputOp\left((\hidOp \spreadInputOp x)[t]\right), \qquad \qquad \qquad \qquad t\in\No.
	\end{equation}
    As a notational convention, we extend the sequences $((K\mathcal D x)[t])_{t\in\mathbb N_0}$ and 
$((R\mathcal D x)[t])_{t\in\mathbb N_0}$ to negative time indices by setting
\[
  (K\mathcal D x)[t] = 0 \quad\text{and}\quad (R\mathcal D x)[t] = 0, \qquad \text{for all } t<0.
\]
This convention is used for expositional convenience only and does not affect the RNN dynamics on $t\ge0$.
\end{remark}
We start by devising an RNN that approximates the squaring function. To this end we make use of the following feedforward neural network construction from \cite{deepAT2019, yarotskyErrorBoundsApproximations2017}.

\begin{lemma}\label{lm:import_interpolation_dennis}
	Let $F(x)\coloneqq x-x^2$, $x\in[0,1]$. Further, with $m\in\N$, let $I_m: [0,1] \to [0,1]$ be the piecewise linear interpolant of $F$ at the points $\frac{k}{2^m}$, with $k\in\{0, \dots, 2^m\}$, that is, 
	\[
		I_m\!\left(\frac{k}{2^m}\right) = F\!\left(\frac{k}{2^m}\right), \quad \textrm{ for } k\in\{0, \dots, 2^m\},
	\]
	and $I_m$ is affine on the intervals $\left[\frac{k}{2^m}, \frac{k+1}{2^m}\right]$, $k\in \{0, \dots, 2^m-1\}$. It holds that 
\begin{equation}
		\sup_{x\in[0,1]} |F(x) - I_m(x)| = 2^{-2m-2}.
	\end{equation}
	Furthermore, define 
\begin{equation}
		s_\ell(\cdot) \coloneqq 2^{-1}\relu(\cdot) - \relu(\cdot - 2^{-2\ell-1}), \quad \ell\in\No,
	\end{equation}
	and recursively $H_\ell = s_\ell \circ H_{\ell-1}$, for $\ell\in\N$, with $H_0 = s_0$. Then, it holds that 
	\begin{equation}
		I_m(x) = \sum_{\ell=0}^{m-1} H_\ell(x), \qquad \textrm{ for } x \in [0, 1].
	\end{equation}
\end{lemma}
\begin{proof} See the proof of Proposition III.2 in \cite{deepAT2019}.
\end{proof}

We furthermore need the following property of $I_m$, which is not stated in \cite{deepAT2019}.
\begin{corollary}\label{cor:dennis_properties}
	For every $m\in \N$ and all $x \in [0,1]$,
	\begin{equation}
		0\leq I_m(x) \leq x, 
	\end{equation}
    where $I_m$ is defined in Lemma \ref{lm:import_interpolation_dennis}.
\end{corollary}
\begin{proof}
	First, note that $F(x)=x-x^2 \in[0,1/4]$, for all $x\in[0,1]$. Thus, $I_m(x)\in[0, 1/4]$, for all $x\in[0,1]$, and in particular $I_m(x)\geq0$. Furthermore,
	since  $F(x) = x-x^2$ is concave, we have $ I_m(x) \leq F(x)$, for all $x\in[0,1]$. This implies 
	\[
		I_m(x) \leq F(x) = x-x^2 \leq x, \quad \textrm{ for all } x \in [0,1],
	\]
	thereby completing the proof.
\end{proof}
We are now ready to construct an RNN that approximates the squaring function.

\begin{theorem}\label{thm:square_net}
	For $D\geq1$, there is an RNN $\rnnSquare = \outSquare \hidSquare $, with $\rnnSizeFunctionIn{\rnnSquare} = \rnnSizeFunctionOut{\rnnSquare}=1$, $\rnnSizeFunctionHid{\rnnSquare}= 7$, such that, for all $x\in[-D, D]$ and all $t\in\N_0$,
	\[
		|(\rnnSquare{}\spreadInputOp x)[t] - x^2| \leq \frac{D^2}{4} \phantom{\cdot} 4^{-t},
	\]
	as well as
	\begin{align}
		\inorm{(\hidSquare \spreadInputOp{} x)[t]} \leq 1\; \textrm{ and }\; 0 \leq (\rnnSquare{} \spreadInputOp{} x)[t] \leq D^2.
	\end{align}
\end{theorem}
\begin{proof}
	\newcommand{\polA}{z}
	\newcommand{\polB}{\textcolor{red}{z_2}}

	The weights of $\rnnSquare$ are as follows:

	\[
		A_h = 
		\NiceMatrixOptions
		{nullify-dots,code-for-first-col = \color{blue},code-for-first-row=\color{blue}, code-for-last-col=\color{blue} }
		\begin{pNiceArray}{cc|ccc|cc}
			0 & 0  & 0 & 0 & 0 &0 &0\\
			0 & 0  & 0 & 0 & 0 &0 &0\\
			\hline
			1&1  &2^{-1}&-1 &0   &0 &0\\
			1&1  &2^{-1}&-1 &0  &-1 &0\\
			1&1  &-2^{-1}&1 &1 &0 &0\\
			\hline
			0 & 0 &0&0 & 0 &2^{-2} &-2^{-1}\\
			0 & 0 &0&0 & 0  &0 &0\\
		\end{pNiceArray},
		\qquad
		b_h = \begin{pNiceArray}{c}
			0  \\
			0  \\
			\hline
			0\\
			0\\
			0\\
			\hline
			2^{-1}\\
			1\\
		\end{pNiceArray},
	\]
	\[
		A_x = \frac{1}{D}\begin{pNiceArray}{c}
			1  \\
			-1 \\
			\hline
			0\\
			0\\
			0\\
			\hline
			0\\
			0\\
		\end{pNiceArray},
		\qquad
		A_o = D^2
		\begin{pNiceArray}{cc|ccc|cc}
			0 & 0& -2^{-1}  & 1 & 1  & 0 & 0
		\end{pNiceArray},
		\qquad b_o=0.
	\]
	Next, we arbitrarily fix $x \in [-D, D]$, let $\polA \coloneqq \frac{|x|}{D}$, and compute the sequence $h[\cdot] \coloneqq (\hidOp{}\spreadInputOp x)[\cdot]$ of hidden states corresponding to the input sequence $\inpSeq \coloneqq (\spreadInputOp x)[\cdot] = x\ind{\cdot=0}$ according to Definition \ref{def:elman_rnn}. 
    We start by proving through induction that
	\begin{equation}\label{eq:multiplication_rnn_base_case}
		h[t] =
		\begin{pNiceArray}{c}
			0 \\ 0 \\
			\hline
			\relu(H_{t-2}(\polA)) \\
			\relu(H_{t-2}(\polA) - 2^{-2t+1}) \\
			\polA - \sum_{i=0}^{t-2}H_i(\polA) \\
			\hline
			2^{-2t-1} \\
			1
		\end{pNiceArray}, \qquad \text{ for all } t \geq 2,
	\end{equation}
	with $H_t$, $t\in\No$, as defined in Lemma \ref{lm:import_interpolation_dennis}.
	Starting from $h[-1]=0$, we compute
	\begin{align}
		h[0] & = \relu(A_h h[-1] + A_x \inpSeq[0] + b_h) \\
		     & = \relu(A_x x + b_h)                      \\
		     & =\begin{pNiceArray}{c}
			        \relu(\frac{x}{D})\\
			        \relu(-\frac{x}{D})\\
			        \hline
			        0\\0\\0\\
			        \hline
			        2^{-1}\\
			        1
		        \end{pNiceArray}.
	\end{align}
	Using
	$
		\polA = \relu\left(\frac{x }{D}\right) + \relu\left(-\frac{x}{D}\right) \geq 0,
	$
	we get
	\begin{align}
		h[1] & = \relu(A_h h[0] + A_x \inpSeq[1] + b_h) \\
		     & = \relu(A_h h[0] + b_h)                  \\
		     & =
		\begin{pNiceArray}{c}
			0 \\ 0 \\
			\hline
			\polA \\
			\relu(\polA - 2^{-1})\\
			\polA \\
			\hline
			2^{-3}\\
			1
		\end{pNiceArray}.
	\end{align}
	Further,
	\begin{align}
		h[2]
		 & = \relu(A_h h[1] + b_h) \\
		 & = \relu
		\begin{pNiceArray}{c}
			0 \\ 0 \\
			\hline
			2^{-1} \relu(\polA) - \relu(\polA - 2^{-1}) \\
			2^{-1} \relu(\polA) - \relu(\polA - 2^{-1}) -2^{-3}\\
			\polA - (2^{-1} \polA - \relu(\polA - 2^{-1}))\\
			\hline
			2^{-2} \cdot 2^{-3} - 2^{-1} + 2^{-1}\\
			1
		\end{pNiceArray}
		=
		\begin{pNiceArray}{c}
			0 \\ 0\\
			\hline
			\relu(s_0(\polA)) \\
			\relu(s_0(\polA) -2^{-3})\\
			\polA - s_0(\polA)\\
			\hline
			2^{-2 \cdot 2-1}\\
			1
		\end{pNiceArray},
	\end{align}
	where $s_t(\cdot), t\in\No$, was defined in Lemma \ref{lm:import_interpolation_dennis}. Hence, recalling that $H_0=s_0$ as per Lemma \ref{lm:import_interpolation_dennis}, we established the base case $t=2$ of the induction.

	We proceed to the induction step. To this end, we assume that \eqref{eq:multiplication_rnn_base_case} holds for some $t\geq2$. 
	Now, note that
	\begin{align}
		h[t+1]
		 & = \relu(A_h h[t] + b_h)                                                                                        \\
		 & = \relu
		\begin{pNiceArray}{c}
			0 \\ 0 \\
			\hline
			2^{-1} \relu(H_{t-2}(\polA))- \relu(H_{t-2}(\polA )-2^{-2t+1})\\
			2^{-1} \relu(H_{t-2}(\polA))- \relu(H_{t-2}(\polA )-2^{-2t+1}) - 2^{-2t-1}\\
			\polA - \sum_{i=0}^{t-2}H_i(\polA) -  \left(2^{-1} \relu(H_{t-2}(\polA))- \relu(H_{t-2}(\polA )-2^{-2t+1}) \right)\\
			\hline
			2^{-2}\cdot 2^{-2t-1} -2^{-1} + 2^{-1}\\
			1
		\end{pNiceArray} \\
		 & = \relu
		\begin{pNiceArray}{c}
			0 \\ 0\\
			\hline
			s_{t-1}(H_{t-2}(\polA))\\
			s_{t-1}(H_{t-2}(\polA)) - 2^{-2t-1}\\
			\polA - \sum_{i=0}^{t-2}H_i(\polA) - s_{t-1}(H_{t-2}(\polA))\\
			\hline
			2^{-2(t+1)-1}\\
			1
		\end{pNiceArray}
		=
		\begin{pNiceArray}{c}
			0 \\ 0\\
			\hline
			\relu(H_{t-1}(\polA))\\
			\relu(H_{t-1}(\polA) - 2^{-2(t+1)+1})\\
			\polA - \sum_{i=0}^{t-1}H_i(\polA)\\
			\hline
			2^{-2(t+1)-1}\\
			1
		\end{pNiceArray},
	\end{align}
	where we used $s_{t-1}(\cdot) = 2^{-1}\relu(\cdot) - \relu(\cdot - 2^{-2t+1})$ and $H_t = s_t \circ H_{t-1}$, both as per Lemma \ref{lm:import_interpolation_dennis}. This completes the proof of \eqref{eq:multiplication_rnn_base_case}.
	Furthermore, we see directly that $\inorm{h[0]} \leq 1$ and $\inorm{h[1]} \leq 1$. By \eqref{eq:multiplication_rnn_base_case}, we obtain $\inorm{h[t]} \leq 1$, for all $t\geq 2$, upon using Corollary \ref{cor:dennis_properties}.

	Next, we compute the RNN output $(\rnnSquare{}x)[t]$, for $t \geq 2$, as follows
	\begin{align}
		(\rnnSquare{}x)[t] & = A_o h[t] + b_o                                                                                                                       \\
		                   & = D^2 \left(\polA - \sum_{i=0}^{t-2}H_i(\polA) -  \left(2^{-1} \relu(H_{t-2}(\polA))- \relu(H_{t-2}(\polA )-2^{-2t+1}) \right) \right) \\
		                   & = D^2 \left(\polA - \sum_{i=0}^{t-2}H_i(\polA) -  H_{t-1}(\polA) \right)                                                               \\
		                   & = D^2 \left(\polA - \sum_{i=0}^{t-1}H_i(\polA) ) \right)                                                                               \\
		                   & = D^2 \left(\polA - I_t(\polA) \right).
	\end{align}
	As $\polA \in [0, 1]$, we have, by Corollary \ref{cor:dennis_properties}, that
	$\polA - I_t(\polA)\in [0, 1]$. Hence, $ 0 \leq (\rnnSquare{}x)[t] \leq D^2$, for all $t \geq 2$.
	Furthermore,
	\begin{align}
		\left| x^2 -  (\rnnSquare{}\spreadInputOp x)[t] \right| & =
		D^2\left|\polA^2 -   \left(\polA - I_t(\polA) \right)\right|                                                                      \\
		                                                        & \leq  D^2 \sup_{y\in[0,1]}\left|y^2 -   \left(y - I_t(y) \right)\right| \\
		                                                        & \leq D^2 \phantom{\cdot} 2^{-2t-2} = \frac{D^2}{4}4^{-t},
	\end{align}
	where the second inequality follows from Lemma \ref{lm:import_interpolation_dennis}. The proof is finalized upon noting that $x\in[-D, D]$ was arbitrary.
\end{proof}
A result that is, at first glance, similar to Theorem~\ref{thm:square_net} was established in \cite[Theorem 4.4]{bohnRecurrentNeuralNetworks2021}, albeit using an RNN architecture that does not lead to quantifier reversal.

We continue our RNN construction program by approximating the multiplication operation. This will be done through the polarization identity
\begin{equation}\label{eq:polarisation_identity}
	x_1 \cdot x_2 = \left( \frac{x_1 + x_2}{2}\right)^2 - \left(\frac{x_1 - x_2}{2}\right)^2.
\end{equation}
Specifically, we will first map the input $(x_1, x_2)$ to $((x_1 + x_2)/2, (x_1 - x_2)/2)$ through an affine transformation and then
apply two instances of $\rnnSquare$ in parallel followed by a linear combination of their outputs. We proceed to develop auxiliary results needed in 
the construction of the RNN approximating the multiplication operation. The first lemma provides a formal way to run several RNNs in parallel inside one larger RNN.

\begin{lemma}\label{lm:parallel_net}
	For $N\in\N$, let $\rnnOpp^1 = \outputOp^1 \hidOp^1,  \dots, \rnnOpp^N = \outputOp^N \hidOp^N $ be RNNs. There exists an RNN $\rnnOpp = \outputOp \hidOp$ such that for all $x_1 \in \R^{\rnnSizeFunctionIn{\rnnOpp^1}}, \dots,  x_N \in \R^{\rnnSizeFunctionIn{\rnnOpp^N}}$, and all $t \in \No$,
	\begin{equation}\label{eq:parallel}
		\left(\hidOp \spreadInputOp{}\!\! \begin{pmatrix}x_1 \\ \vdots \\  x_N \end{pmatrix}\right)[t]  =    \begin{pmatrix}
			(\hidOp^1 \spreadInputOp x_1)[t] \\
			\vdots                           \\
			(\hidOp^N \spreadInputOp x_N)[t]
		\end{pmatrix}, 
	\end{equation}
	and
	\begin{equation}\label{eq:parallel_output}
		\left(\rnnOpp \spreadInputOp{} \!\! \begin{pmatrix}x_1 \\ \vdots \\  x_N \end{pmatrix}\right)[t]
		= \begin{pmatrix}
			(\rnnOpp^1 \spreadInputOp x_1)[t] \\
			\vdots                            \\
			(\rnnOpp^N \spreadInputOp x_N)[t]
		\end{pmatrix}. 
	\end{equation}
\end{lemma}
\begin{proof}
	The weights of $\rnnOpp = \outputOp \hidOp$ are given by
	\begin{align}\label{eq:parallel_rnn_weights}
		A_h & = \begin{pmatrix}
			        \Ah{1} & \hdots & 0      \\
			        \vdots & \ddots & \vdots \\
			        0      & \hdots & \Ah{N} \\
		        \end{pmatrix},
		\qquad
		A_x =
		\begin{pmatrix}
			\Ax{1} & \hdots & 0      \\
			\vdots & \ddots & \vdots \\
			0      & \hdots & \Ax{N} \\
		\end{pmatrix},
		\quad
		b_h =
		\begin{pmatrix} \bh{1} \\ \vdots \\ \bh{N} \end{pmatrix}, \\
		A_o & =
		\begin{pmatrix}
			\Ao{1} & \hdots & 0      \\
			\vdots & \ddots & \vdots \\
			0      & \hdots & \Ao{N} \\
		\end{pmatrix},
		\qquad b_o = \begin{pmatrix} \bo{1} \\ \vdots \\ \bo{N} \end{pmatrix}.
	\end{align}
	Now, arbitrarily fix $x = \begin{pmatrix} x_1 \\ \vdots \\ x_N \end{pmatrix}$. We first show \eqref{eq:parallel}, by induction.
	Indeed, for $t=0$, we have by Remark \ref{rm:operator_rnn_description},
	\[
		(\hidOp \spreadInputOp x)[0] = \relu(\Ax{} x + \bh{}) =
		\relu \begin{pmatrix}
			\Ax{1} x_1 + \bh{1} \\
			\vdots              \\
			\Ax{N} x_N + \bh{N}
		\end{pmatrix}=
		\begin{pmatrix}
			(\hidOp^1 \spreadInputOp x_1)[0] \\
			\vdots                         \\
			(\hidOp^N \spreadInputOp x_N)[0]
		\end{pmatrix}.
	\]
	Next, assume that \eqref{eq:parallel} holds for some $t\in\No$ and compute
	\begin{align}
		(\hidOp \spreadInputOp x)[t+1] & =
		\relu\left( A_h \left( (\hidOp \spreadInputOp x)[t]\right) + b_h\right)                             \\
		                               & = \relu\left( A_h \begin{pmatrix}
				                                                   (\hidOp^1 \spreadInputOp x_1)[t] \\
				                                                   \vdots                         \\
				                                                   (\hidOp^N \spreadInputOp x_N)[t]
			                                                   \end{pmatrix} + b_h\right)                   \\
		                               & = \relu \begin{pmatrix}
			                                         \Ah{1}\left((\hidOp^1 \spreadInputOp x_1)[t]\right) + \bh{1} \\
			                                         \vdots                                                     \\
			                                         \Ah{N}\left((\hidOp^N \spreadInputOp x_N)[t]\right) + \bh{N}
		                                         \end{pmatrix}  = \begin{pmatrix}
			                                                          (\hidOp^1 \spreadInputOp x_1)[t+1] \\
			                                                          \vdots                         \\
			                                                          (\hidOp^N \spreadInputOp x_N)[t+1]
		                                                          \end{pmatrix}.
	\end{align}
	This completes the proof of \eqref{eq:parallel}. To establish \eqref{eq:parallel_output}, we note that
	\begin{align}
		(\rnnOpp \spreadInputOp x)[t]
		 & = \Ao{} (\hidOp \spreadInputOp x)[t] + \bo{}
		= \Ao{} \begin{pmatrix}
			        (\hidOp^1 \spreadInputOp x_1)[t] \\
			        \vdots                         \\
			        (\hidOp^N \spreadInputOp x_N)[t]
		        \end{pmatrix} + \bo{}              \\
		 & =  \begin{pmatrix}
			      \Ao{1}(\hidOp^1 \spreadInputOp x_1)[t] + \bo{1} \\
			      \vdots                                        \\
			      \Ao{N}(\hidOp^N \spreadInputOp x_N)[t] + \bo{N}
		      \end{pmatrix}
		=
		\begin{pmatrix}
			(\rnnOpp^1 \spreadInputOp x_1)[t] \\
			\vdots                            \\
			(\rnnOpp^N \spreadInputOp x_N)[t]
		\end{pmatrix}, \qquad \delforall t \in \No. \qedhere
	\end{align}
\end{proof}
The next result shows how a fixed linear map can be absorbed into the first layer of an RNN.
\begin{lemma}\label{lm:linear_map_rnn}
	Let $\rnnOpp = \outputOp \hidOp$ be an RNN with $\rnnSizeFunctionIn{\rnnOpp} = d$ and let $A \in \R^{d \times d'}$, $d'\in\N$. Then, there exists an RNN $\rnnOpp' = \outputOp \hidOp'$ such that, for all $x\in\R^{d'}$, and all $t \in \No$,
	\begin{align}
		(\rnnOpp' \spreadInputOp x)[t] & =  \left(\rnnOpp \spreadInputOp (Ax)\right)[t] \qquad \textrm{and} &
		(\hidOp' \spreadInputOp x)[t]  & =  \left(\hidOp \spreadInputOp (Ax)\right)[t].
	\end{align}
\end{lemma}
\begin{proof}
	Let $\Ah{}, \Ax{}, \Ao{}, \bh{}, \bo{}$ be the weights of the RNN $\rnnOpp$.  We define the corresponding RNN $\rnnOpp'=\outputOp \hidOp'$ with weights $A_h'=\Ah{}, A_x' = \Ax{}A, A_o' = \Ao{},  b_h'=\bh{}, b_o' = \bo{}$, in particular the output mapping of $\rnnOpp'$ is identical to that of $\rnnOpp$. 
    First, we establish, through induction, that
	\begin{equation}\label{teq:inductAssume}
		(\hidOp' \spreadInputOp x)[t]  =  \left(\hidOp \spreadInputOp (Ax)\right)[t], \qquad \text{for all } t\in\No.
	\end{equation}
	The base case follows from 
	\begin{align}
		(\hidOp' \spreadInputOp x)[0] & = \relu\left( A_x' x + b_h'\right)                                        &  & \\
		                              & = \relu\left( \Ax{}A x + b_h\right)                                       &  & \\
		                              & = \relu\left( \Ax{}(A x) + b_h\right)  = (\hidOp \spreadInputOp (Ax))[0].
	\end{align}
	Next, assume that \eqref{teq:inductAssume} holds for some $t\in\No$. We have by Remark \ref{rm:operator_rnn_description},
	\begin{align}
		(\hidOp' \spreadInputOp x)[t+1] 
		&= \relu \left(A_h' \left((\hidOp' \spreadInputOp x)[t]\right) + b_h'\right)\\
		&= \relu \left(A_h \left((\hidOp' \spreadInputOp x)[t]\right) + b_h\right)\\
		&= \relu \left(A_h \left( \left(\hidOp \spreadInputOp (Ax)\right)[t]\right) + b_h\right)\\
		&= \left(\hidOp \spreadInputOp (Ax)\right)[t+1],
	\end{align}
	which concludes the induction argument. 
    Finally, $(\rnnOpp' \spreadInputOp x)[t] =  \left(\rnnOpp \spreadInputOp (Ax)\right)[t]$, for all $t\in\No$, follows from \eqref{teq:inductAssume} since $A_o'=A_o$ and $b_o' = b_o$.

\end{proof}
The last auxiliary result needed in the approximation of the multiplication operation shows that the application of an affine transformation to the output of an RNN can be absorbed into the RNN.
\begin{lemma}\label{lm:linear_map_rnn_output}
	Let $\rnnOpp=\outputOp \hidOp$ be an RNN with $\rnnSizeFunctionOut{\rnnOpp} = d$ and let $A \in \R^{d' \times d}, b\in\R^{d'}$, $d'\in\N$. Then, there exists an RNN $\rnnOpp'=\outputOp' \hidOp$ such that, for all $x\in \R^{\rnnSizeFunctionIn{\rnnOpp}}$, and all $t \in \No$,
	\begin{align}
		(\rnnOpp' \spreadInputOp x)[t] & =  A\left((\rnnOpp \spreadInputOp x)[t]\right) + b. 
	\end{align}
\end{lemma}
\begin{proof}
	For $ \rnnOpp = \outputOp{} \hidOp$, where $\outputOp{}(h) = \Ao{}h + \bo{}$, define
	\[
		\rnnOpp' = \outputOp{}' \hidOp, \quad
		\outputOp{}'(h) = A_o'h + b_o', \textrm{ with } A_o' = A \Ao{}, b_o' = A \bo{} + b.
	\]
	That is, we take the hidden state operator of the modified RNN $\rnnOpp'$ to be identical to that of $\rnnOpp$.
	Noting that 
	\begin{align}
		(\rnnOpp' \spreadInputOp x)[t] = \left(\outputOp{}' \hidOp \spreadInputOp x\right)[t]
		 & = A_o' \left((\hidOp \spreadInputOp x)[t]\right) + b_o'             \\
		 & = A \Ao{} \left((\hidOp \spreadInputOp x)[t]\right) + A \bo{} + b   \\
		 & = A \left( \Ao{} (\hidOp \spreadInputOp x)[t] + \bo{} \right)   + b \\
		 & = A (\rnnOpp{}\spreadInputOp x)[t]   + b, \qquad \text{for all } t\in\No,
	\end{align}
	the proof is completed.
\end{proof}

We can now proceed to the derivation of the approximation result for the multiplication operation.

\begin{theorem}\label{thm:multiplication_net}
	For $D\geq1$, there is an RNN $\rnnMult = \outMult \hidMult$ such that, for all $x=(x_1,x_2)\in[-D, D]^2$, and all $t \in \No$,
	\[
		|(\rnnMult{}\spreadInputOp x)[t] - (x_1 \cdot x_2) | \leq \frac{D^2}{2} \phantom{\cdot} 4^{-t},
	\]
	and
	\begin{align}
		\norm{\hidMult \spreadInputOp{} x)[t]}_\infty \leq 1, \quad |(\rnnMult{} \spreadInputOp{} x)[t]| \leq D^2.
	\end{align}
	Furthermore, $\rnnSizeFunctionHid{ \rnnMult{}} = 14$.
\end{theorem}
\begin{proof}
	We start by employing Lemma \ref{lm:parallel_net} with $N=2$ and $\rnnOpp^1 = \rnnSquare$, $\rnnOpp^2 = \rnnSquare$ according to Theorem \ref{thm:square_net}. This yields an RNN $\rnnOpp = \outputOp \hidOp$ satisfying, for all $z=(z_1,z_2)\in[-D, D]^2$, 
	\begin{equation}\label{teq:double_square_rnn_prop}
		\inorm{(\rnnOpp \spreadInputOp z)[t] - \begin{pmatrix}
				z_1^2 \\
				z_2^2
			\end{pmatrix}} \leq \frac{D^2}{4} 4^{-t}, \quad \inorm{(\hidOp \spreadInputOp z)[t]} \leq 1, \quad \textrm{ and } 0  \, \vecleq \, (\rnnOpp \spreadInputOp z)[t] \, \vecleq \, D^2, \quad \text{for all } t\in\No.  
	\end{equation}
	Next, consider the matrix
	\begin{equation}\label{teq:matdef}
		A = \frac{1}{2}\begin{pmatrix}
			1 & 1  \\
			1 & -1
		\end{pmatrix},
	\end{equation}
	and observe that, for all $x\in[-D, D]^2$,  $Ax \in [-D, D]^2$. We now set $z = A x$ in \eqref{teq:double_square_rnn_prop} and apply Lemma \ref{lm:linear_map_rnn} with $A$ in \eqref{teq:matdef} and $\rnnOpp$ as per \eqref{teq:double_square_rnn_prop} to obtain the RNN $\widetilde{\rnnOpp} = \outputOp \widetilde{\hidOp}$ satisfying, for all $x = (x_1,x_2) \in[-D, D]^2$, 
	\begin{align}
		\inorm{(\tilde{\rnnOpp} \spreadInputOp x)[t] - \begin{pmatrix}
				                                        (\frac{x_1+x_2}{2})^2 \\[1mm]
				                                        (\frac{x_1-x_2}{2})^2
			                                        \end{pmatrix}} & \leq \frac{D^2}{4} 4^{-t}, \label{teq:aa1}        \\[1mm]
		\inorm{(\widetilde{\hidOp} \spreadInputOp x)[t]}                        & \leq 1, \quad \textrm{and } \label{teq:aa2}      \\[1mm]
		0  \, \vecleq \, (\widetilde{\rnnOpp} \spreadInputOp x)[t]       \,           & \vecleq \, D^2,  \label{teq:aa3}
	\end{align}
    for all $t\in\No$. 
    Finally, we apply Lemma \ref{lm:linear_map_rnn_output} to $\widetilde{\rnnOpp}$, with $d'=1,d=2$, $A = \begin{pmatrix} 1 & -1\end{pmatrix}$, and $b=0$ to obtain the RNN $\rnnMult = \outMult \hidMult$, with $\hidMult=\widetilde{\hidOp}$, and note that
	\begin{equation}\label{teq:aa4}
		(\rnnMult{}\spreadInputOp x)[t] = ((\widetilde{\rnnOpp} \spreadInputOp x)[t])_1 - ((\widetilde{\rnnOpp} \spreadInputOp x)[t])_2, \qquad \text{for all }t\in \No.
	\end{equation}
	It thus follows from \eqref{teq:aa3} that $-D^2 \, \vecleq \, (\rnnMult{}\spreadInputOp x)[t] \, \vecleq \, D^2$, for all $t\in\No$.  Finally, we have
	\begin{align}
		|(\rnnMult{}\spreadInputOp x)[t] - (x_1 \cdot x_2)|
		\overset{\eqref{eq:polarisation_identity}} & {=}
		\left|(\rnnMult{}\spreadInputOp x)[t] - \left( \left(\frac{x_1+x_2}{2}\right)^2 - \left(\frac{x_1-x_2}{2}\right)^2\right)\right|                                                                                                     \\
		\overset{\eqref{teq:aa4}}                  & {=}
		\left|((\widetilde{\rnnOpp} \spreadInputOp x)[t])_1 - \left(\frac{x_1+x_2}{2}\right)^2 + \left(\frac{x_1-x_2}{2}\right)^2 -  \left((\widetilde{\rnnOpp} \spreadInputOp x)[t]\right)_2 \right|                                                              \\
		                                           & \leq \left|((\widetilde{\rnnOpp} \spreadInputOp x)[t])_1 - \left(\frac{x_1+x_2}{2}\right)^2\right| + \left|\left((\widetilde{\rnnOpp} \spreadInputOp x)[t]\right)_2 - \left(\frac{x_1-x_2}{2}\right)^2\right| \\
		\overset{\eqref{teq:aa1}}                  & {\leq} \frac{D^2}{4} 4^{-t} + \frac{D^2}{4} 4^{-t} = \frac{D^2}{2} 4^{-t}, \qquad \text{for all } x\in [-D, D], \text{ and all } t\in\No.
	\end{align}
	The proof is concluded upon noting that $\rnnSizeFunctionHid{ \rnnMult{}} = 14$ by Lemma~\ref{lm:parallel_net} applied to $\rnnMult{}$ which contains two parallel instances of $\rnnSquare$ with $\rnnSizeFunctionHid{ \rnnSquare{}} = 7$.  
\end{proof}

\section{Function composition through RNN concatenation}\label{sec:concatenate-RNNs}

The present section develops the recurrent analogue of the structural operations
that play a central role in classical deep feed-forward ReLU network approximation theory. A key objective is to
realize function composition within a single recurrent architecture, something that
cannot be achieved simply by stacking networks as in the feed-forward setting.  
Theorem~\ref{thm:rnn4func_concat} establishes a clocked concatenation mechanism in
which two subnetworks operate in parallel and a time-schedule determines
when the intermediate output of one serves as the input to the other.  
Through this mechanism, RNNs emulate feed-forward network composition
while maintaining a fixed recursive architecture and reusing the same
weights over time.

Our ultimate goal is the approximation of general polynomials in $x$, which requires RNNs that approximate higher powers. 
This will be realized by composing functions, e.g., the map 
 $x \to x^4$ can be expressed by composing the squaring function with itself.
  More generally, consider the functions $f,g$ with associated RNNs $\rnnOpp{}^f$ and $\rnnOpp{}^g$ approximating $f$ and $g$, respectively, 
  in the sense of
\begin{align}
	|(\rnnOpp{}^f \spreadInputOp x)[t] - f(x)| &\leq c_1 4^{-c_2 t}, \qquad \textrm{and }\\
	|(\rnnOpp{}^g \spreadInputOp x)[t] - g(x)| &\leq c_1 4^{-c_2 t}, 
	\qquad \text{for all } t\in\No, \qquad \textrm{ for some } c_1, c_2 > 0.
\end{align}
We wish to construct a new RNN $\rnnOpp'$ that approximates $g \circ f$ such that
\begin{equation}
	|(\rnnOpp' \spreadInputOp x)[t] - g(f(x))| \leq c_3 4^{-c_4 t},
	\qquad \text{for all } t\in\No, \qquad \textrm{ for some } c_3, c_4 > 0.
\end{equation}



\begin{figure}[t]
	\centering
	\includegraphics[width=0.99\textwidth]{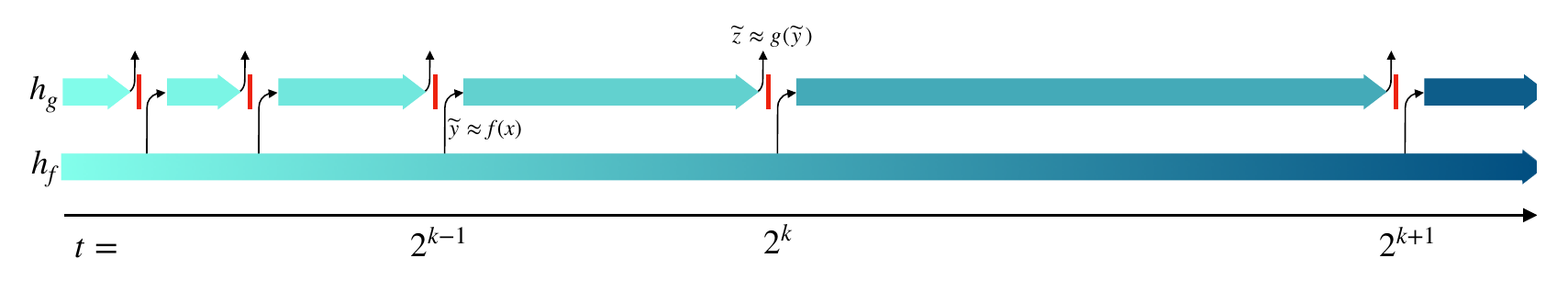}
	\caption{Evolution of the hidden state sequence of the RNN approximating $g\circ f$.}
	\label{fig:concat}
\end{figure}

The core idea is to construct the RNN $\rnnOpp'$ such that it internally runs $\rnnOpp{}^f$ and $\rnnOpp{}^g$ in parallel. Specifically, the hidden state vector of $\rnnOpp'$ consists of two parts, as illustrated in Figure \ref{fig:concat}. The part labeled $h_f$ follows exactly the same recursion as the hidden state of $\rnnOpp_f$ and therefore produces the same output sequence as $\rnnOpp_f$, yielding progressively more accurate approximations of $f(x)$.
The other part, labeled $h_g$, evolves as prescribed by $\rnnOpp_g$, but with the following modification. 
Whenever the time index is a power of $2$, we reinitialize $h_g$ based on the present output of the subnetwork $\rnnOpp{}^f$.
Specifically, as visualized in Figure \ref{fig:concat}, at time index $t=2^{k-1}$ the hidden state $h_g$ is 
set to $0$. Then, the current value of $h_f$ is used to produce an approximation $\ytilde$ of $f(x)$ which, in turn, is employed to reinitialize $h_g$. Subsequently, at time index $t=2^k$, an output approximation $\ztilde$ of $g(f(x))$ is produced, where $\rnnOpp{}^f$ had been running for $2^{k-1}$ time steps to compute $\ytilde \approx f(x)$ and $\rnnOpp{}^g$ ran for $2^{k-1}$ time steps with input $\ytilde$ to approximate $g(\ytilde)$. 
Importantly, $\rnnOpp^f$ continues to run in parallel such that at time step $t=2^{k}$ a refined approximation of $f(x)$ is available to initialize $h_g$ for the next readout at time step $t=2^{k+1}$.

We next show how to realize the clocking mechanism.

\begin{lemma}\label{lem:switch_net}
	Define the sequence
	\[
		\switch[t] = \begin{cases}
			1, & \text{if } t=2^k, \text{ for } k\in\N \text{ with } k\geq2 \\
			0, & \text{else.}
		\end{cases}
	\]
	Let
	\begin{equation}\label{eq:switchnet_weights}
		\switchNetA \coloneqq
		\begin{pmatrix}
			-4 & 2 & 0   & 0 & 0    \\
			-4 & 2 & 0   & 2 & -1/2 \\
			0  & 0 & 1/2 & 0 & -1   \\
			0  & 1 & 1   & 0 & 0    \\
			0  & 0 & 0   & 0 & 0
		\end{pmatrix}
		\quad \textrm{and} \quad 
		\switchNetB \coloneqq
		\begin{pmatrix}
			-1  \\
			1/2 \\
			1   \\
			-2  \\
			1
		\end{pmatrix},
	\end{equation}
	and consider the recursively defined sequence $h[t] \in \R^{5}$, with $h[-1] = 0$ and $h[t] = \relu(\switchNetA h[t-1] + \switchNetB)$, for $t \in \No$. 
    It holds that
	\begin{align}
		(h[t])_1 & = \switch[t+2]
	\end{align}
	and $\inorm{h[t]}\leq 2$, both for all $t\in\No$.
\end{lemma}
{
\newcommand{\hit}[2]{\ensuremath{(h[#2])_{#1}}}
\newcommand{\hi}[1]{\hit{#1}{t}}

\begin{proof}
	The proof is effected through induction over time. We start by showing that
	\begin{equation}
		\label{eq:interleave_induction_base}
		h[2^k-2] =
		\begin{pmatrix}
			1         \\
			2         \\
			2^{2-2^k} \\
			0         \\
			1
		\end{pmatrix}, \qquad
		\text{for all } k\in\N,\; \text{ with } k \geq 2.
	\end{equation}
	The base case, $k=2$, is established through direct computation as
	\begin{equation}
		\begin{aligned}
			h[0] & =
			\relu
			\begin{pmatrix}
				-1  \\
				1/2 \\
				1   \\
				-2  \\
				1
			\end{pmatrix}
			=
			\begin{pmatrix}
				0   \\
				1/2 \\
				1   \\
				0   \\
				1
			\end{pmatrix},
			\qquad
			h[1] =
			\relu
			\begin{pmatrix}
				0    \\
				1    \\
				1/2  \\
				-1/2 \\
				1
			\end{pmatrix}
			=
			\begin{pmatrix}
				0   \\
				1   \\
				1/2 \\
				0   \\
				1
			\end{pmatrix},
			\quad
			\textrm{and}
			\quad
			h[2] =
			\relu
			\begin{pmatrix}
				1      \\
				2      \\
				2^{-2} \\
				-1/2   \\
				1
			\end{pmatrix}
			=
			\begin{pmatrix}
				1      \\
				2      \\
				2^{-2} \\
				0      \\
				1
			\end{pmatrix}.
		\end{aligned}
		\label{depr-teq:first_calculations}
	\end{equation}
 For the induction step, fix $k\geq 2$ and assume that \eqref{eq:interleave_induction_base} holds for this $k$. We first compute
	\begin{align}\label{teq:7}
		h[2^k-1] & = \relu
		\begin{pmatrix}
			-1        \\
			0         \\
			2^{1-2^k} \\
			2^{2-2^k} \\
			1
		\end{pmatrix}
		=
		\begin{pmatrix}
			0         \\
			0         \\
			2^{1-2^k} \\
			2^{2-2^k} \\
			1
		\end{pmatrix},
		         &
		h[2^k]   & = \relu
		\begin{pmatrix}
			-1          \\
			2^{3-2^k}   \\
			2^{-2^k}    \\
			2^{1-2^k}-2 \\
			1
		\end{pmatrix}
		=
		\begin{pmatrix}
			0         \\
			2^{3-2^k} \\
			2^{-2^k}  \\
			0         \\
			1
		\end{pmatrix}.
	\end{align}
	Next, we show that 
	\begin{equation}\label{teq:nested_induction}
		h[2^k+t]
		=
		\begin{pmatrix}
			0           \\
			2^{t+3-2^k} \\
			2^{-2^k-t}  \\
			0           \\
			1
		\end{pmatrix}, \qquad \text{ for all } t\in\{0, 1, \dots, 2^k-3\},
	\end{equation}
    by induction over $t$. The base case, $t=0$, was already established in \eqref{teq:7}. For the induction step, assume that \eqref{teq:nested_induction} holds for some $t\in\{0, 1, \dots, 2^k-4\}$ and compute
	\begin{equation}
		h[2^k+t+1]
		= \relu
		\begin{pmatrix}
			2^{t+4-2^k} - 1             \\
			2^{t+4-2^k}                 \\
			2^{-2^k-t-1}                \\
			2^{t+3-2^k} + 2^{-2^k-t} -2 \\
			1
		\end{pmatrix}
		=
		\begin{pmatrix}
			0              \\
			2^{t+4-2^k}    \\
			2^{-2^k-(t+1)} \\
			0              \\
			1
		\end{pmatrix},
	\end{equation}
	where we used $  2^{t+4-2^k}  \leq 1$ thanks to $t\leq 2^k-4$. This completes the induction over $t$ and thus establishes \eqref{teq:nested_induction}. Particularizing \eqref{teq:nested_induction} to $t=2^k-3$, we have
	\begin{equation}
		h[2^{k+1} - 3 ] 
		= 
		\begin{pmatrix}
			0              \\
			1              \\
			2^{3-2^{k+1} } \\
			0              \\
			1
		\end{pmatrix}.
	\end{equation}
	Next, we compute
	\begin{equation}
		h[2^{k+1}-2]
		= \relu
		\begin{pmatrix}
			1                \\
			2                  \\
			2^{2-2^{k+1}}      \\
			1+2^{3-2^{k+1}} -2 \\
			1
		\end{pmatrix}
		=
		\begin{pmatrix}
			1             \\
			2             \\
			2^{2-2^{k+1}} \\
			0             \\
			1
		\end{pmatrix},
	\end{equation}
	where we used $2^{3-2^{k+1}} \leq 1$, for $k\geq 2$. This
	completes the induction over $k$ and thus establishes \eqref{eq:interleave_induction_base} and, in particular, that
	\begin{align}
		(h[t])_1 = 1, \quad \text{ if } t=2^{k}-2 \text{ for } k \in \N,\; k\geq 2.
	\end{align}
	Inspection of \eqref{teq:7} and \eqref{teq:nested_induction} reveals that, for all other $t$, $(h[t])_1 = 0$.
	Finally, $\inorm{h[t]} \leq 2$, for all $t\in\No$, follows from \eqref{eq:interleave_induction_base} through \eqref{teq:nested_induction}. 
\end{proof}
}

We next define a map, which, based on Lemma \ref{lem:switch_net}, will then be shown to produce an RNN that realizes the desired behavior.
\begin{definition}\label{def:concat}
	Let $\boundOutF, \boundHidG \posConst{}$, and let $\rnnf{}$ and $\rnng $ be RNNs such that $\rnnSizeFunctionIn{\rnnOpp^g} = \rnnSizeFunctionOut{\rnnOpp^f}=:\outputSize{f}$.
	We identify the weights of $\rnnf{}$ as 
	$\Ax{f}, \Ah{f}, \bh{f}, \Ao{f}, \bo{f}$
	and the weights of $\rnng $ as
	$\Ax{g}, \Ah{g}, \bh{g}, \Ao{g}, \bo{g}$, further let  $ \hiddenStateSize{g} := \rnnSizeFunctionHid{\rnng} $, $\hiddenStateSize{f} := \rnnSizeFunctionHid{\rnnf}$, and $\inputSize{f} := \rnnSizeFunctionIn{\rnnf}$.
	We define the mapping $\rnnOpp = \rnnConcatMap{\boundOutF, \boundHidG}{\rnnOpp^g, \rnnOpp^f}$, with 
	weights given by
	\begin{align}
		\begin{split}
			A_h & = \begin{pNiceArray}{c|cc|c|c}
				        \Ah{f} & 0&0 & 0  & 0\\
				        \hline
				        \Ao{f} & 0 & 0  & 0& \boundOutF{}\Ivec{\outputSize{f}}\switchNetReadout{}\\
				        -\Ao{f} & 0 & 0 & 0 & \boundOutF{}\Ivec{\outputSize{f}}\switchNetReadout{}\\
				        \hline
				        0 &  \Ax{g} & -\Ax{g}&\Ah{g} & - \boundHidG{} \Ivec{\hiddenStateSize{g}}\switchNetReadout{}\\
				        \hline
				        0 & 0 & 0 & 0 &  \switchNetA{}\\
			        \end{pNiceArray},   
					 \quad
			b_h = \begin{pNiceArray}{c}
				      \bh{f} \\
				      \hline
				      \bo{f}-\boundOutF{} \Ivec{\outputSize{f}} \\
				      -\bo{f}-\boundOutF{} \Ivec{\outputSize{f}} \\
				      \hline
				      \bh{g} \\
				      \hline
				      \switchNetB{}
			      \end{pNiceArray},
			\quad
			A_x = \begin{pNiceArray}{c}
				      \Ax{f} \\
				      \hline
				      0 \\
				      0 \\
				      \hline
				      0 \\
				      \hline
				      0
			      \end{pNiceArray},
			\\
			A_o & = \begin{pNiceArray}{c|cc|c|c}
				        0 & 0 & 0 & \Ao{g} & 0
			        \end{pNiceArray}, 
					\;\textrm{ and }\;
			b_o = \bo{g},\;
			\textrm{ where }\; \switchNetReadout \coloneqq
			\begin{pmatrix}1 & 0 & 0 & 0& 0\end{pmatrix},
		\end{split}
		\label{eq:def_concat_net}
	\end{align}
	 and 
	$\switchNetA$ and $\switchNetB$ are as in \eqref{eq:switchnet_weights}.
	
	It holds that
	\[
		\rnnSizeFunctionHid{\rnnOpp} = \rnnSizeFunctionHid{\rnnOpp^f} + 2 \rnnSizeFunctionOut{\rnnOpp^f} + \rnnSizeFunctionHid{\rnnOpp^g} + 5.
	\]
	Furthermore, define the mappings
	\begin{equation}
		\label{eq:def_read_out_matrices}
		\subMatInMap{}({\rnnOpp^g, \rnnOpp^f}) \coloneqq \begin{pNiceArray}{c|cc|c|c}
			\Imat{\hiddenStateSize{f}} & 0 & 0 & 0 & 0
		\end{pNiceArray}, \quad \textrm{and} \quad
		\subMatOutMap{}({\rnnOpp^g, \rnnOpp^f}) \coloneqq \begin{pNiceArray}{c|cc|c|c}
			0 & 0 & 0 & \Imat{\hiddenStateSize{g}} & 0
		\end{pNiceArray}.
	\end{equation}
\end{definition}
We next establish the properties of the RNN produced by the map $\rnnConcatMap{\boundOutF, \boundHidG}{\rnnOpp^g, \rnnOpp^f}$.
\begin{theorem}\label{thm:rnn4func_concat}
	Let $D, \boundOutG \posConst{}$, and let $\boundOutF, \boundHidG \posConst{}$, and $\rnnf=\outputOp{}^f \hidOp{}^f$, $\rnng = \outputOp{}^g \hidOp{}^g$ be as in Definition \ref{def:concat}.
	Assume that 
	\begin{enumerate}[label=(A\arabic*)]
		\item \label{assume:first_output_bounded} $\inorm{(\rnnOp{}^f \spreadInputOp x)[t]} \leq \boundOutF$,\; for $x\in[-D, D]^{\inputSize{f}}$,\; $t\in \No$.
		\item \label{assume:second_hidden_bounded} $\inorm{(\hidOp^g \spreadInputOp y)[t]} \leq \boundHidG$,\; for $y\in[-\boundOutF, \boundOutF]^{\outputSize{f}}$,\; $t\in \No$.
	\end{enumerate}
	Then, the RNN $ \rnnConcatMap{\boundOutF, \boundHidG}{\rnnOpp^g, \rnnOpp^f} =: \rnnOp{}=\outputOp{} \hidOp{}$ specified in Definition \ref{def:concat} satisfies, for all $x\in[-D, D]^{\inputSize{f}}$,
	%
    \begin{align}
		\begin{split}
			\label{eq:concat_rnn_output}
			(\rnnOp{} \spreadInputOp{}x)[2^k - 2] =
			  \bigl(\rnnOp{}^{g} \spreadInputOp \bigl(            
			     (\rnnOp{}^f \spreadInputOp{} x)[2^{k-1}-2] 
			  \bigr)\bigr)[2^{k-1}-2], \quad \text{ for all } k \in \N, \text{ with } k\geq 3.
		\end{split}
	\end{align}
	Furthermore, with $\subMatIn = \subMatInMap{}({\rnnOpp^g, \rnnOpp^f})$ and $\subMatOut = \subMatOutMap{}({\rnnOpp^g, \rnnOpp^f})$, we have
    \begin{align}
		\subMatIn{} (\hidOp{} \spreadInputOp{} x) [t]      & = (\hidOp{}^f \spreadInputOp{}x)[t],                                                                            
		& \text{for all } & t\in \No, \label{eq:prop_subMatIn}             \\
		\subMatOut{} (\hidOp{} \spreadInputOp{} x) [2^k-2] & = \left(\hidOp{}^{g} \spreadInputOp \left((\rnnOp{}^f \spreadInputOp{} x)[2^{k-1}-2] \right)\right)[2^{k-1}-2], & \text{for all } &  k \in \N, \text{ with } k\geq 3. \label{eq:prop_subMatOut}
	\end{align}
	Additionally assuming  
	\begin{enumerate}[label=(A\arabic*), resume]
		\item \label{assume:f_hid_bound} $\inorm{(\hidOp{}^f \spreadInputOp x)[t]} \leq \boundHidF$,\;  for $x\in[-D, D]^{\inputSize{f}}$,\; $t\in \No$,
		\item \label{assume:g_out_bound} $\inorm{(\rnnOpp{}^g \spreadInputOp y)[t]} \leq \boundOutG$,\; for  $y\in[-\boundOutF, \boundOutF]^{\outputSize{f}}$,\; $t\in \No$,
	\end{enumerate}
	we have, for all $x\in[-D, D]^{\inputSize{f}}$, 
	\begin{enumerate}[label=(B\arabic*)]
		\item \label{bound_hid}  $\inorm{(\hidOp{} \spreadInputOp x)[t]} \leq \max\{2, \boundOutF, \boundHidG\}$,\; for all $t\in \No$,
		\item \label{bound_out}  $\inorm{(\rnnOp{} \spreadInputOp x)[t]} \leq \boundOutG$,\; for all $t\in \No$.
	\end{enumerate}
\end{theorem}
\begin{proof}
	Arbitrarily fix $x\in[-D, D]^{\inputSize{f}}$ and consider the hidden state sequence in response to the input sequence $(\spreadInputOp x)[\cdot]$, i.e., $h[t] = (\hidOp \spreadInputOp{} x)[t]$. We divide this hidden state sequence corresponding to the blocks in \eqref{eq:def_concat_net} according to
	\begin{equation}
		h[t] = \begin{pNiceArray}{c}h_1[t] \\\hline  h_2[t]  \\\hline  h_3[t]  \\\hline h_4[t] \end{pNiceArray}  = \relu \left( A_h h[t-1] + A_x( (Dx)[t] ) + b_h  \right),
	\end{equation}
	and analyze each block separately. First, note that $h_4[\cdot]$ follows 
	\begin{equation}\label{teq:switchnetrecursion}
		h_4[t] = \relu(\switchNetA h_4[t-1] + \switchNetB{}), \, t \in \No, \text{ with } h_4[-1] = 0.
	\end{equation}
	This recursion implements the clocking mechanism defined in Lemma \ref{lem:switch_net}, generating the control pulses that govern the alternation between the $\rnnOpp^f$ and $\rnnOpp^g$ computations. By Lemma \ref{lem:switch_net}, we have, for all $t \in \No$,
	\begin{equation}\label{teq:switch}
		 \switchNetReadout h_4[t] = \switch[t+2].
	\end{equation}
	Next, we note that $h_1[t]$ follows the recursion
	\begin{equation}
		h_1[t] = \relu\left(\Ah{f} h_1[t-1] + \Ax{f} (\spreadInputOp{} x)[t] + \bh{f}\right), \, t \in \No, \quad \textrm{with } h_1[-1] = 0.
	\end{equation}
	Thus, recalling Definition \ref{def:elman_rnn}, we have 
	\begin{equation}\label{teq:h1_evolution}
	h_1[t] = (\hidOp^f \spreadInputOp{} x)[t].
	\end{equation}
	Next, we consider the sequence $h_2[t]$, which is given by
	\begin{align}
		h_2[t]
		 & = \relu
		\begin{pmatrix}
			\Ao{f} h_1[t-1] + \boundOutF{}\Ivec{\outputSize{f}}\switchNetReadout h_4[t-1] + \bo{f} - \boundOutF{}{} \Ivec{\outputSize{f}}  \\
			-\Ao{f} h_1[t-1] + \boundOutF{}\Ivec{\outputSize{f}}\switchNetReadout h_4[t-1] - \bo{f} - \boundOutF{}{} \Ivec{\outputSize{f}} \\
		\end{pmatrix}                   \\
		 & = \relu
		\begin{pmatrix}
			(\Ao{f}  (\hidOp^f \spreadInputOp{} x)[t-1]+ \bo{f}) - \boundOutF{}\Ivec{\outputSize{f}}(1 - \switchNetReadout h_4[t-1])   \\
			-(\Ao{f}  (\hidOp^f \spreadInputOp{} x)[t-1] + \bo{f}) - \boundOutF{}\Ivec{\outputSize{f}}(1 - \switchNetReadout h_4[t-1]) \\
		\end{pmatrix} \\
		  \overset{\eqref{teq:switch}}&{=}\relu
		\begin{pmatrix}
			(\outputOp^f \hidOp^f \spreadInputOp{} x)[t-1] - \boundOutF{}\Ivec{\outputSize{f}}(1 - \switch[t+1])   \\
			-(\outputOp^f  \hidOp^f \spreadInputOp{} x)[t-1] - \boundOutF{}\Ivec{\outputSize{f}}(1 - \switch[t+1]) \\
		\end{pmatrix}          \\
		 \overset{(i)}&{=} \relu
		\begin{pmatrix}
			(\rnnOp{}^f \spreadInputOp{} x)[t-1]  \\
			-(\rnnOp{}^f \spreadInputOp{} x)[t-1] \\
		\end{pmatrix} \switch[t+1],
			\label{teq:h2}
	\end{align}
	where in (i) we used $\inorm{(\outputOp^f \hidOp^f \spreadInputOp{} x)[t-1] }
	= \inorm{(\rnnOp{}^f \spreadInputOp{} x)[t-1]} \leq \boundOutF{}$, for all $t\in\No$, by assumption \ref{assume:first_output_bounded}.
	From \eqref{teq:h2} it now follows that $h_3[t]$, $t \in \No$, is given by the following recursion, with $h_3[-1]=0$,
    \begin{align}
		h_3[t] 
		&=
		\relu\biggl(
		\Ax{g} \relu((\rnnOp{}^f \spreadInputOp{} x)[t-2]) \switch[t] - \Ax{g}\relu(-(\rnnOp{}^f \spreadInputOp{} x)[t-2])\switch[t] \\
		         & + \Ah{g}h_3[t-1] + \bh{g} - \boundHidG{} \Ivec{\hiddenStateSize{g}}\switchNetReadout h_4[t-1]
		\biggr)                                                                                                                          \\
		         \overset{(ii)}&{=} \relu\left(
		\Ah{g}h_3[t-1] + \bh{g} - \boundHidG{} \Ivec{\hiddenStateSize{g}}\switchNetReadout h_4[t-1] +  \switch[t] \cdot \Ax{g}(\rnnOp{}^f \spreadInputOp{} x)[t-2]
		\right)\\
		\overset{\eqref{teq:switch}}&{=} \relu\left(
		\Ah{g}h_3[t-1] + \bh{g} - \boundHidG{} \Ivec{\hiddenStateSize{g}}\switch[t+1] +  \switch[t] \cdot \Ax{g}(\rnnOp{}^f \spreadInputOp{} x)[t-2]
		\right),
	\end{align}
	where in (ii) we used the identity $x = \relu(x)-\relu(-x)$.
	As $\switch[\ell]=0,$ for $\ell\in\{0,\dots, 3\}$, it follows that
	\begin{equation}\label{teq:h3_start}
		h_3[t]=(\hidOp^g \spreadInputOp{}0)[t],\quad \textrm{ for } t\in\{0, 1, 2\}.
	\end{equation} 
    Next, we compute
	\begin{align}
		h_3[3] & = \relu\left(
		\Ah{g}h_3[2] + \bh{g} - \boundHidG{} \Ivec{\hiddenStateSize{g}}\switch[4] +  \switch[3] \cdot \Ax{g}(\rnnOp{}^f \spreadInputOp{} x)[1]
		\right)                \\
		       & = \relu\left(
		\Ah{g}h_3[2] + \bh{g} - \boundHidG{} \Ivec{\hiddenStateSize{g}}
		\right)=0, \label{teq:compute3}
	\end{align}
	where we used 
    $ \inorm{(\hidOp^g \spreadInputOp{}0)[3]} = \inorm{\relu(\Ah{g}h_3[2] + \bh{g})} \leq\boundHidG$, thanks to Assumption \ref{assume:second_hidden_bounded}.
	We now prove by nested induction that
	\begin{align}
		h_3[2^k-1]      & = 0,             & \text{for } k                                      & \in\N,\; k\geq 2, \quad \label{teq:induction_3a} \\
		\text{and}\quad
		h_3[2^k + \ell] & =
		\left( \hidOp^g \spreadInputOp \left((\rnnOp{}^f \spreadInputOp{} x)[2^k-2] \right)\right)[\ell],
		                & \text{for } \ell & \in \{0, \dots, 2^k - 2\}.\label{teq:induction_3b}
	\end{align}
	The base case for the induction over $k$, i.e., \eqref{teq:induction_3a} for $k=2$, was already established in \eqref{teq:compute3}.
    Next, we assume that \eqref{teq:induction_3a} holds for some $k\geq2$ and compute
	\begin{align}
		h_3[2^k] & = \relu\left(
		\Ah{g}h_3[2^k-1] + \bh{g} - \boundHidG{} \Ivec{\hiddenStateSize{g}}\switch[2^k+1] +  \switch[2^k] \cdot \Ax{g}(\rnnOp{}^f \spreadInputOp{} x)[2^k-2]
		\right)                                                                                                     \\
		         & = \relu\left(
		\bh{g}  + \Ax{g}(\rnnOp{}^f \spreadInputOp{} x)[2^k-2]
		\right)                                                                                                     \\
		         & = \left(\hidOp^g \spreadInputOp{}\left((\rnnOp{}^f \spreadInputOp{} x)[2^k-2] \right)\right)[0],
	\end{align}
	where in the last step we used Remark \ref{rm:operator_rnn_description}.
	This establishes the base case $\ell = 0$ for the induction over $\ell$ in \eqref{teq:induction_3b}. Next, assume that \eqref{teq:induction_3b} holds for some $\ell\in\{0, \dots, 2^k-3\}$ and compute
	\begin{align}
		h_3[2^k + \ell+1] & = \relu\biggl(
		\Ah{g}h_3[2^k+\ell] + \bh{g} - \boundHidG{} \Ivec{\hiddenStateSize{g}}\switch[2^k+\ell+2]                                 \\
		                  &  +  \switch[2^k+\ell+1] \cdot \Ax{g}(\rnnOp{}^f \spreadInputOp{} x)[2^k+\ell-1]
		\biggr)                                                                                                                   \\
		                  & = \relu\left(
		\Ah{g}h_3[2^k +\ell]  + \bh{g}
		\right)                                                                                                                   \\
		                   \overset{\eqref{teq:induction_3b}}&{=} \relu\left(
		\Ah{g}\left( \hidOp^g \spreadInputOp \left((\rnnOp{}^f \spreadInputOp{} x)[2^k-2] \right)\right)[\ell]  + \bh{g}
		\right)                                                                                                                   \\
		                   \overset{\textrm{Rem. \ref{rm:operator_rnn_description}}}&{=} \left( \hidOp^g \spreadInputOp \left((\rnnOp{}^f \spreadInputOp{} x)[2^k-2] \right)\right)[\ell+1].
	\end{align}
	This completes the induction over $\ell$ and thus establishes \eqref{teq:induction_3b}.
	We now proceed by noting that
	\begin{align}
		h_3[2^{k+1}-1]  & = \relu\biggl(
		\Ah{g}h_3[2^k+2^k-2] + \bh{g} - \boundHidG{} \Ivec{\hiddenStateSize{g}}\switch[2^{k+1}]                                       \\
		                                        &  +  \switch[2^{k+1}-1] \cdot \Ax{g}(\rnnOp{}^f \spreadInputOp{} x)[2^{k+1}-3]
		\biggr)                                                                                                                       \\
		                                        \overset{(iii)} &{=} \relu\left(
		\Ah{g}\left( \hidOp^g \spreadInputOp \left((\rnnOp{}^f \spreadInputOp{} x)[2^k-2] \right)\right)[2^k-2]  + \bh{g} - \boundHidG{}\Ivec{\hiddenStateSize{g}}
		\right)                                                                                                                       \\
		                                         \overset{(iv)}&{=} 0, \label{teq:use_bound_on_hidden_state}
	\end{align}
	where (iii) follows from \eqref{teq:induction_3b} with $\ell=2^k-2$ and (iv) holds because
	\begin{align}
		\Ah{g}\left( \hidOp^g \spreadInputOp \left((\rnnOp{}^f \spreadInputOp{} x)[2^k-2] \right)\right)[2^k-2]  + \bh{g} 
		& \, \vecleq \,
		\norm{\relu(\Ah{g}\left( \hidOp^g \spreadInputOp \left((\rnnOp{}^f \spreadInputOp{} x)[2^k-2] \right)\right)[2^k-2]  + \bh{g})}_\infty                                                                                  \\
		\overset{\ref{assume:first_output_bounded}}&{\leq}
		\max_{\inorm{x'}\leq \boundOutF{}{}} \norm{ \relu(\Ah{g}\left( \hidOp^g \spreadInputOp x' \right)[2^k-2]  + \bh{g})}_\infty                                                                                                 \\
		\overset{\textrm{Rem. \ref{rm:operator_rnn_description}}}&{=} \max_{\inorm{x'}\leq \boundOutF{}{}} \norm{\left( \hidOp^g \spreadInputOp x' \right)[2^k-1]}_\infty
		\overset{\ref{assume:second_hidden_bounded}}{\leq} \boundHidG{}.
	\end{align}
	This establishes that \eqref{teq:induction_3a} holds for $k+1$ as well and thus completes the nested induction.
	The overall network output is given by
	$(\rnnOp{} \spreadInputOp{}x)[t] = \Ao{g}h_3[t] + \bo{g},$
	which, using \eqref{teq:induction_3b}, yields
	\begin{align}
		(\rnnOp{} \spreadInputOp{}x)[2^k + \ell] & =
		\left(\rnnOp{}^{g} \spreadInputOp \left((\rnnOp{}^f \spreadInputOp{} x)[2^{k}-2] \right)\right)[\ell],  &   & \text{for all } k \in \N, \text{with } k\geq2, \textrm{ and all }\ell\in\{0,\dots,  2^k-2\}. \label{teq:rnn_output}
	\end{align}
	In particular, setting $\ell = 2^k-2$, we obtain 
	\begin{equation}\label{teq:h3_evolution}
		(\rnnOp{} \spreadInputOp{}x)[2^{k+1} - 2]  =  \left( \rnnOpp^g \spreadInputOp \left((\rnnOp{}^f \spreadInputOp{} x)[2^k-2] \right)\right)[2^k-2],
	\end{equation}
	which establishes \eqref{eq:concat_rnn_output}.
	Moreover, \eqref{eq:prop_subMatIn} is an immediate consequence of Definition \ref{def:concat} and \eqref{teq:h1_evolution}. Finally,
	\eqref{eq:prop_subMatOut} follows from Definition \ref{def:concat} and \eqref{teq:induction_3b} with $\ell=2^k-2$.

	To establish \ref{bound_hid}, we note that 
    $\inorm{h_1[t]} \leq \boundHidF$ by \eqref{teq:h1_evolution} and Assumption \ref{assume:f_hid_bound}, $\inorm{h_2[t]} \leq \boundOutF$ by \eqref{teq:h2} and Assumption \ref{assume:first_output_bounded}, $\inorm{h_3[t]}\leq \boundHidG$ by \eqref{teq:induction_3a} and \eqref{teq:induction_3b} together with Assumptions \ref{assume:first_output_bounded} and \ref{assume:second_hidden_bounded}, and finally
	$\inorm{h_4[t]} \leq 2$ by \eqref{teq:switchnetrecursion} and Lemma \ref{lem:switch_net}. Taking the maximum over these bounds, we arrive at \ref{bound_hid}. The proof is concluded upon noting that \ref{bound_out} follows from \eqref{teq:rnn_output} together with Assumption \ref{assume:g_out_bound}.
\end{proof}

\begin{figure}[h]
	\centering
	\includegraphics[width=0.7\textwidth]{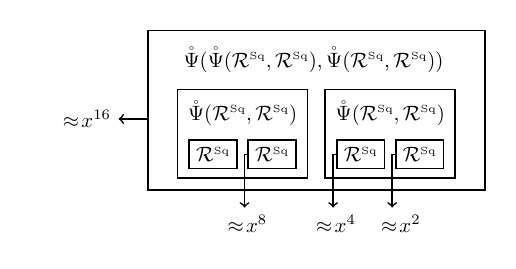}
	\caption{RNN approximating $x\to x^{16}$.}
	\label{fig:multiconcat}
\end{figure}

We now illustrate, through a simple example, the concatenation mechanism developed in 
Theorem~\ref{thm:rnn4func_concat} and prepare for its generalization in 
Lemma~\ref{lem:multiconcat_rnn_tree}. 
Consider the function $x \mapsto x^{16}$, which can be expressed as 
$\squareFunc \circ \squareFunc \circ \squareFunc \circ \squareFunc$, where 
$\squareFunc: x \mapsto x^2$. 
Accordingly, $x^{16}$ can be approximated by concatenating four copies of $\rnnSquare$, 
the RNN introduced in Theorem~\ref{thm:square_net}. 
Using the map from Definition~\ref{def:concat}, we construct the network 
\[
\rnnConcatMap{}{\rnnConcatMap{}{\rnnSquare, \rnnSquare},\,
\rnnConcatMap{}{\rnnSquare, \rnnSquare}},
\]
depicted in Figure~\ref{fig:multiconcat}, whose output sequence approximates $x^{16}$. 
Furthermore, suitably exploiting \eqref{eq:prop_subMatIn} and \eqref{eq:prop_subMatOut} 
makes it possible to also recover approximations of the intermediate functions 
$x^2$, $x^4$, and~$x^8$. 
The following result, Lemma~\ref{lem:multiconcat_rnn_tree}, formalizes this idea by extending the 
two-network concatenation, which implements function composition, of Theorem~\ref{thm:rnn4func_concat} to the concatenation of multiple RNNs.

\begin{lemma}\label{lem:multiconcat_rnn_tree}
	Let $D>0$, $L \in \N$, and let $\rnnOp{}^1, \dots, \rnnOpp{}^{2^L}$ be RNNs with $\rnnSizeFunctionIn{\rnnOpp^{\ell+1}}=\rnnSizeFunctionOut{\rnnOpp^{\ell}}$, $\ell \in \{1, \dots, 2^{L}-1\}$.
    Assume that there are constants $D_1, \dots, D_{2^L}$ and $\hidBound{}$ with
    $\max\{2, \max_{\ell\in\{1,\dots, 2^L\}} D_\ell\} \leq \hidBound{}$ such that,
	for all $\ell\in\{1, \dots, 2^L\}$, 
	\begin{align}
		\sup_{\inorm{x} \leq D_{\ell-1}, t\in\No }\inorm{(\rnnOpp^{\ell} \spreadInputOp{} x)[t]} & \leq D_{\ell}, \label{teq:assumption_concat3}  \\
		\sup_{\inorm{x} \leq D_{\ell-1}, t\in\No }\inorm{(\hidOp^{\ell} \spreadInputOp{} x)[t]}  & \leq \hidBound{},       \label{teq:assumption_concat4}
	\end{align}
	where we set $D_0 = D$. Define the mappings 
	\begin{equation}\label{eq:def_multiconcat_maps}
		\multiconcatRNNmap{\ell}{k}: 
        \R^{\rnnSizeFunctionIn{\rnnOpp^{\ell}}} \to \R^{\rnnSizeFunctionOut{\rnnOpp^{\ell}}}
        \; \textrm{ as } \; x \to \left(\rnnOpp^\ell \spreadInputOp{} x\right)[2^k -2], \qquad
		\ell \in \{1, \dots, 2^{L}\}, \, k\geq 2.
	\end{equation}
	Then, there exists an RNN
	$\rnnMultiConcatTree^L = \outOpMultiConcatTree{}^L \hidOpMultiConcatTree{}^L $ such that
	\begin{align}\label{eq:bound_of_multiconcat_net}
		\sup_{\inorm{x} \leq D, t\in\No } \inorm{\left(\rnnMultiConcatTree^L \spreadInputOp x\right)[t]} \leq D_{2^L},
		\quad 
		\sup_{\inorm{x} \leq D, t\in\No } \inorm{\left(\hidOpMultiConcatTree^L \spreadInputOp x\right)[t]} \leq \hidBound{},
	\end{align}
	and
	\begin{equation}\label{eq:multiconcat_rnn_output}
		\left(\rnnMultiConcatTree^L \spreadInputOp x\right)[2^k - 2 ] =
		\left(
		\multiconcatRNNmap{2^L}{k-L} \circ \dots \circ \multiconcatRNNmap{1}{k-L}
		\right)(x), \quad
		k \geq L+2, \; x\in[-D, D]^{\rnnSizeFunctionIn{\rnnOpp^1}}.
	\end{equation} 
	Furthermore, for every $\ell \in \{1, \dots, 2^L\}$, there are $A^\ell, b^\ell$, and $\timeOffsetIdx{1}, \dots, \timeOffsetIdx{\ell} \in \{0, \dots,  \ceil{\log(\ell)}\}$ such that
	\begin{equation}\label{eq:multiconcat_hid_outputs}
		A^\ell\left(\hidOpMultiConcatTree{}^L \spreadInputOp x\right)[2^k - 2 ] + b^\ell =
		\left(
		\multiconcatRNNmap{\ell}{k-\timeOffsetIdx{\ell}} \circ \dots \circ \multiconcatRNNmap{1}{k-\timeOffsetIdx{1}}
		\right)(x),\qquad
		\delforall k \geq L+2, \; \delforall x\in[-D, D]^{\rnnSizeFunctionIn{\rnnOpp^1}}.
	\end{equation}
	Finally, we have
	\begin{equation}\label{eq:multiconcat_sizes}
		\rnnSizeFunctionHid{\rnnMultiConcatTree{}^L} = \sum_{\ell=1}^{2^L} \rnnSizeFunctionHid{\rnnOpp^\ell} + 2 \sum_{\ell=1}^{2^L-1} \rnnSizeFunctionOut{\rnnOpp^\ell} + 5 (2^L-1).
	\end{equation}
\end{lemma}

\begin{proof}
	We prove the statement by induction over $L$. For the base case, $L=1$, given $\rnnOpp^1 = \outputOp^1 \hidOp^1$ and $\rnnOpp^2 = \outputOp^2 \hidOp^2$, 
    we take $\rnnMultiConcatTree^1 = \outOpMultiConcatTree{}^1 \hidOpMultiConcatTree{}^1  = \rnnConcatMap{\hidBound{}, \hidBound{}}{ \rnnOpp^2, \rnnOpp^1}$ according to Definition \ref{def:concat} and $\subMatIn^1 = \subMatInMap(\rnnOpp^2, \rnnOpp^1)$, $\subMatOut^1 = \subMatOutMap(\rnnOpp^2, \rnnOpp^1)$.
    By assumption, we have
	\begin{equation}\label{teq:16}
		\sup_{\inorm{x}\leq D, t\in\No}\inorm{(\rnnOpp^1 \spreadInputOp x)[t]} \leq D_1\leq \hidBound{}
		\;\textrm{ and } \;
		\sup_{\inorm{x'}\leq D_1, t\in\No}\inorm{(\hidOp^2 \spreadInputOp x')[t]} \leq \hidBound{},
	\end{equation}
	as well as
	\begin{equation}\label{teq:17}
		\sup_{\inorm{x}\leq D, t\in\No}\inorm{(\hidOp^1 \spreadInputOp x)[t]} \leq \hidBound{}
		\;\textrm{ and }\;
		\sup_{\inorm{x'}\leq D_1, t\in\No}\inorm{(\rnnOpp^2 \spreadInputOp x')[t]} \leq D_2.
	\end{equation}
	We now invoke Theorem \ref{thm:rnn4func_concat} with $\rnnf=\rnnOpp^1$ and $\rnng=\rnnOpp^2$ and note that
    its conditions \ref{assume:first_output_bounded} and \ref{assume:second_hidden_bounded} are satisfied thanks to \eqref{teq:16}, and
    \ref{assume:f_hid_bound} and \ref{assume:g_out_bound} are met owing to \eqref{teq:17}. We thus have \eqref{eq:bound_of_multiconcat_net} for $L=1$ as a consequence of \ref{bound_hid} and \ref{bound_out} in Theorem \ref{thm:rnn4func_concat}. Furthermore, it follows from \eqref{eq:concat_rnn_output} in Theorem \ref{thm:rnn4func_concat} that
    \begin{align} \label{beq:1}
		(\rnnMultiConcatTree^1 \spreadInputOp{}x)[2^k - 2] = 
		  \bigl(\rnnOpp^2 \spreadInputOp \bigl(              
		    (\rnnOpp^1 \spreadInputOp{} x)[2^{k-1}-2] 
		  \bigr)\bigr)[2^{k-1}-2],
		\,\,  \text{for all } k\geq3,  \; \text{and all } x\in[-D, D]^{\rnnSizeFunctionIn{\rnnOpp^1}},
	\end{align}
	which, upon invoking \eqref{eq:def_multiconcat_maps}, yields \eqref{eq:multiconcat_rnn_output}.
	Next, we establish \eqref{eq:multiconcat_hid_outputs}. To this end, let
	\begin{equation}
		A^1 \coloneqq \Ao{1} \subMatIn^1, \: b^1 = \bo{1},
		\textrm{ and }
		A^2 \coloneqq \Ao{2} \subMatOut^1, \: b^2 = \bo{2},
	\end{equation}
	where $\Ao{1}, \bo{1}$ and $\Ao{2}, \bo{2}$ are the weights of the affine output mappings for the RNNs $\rnnOpp{}^1$ and $\rnnOpp{}^2$, respectively. We have, for all $x$ with $\inorm{x}\leq D$ and all $k\geq 3$, 
	\begin{align}
		A^1\left(\hidOpMultiConcatTree{}^1 \spreadInputOp x\right)[2^k - 2 ] + b^1 & =
		\Ao{1} \subMatIn^1 \left(\hidOpMultiConcatTree{}^1 \spreadInputOp x\right)[2^k - 2 ] + \bo{1}\\
		\overset{\eqref{eq:prop_subMatIn}}&{=} \Ao{1}  \left(\hidOp{}^1 \spreadInputOp x\right)[2^k - 2 ] + \bo{1} \\
		& = \left(\rnnOpp{}^1 \spreadInputOp x\right)[2^k - 2 ]                 \\
		\overset{\eqref{eq:def_multiconcat_maps}}&{=}  \multiconcatRNNmap{1}{k-0}(x).
	\end{align}
	Hence, this establishes \eqref{eq:multiconcat_hid_outputs} for $\ell=1$, with $\timeOffsetIdx{1} = 0$. Furthermore, we have, for all $x$ with $\inorm{x}\leq D$ and all $k\geq 3$,
	\begin{align}
		A^2\left(\hidOpMultiConcatTree{}^1 \spreadInputOp x\right)[2^k - 2 ] + b^2 & =
		\Ao{2} \subMatOut^1 \left(\hidOpMultiConcatTree{}^1 \spreadInputOp x\right)[2^k - 2 ] + \bo{2}    \\
		\overset{\eqref{eq:prop_subMatOut}}&{=} \Ao{2}  \left(\hidOp{}^2 \spreadInputOp \left(\rnnOpp{}^1 \spreadInputOp x\right)[2^{k-1} - 2 ]\right)[2^{k-1}-2] + \bo{2} \\
		& =  \left(\rnnOpp{}^2 \spreadInputOp \left(\rnnOpp{}^1 \spreadInputOp x\right)[2^{k-1} - 2 ]\right)[2^{k-1}-2]                \\
		\overset{\eqref{eq:def_multiconcat_maps}}&{=} \left(\multiconcatRNNmap{2}{k-1} \circ \multiconcatRNNmap{1}{k-1}\right)(x). 
	\end{align}
    This establishes \eqref{eq:multiconcat_hid_outputs} for $\ell=2$, with $\timeOffsetIdx{1} = \timeOffsetIdx{2} = 1$. Finally, by Definition \ref{def:concat}, we obtain 
	\[
		\rnnSizeFunctionHid{\rnnMultiConcatTree^1} = \rnnSizeFunctionHid{\rnnOpp{}^1} + 2 \rnnSizeFunctionOut{\rnnOpp{}^1} + \rnnSizeFunctionHid{\rnnOpp{}^2} + 5,
	\]
	which yields \eqref{eq:multiconcat_sizes} for $L=1$. This finishes the proof of the base case $L=1$ for the induction over $L$.
	
	We proceed to the induction step and assume that Lemma \ref{lem:multiconcat_rnn_tree} holds for some $L\geq 1$. 
    Specifically, let $\rnnOpp{}^1, \dots, \rnnOpp{}^{2^{L+1}}$ be RNNs satisfying \eqref{teq:assumption_concat3} and \eqref{teq:assumption_concat4} with constants $D_1, \dots, D_{2^{L+1}}, \hidBound{}$. We first invoke Lemma \ref{lem:multiconcat_rnn_tree} for $\rnnOpp{}^1, \dots, \rnnOpp{}^{2^L}$ and denote the resulting overall RNN as $\rnnMultiConcatTree{}^{a} = \outOpMultiConcatTree{}^a \hidOpMultiConcatTree{}^a$. By \eqref{eq:bound_of_multiconcat_net} we have
	\begin{equation}\label{teq:40}
		\inorm{\left(\rnnMultiConcatTree{}^a \spreadInputOp x\right)[t]} \leq D_{2^L}
		\quad \textrm{ and } \quad
		\inorm{\left(\hidOpMultiConcatTree{}^a \spreadInputOp x\right)[t]} \leq \hidBound{},
		\quad \text{for all } t \in \No , \; \text{and all } x\in[-D, D]^{\rnnSizeFunctionIn{\rnnOpp^1}}.
	\end{equation}
	Similarly, we
	invoke Lemma \ref{lem:multiconcat_rnn_tree} for the $2^L$ RNNs $\rnnOpp{}^{2^L+1}, \dots, \rnnOpp{}^{2^{L+1}}$ and denote the resulting overall RNN by $\rnnMultiConcatTree{}^{b}=\outOpMultiConcatTree{}^b \hidOpMultiConcatTree{}^b$, which satisfies
	\begin{equation}\label{teq:41}
		\inorm{\left(\rnnMultiConcatTree{}^b \spreadInputOp x\right)[t]} \leq D_{2^{L+1}}
		\, \textrm{and} \,
		\inorm{\left(\hidOpMultiConcatTree{}^b \spreadInputOp x\right)[t]} \leq \hidBound{},
		\, \text{for all } t \in \No , \; \text{and all } x\in[-D_{2^{L}}, D_{2^L}]^{\rnnSizeFunctionIn{\rnnOpp^{2^{L}+1}}}.
	\end{equation}
	Now, set $\rnnMultiConcatTree{}^{L+1} = \outOpMultiConcatTree{}^{L+1} \hidOpMultiConcatTree{}^{L+1}= \rnnConcatMap{\hidBound{}, \hidBound{}}{ \rnnMultiConcatTree{}^b , \rnnMultiConcatTree{}^a}$ and invoke Theorem \ref{thm:rnn4func_concat} with the following correspondence of quantities:
    \begin{center}
    \begin{tabular}{c|c}
        Here & Theorem \ref{thm:rnn4func_concat}
         \\
         \hline 
         $\rnnMultiConcatTree{}^a$ & $\rnnOpp^f$ \\
        $\rnnMultiConcatTree{}^b$ & $\rnnOpp^g$ \\
        $D$ & $D$ \\
        $\hidBound{}$ &  $\hidBound{}$ \\
        $D_{2^L}$ &  $\boundOutF$\\
        $D_{2^{L+1}}$ & $ \boundOutG$ \\
    \end{tabular}
\end{center}
    By \eqref{teq:40}, Conditions \ref{assume:first_output_bounded} and \ref{assume:f_hid_bound} of Theorem \ref{thm:rnn4func_concat} are satisfied. 
    Further, by \eqref{teq:41}, Conditions \ref{assume:g_out_bound} and \ref{assume:second_hidden_bounded} of Theorem \ref{thm:rnn4func_concat} are met. 
    %
%
%
%
	Hence, \eqref{eq:bound_of_multiconcat_net} holds for $\rnnMultiConcatTree{}^{L+1}$ as a consequence of \ref{bound_hid} and \ref{bound_out} in Theorem \ref{thm:rnn4func_concat}. By \eqref{eq:concat_rnn_output} in Theorem \ref{thm:rnn4func_concat} we have
    \begin{equation}
        \label{beq:2}
		(\rnnMultiConcatTree^{L+1} \spreadInputOp{}x)[2^k - 2] =
		  \bigl(\rnnMultiConcatTree^b \spreadInputOp \bigl(              
		    (\rnnMultiConcatTree^a \spreadInputOp{} x)[2^{k-1}-2] 
		  \bigr)\bigr)[2^{k-1}-2],
		\quad \text{for all } k\geq3,  \; \text{and all }  x\in[-D, D]^{\rnnSizeFunctionIn{\rnnOpp^1}},
        \end{equation}
	which, upon using \eqref{eq:multiconcat_rnn_output} for $\rnnMultiConcatTree^b$ and $\rnnMultiConcatTree^a$, yields
    \begin{align}\label{beq:3}
		( \rnnMultiConcatTree^{L+1} \spreadInputOp{}x)[2^k - 2] =
		  & \left( \multiconcatRNNmap{2^{L+1}}{k-(L+1)} \circ \dots \circ \multiconcatRNNmap{2^L+1}{k-(L+1)}\right)
		\bigl( \left( \multiconcatRNNmap{2^L}{k-(L+1)} \circ \dots \circ \multiconcatRNNmap{1}{k-(L+1)}\right)(x) \bigr) \\
		= & \left(\multiconcatRNNmap{2^{L+1}}{k-(L+1)} \circ \dots \circ \multiconcatRNNmap{1}{k-(L+1)}\right) (x),
		\,\, \text{for all } k\geq L + 3,  \; \text{and all }  x\in[-D, D]^{\rnnSizeFunctionIn{\rnnOpp^1}}.
	\end{align}
	This proves \eqref{eq:multiconcat_rnn_output} for $\rnnMultiConcatTree^{L+1}$.

	Next, we establish \eqref{eq:multiconcat_hid_outputs} for $\rnnMultiConcatTree^{L+1}$ and first treat the case $\ell \in \{1, \dots, 2^L\}$. To this end arbitrarily fix $\ell \in \{1, \dots, 2^L\}$ and note that, using \eqref{eq:multiconcat_hid_outputs} for $\rnnMultiConcatTree^a$, which is possible by the induction assumption, we can conclude that there are $A_a^\ell, b_a^\ell$, and $\timeOffsetIdx{1}^a, \dots, \timeOffsetIdx{\ell}^a \in \{0, \dots, \ceil{\log(\ell)}\}$ such that
	\begin{equation}\label{teq:23}
		A_a^\ell\left(\hidOpMultiConcatTree^a \spreadInputOp x\right)[2^k - 2 ] + b_a^\ell =
		\left(
		\multiconcatRNNmap{\ell}{k-\timeOffsetIdx{\ell}^a} \circ \dots \circ \multiconcatRNNmap{1}{k-\timeOffsetIdx{1}^a}
		\right)(x),\qquad
		 \text{for all } k \geq L+2, \; \text{and all } x\in[-D, D]^{\rnnSizeFunctionIn{\rnnOpp^1}}.
	\end{equation}
	Now, we let
	\begin{align}
			A^\ell \coloneqq A^\ell_a \subMatIn, \quad b^\ell \coloneqq b^\ell_a, \quad \textrm{and} \quad \timeOffsetIdx{1}  \coloneqq \timeOffsetIdx{1}^a,\; \dots,\; \timeOffsetIdx{\ell} \coloneqq \timeOffsetIdx{\ell}^a,
	\end{align}
	with $\subMatIn = \subMatInMap(\rnnMultiConcatTree^b, \rnnMultiConcatTree^a)$, and compute
	\begin{align}
		A^\ell\left(\hidOpMultiConcatTree{}^{L+1} \spreadInputOp x\right)[2^k - 2 ] + b^\ell & =
		A^\ell_a \subMatIn \left(\hidOpMultiConcatTree{}^{L+1} \spreadInputOp x\right)[2^k - 2 ] + b_a^\ell      \\
		\overset{\eqref{eq:prop_subMatIn}}&{=} A_a^{\ell}  \left(\hidOpMultiConcatTree{}^a \spreadInputOp x\right)[2^k - 2 ] + b_a^{\ell} \\
		\overset{\eqref{teq:23}} & {=}
		\left(
		\multiconcatRNNmap{\ell}{k-\timeOffsetIdx{\ell}^a} \circ \dots \circ \multiconcatRNNmap{1}{k-\timeOffsetIdx{1}^a}
		\right)(x) \\
		&=
		\left(
		\multiconcatRNNmap{\ell}{k-\timeOffsetIdx{\ell}} \circ \dots \circ \multiconcatRNNmap{1}{k-\timeOffsetIdx{1}}
		\right)(x)\iter{,}
	\end{align}
    for all $k \geq L + 2$ and all $x \in [-D, D]^{\rnnSizeFunctionIn{\rnnOpp^1}}$. In particular, the same identity holds for all $k \geq L + 3$, 
    so \eqref{eq:multiconcat_hid_outputs} is established for $\ell \in \{1, \dots, 2^L\}$. It remains to treat the case $\ell \in\{ 2^L+1, \dots, 2^{L+1}\}$. To this end, we arbitrarily fix $\ell \in\{ 2^L+1, \dots, 2^{L+1}\}$ and set $\ell'= \ell - 2^L \in \{1, \dots, 2^L\}$. Now, upon application of
    \eqref{eq:multiconcat_hid_outputs} to $\rnnMultiConcatTree^b$, we can conclude the existence of
	$A_b^{\ell'}, b_b^{\ell'}$, and
	\begin{equation}\label{teq:bound_second_idx_offsets}
		\timeOffsetIdx{1}^b, \dots, \timeOffsetIdx{\ell'}^b \in \{0, \dots, \ceil{\log(\ell')}\}
	\end{equation}
	such that
	\begin{equation}\label{teq:24}
	A_b^{\ell'}\left(\hidOpMultiConcatTree^b \spreadInputOp x\right)[2^k - 2 ] + b_b^{\ell'} =
		\left(
		\multiconcatRNNmap{2^L+\ell'}{k-\timeOffsetIdx{\ell'}^{\iter{b}}} \circ \dots \circ \multiconcatRNNmap{2^L+1}{k-\timeOffsetIdx{1}^{\iter{b}}}
		\right)(x), 
	\end{equation}
    for all $k \geq L+2$, and all $x\in[-D_{2^L}, D_{2^L}]^{\rnnSizeFunctionIn{\rnnOpp^{2^L+1}}}$.
	Next, with $\subMatOut = \subMatOutMap(\rnnMultiConcatTree^b, \rnnMultiConcatTree^a)$, let 
	\begin{align} 
		A^\ell \coloneqq A^{\ell'}_b \subMatOut, & \quad b^\ell
		\coloneqq b_b^{\ell'},
		 \quad \textrm{and}\\
		\timeOffsetIdx{1}\coloneqq L+1,\; \dots,\; \timeOffsetIdx{2^L}\coloneqq L+1, \; & \,\timeOffsetIdx{2^L+1 }\coloneqq \timeOffsetIdx{1}^b+1,\; \dots, \; \timeOffsetIdx{\iter{\ell}} \coloneqq \timeOffsetIdx{\ell'}^b+1, \label{teq:45}
	\end{align}
	and note that by \eqref{teq:bound_second_idx_offsets}, $\timeOffsetIdx{1}, \dots, \timeOffsetIdx{\ell} \in \{0, \dots, L+1\}$ because ${\ell'} = \ell - 2^L \leq 2^L$. As $\ell \in\{ 2^L+1, \dots, 2^{L+1}\}$, we have $\ceil{\log(\ell)} = L+1$ and can thus equivalently write $\timeOffsetIdx{1}, \dots, \timeOffsetIdx{\ell} \in \{0, \dots, \ceil{\log(\ell)}\}$.
	Next, we compute
	\begin{align}
   		\begin{split}
			A^\ell\left(\hidOpMultiConcatTree{}^{L+1} \spreadInputOp x\right)[2^k - 2 ] + b^\ell & =
			A^{{\ell'}}_b \subMatOut \left(\hidOpMultiConcatTree{}^{L+1} \spreadInputOp x\right)[2^k - 2 ] + b^{{\ell'}}_b \\
			\overset{\eqref{eq:prop_subMatOut}}&{=} A^{{\ell'}}_b  \left(\hidOpMultiConcatTree{}^b \spreadInputOp \left(\rnnMultiConcatTree{}^a \spreadInputOp x\right)[2^{k-1} - 2 ]\right)[2^{k-1}-2] + b^{{\ell'}}_b                                                            \\
			\overset{\eqref{teq:24}}                                                             & {=}  \left( \multiconcatRNNmap{\ell}{k-1-\timeOffsetIdx{{\ell'}}^b} \circ \dots \circ \multiconcatRNNmap{2^L+1}{k-1-\timeOffsetIdx{1}^b}  \right)\left(\left(\rnnMultiConcatTree{}^a \spreadInputOp x\right)[2^{k-1} - 2 ]\right) \\
			\overset{{\eqref{eq:multiconcat_rnn_output}\textrm{ for }\rnnMultiConcatTree{}^a}}   & {=}
			\left( \multiconcatRNNmap{\ell}{k-1-\timeOffsetIdx{{\ell'}}^b} \circ \dots \circ \multiconcatRNNmap{2^L+1}{k-1-\timeOffsetIdx{1}^b}  \right)\left( \left( \multiconcatRNNmap{2^L}{k-1-L} \circ \dots \circ \multiconcatRNNmap{1}{k-1-L}\right)(x) \right),\\
			\overset{\eqref{teq:45}}&{=}
		\left(\multiconcatRNNmap{\ell}{k-\timeOffsetIdx{\ell}} \circ \dots \circ \multiconcatRNNmap{1}{k-\timeOffsetIdx{1}}\right)(x)\iter{,}
		\end{split}\label{teq:hid_outputs_1}
	\end{align}
	for all $k \in \N$ such that $k-1 \geq L + 2$, i.e., $k \geq L + 3$, and all $x \in [-D, D]^{\rnnSizeFunctionIn{\rnnOpp^1}}$. This establishes \eqref{eq:multiconcat_hid_outputs} for the case $\ell \in\{ 2^L+1, \dots, 2^{L+1}\}$ as well.

	The proof is concluded by establishing \eqref{eq:multiconcat_sizes} as follows
	\begin{align}
		\rnnSizeFunctionHid{\rnnMultiConcatTree^{L+1}} & =
		\rnnSizeFunctionHid{\rnnMultiConcatTree{}^a} + 2 \rnnSizeFunctionOut{\rnnMultiConcatTree{}^a} + \rnnSizeFunctionHid{\rnnMultiConcatTree{}^b} + 5                                                                   \\
		& =\left(\sum_{\ell=1}^{2^L} \rnnSizeFunctionHid{\rnnOpp^\ell} + 2 \sum_{\ell=1}^{2^L-1} \rnnSizeFunctionOut{\rnnOpp^\ell} + 5 (2^L-1)\right)                       \\
	    & +2 \rnnSizeFunctionOut{\rnnOpp{}^{2^L}} + 5                                                                                                                \\
	    & + \left(\sum_{\ell=1}^{2^{L}} \rnnSizeFunctionHid{\rnnOpp^{2^L+\ell}} + 2 \sum_{\ell=1}^{2^L-1} \rnnSizeFunctionOut{\rnnOpp^{2^L+\ell}} + 5 (2^L-1)\right) \\
	    & =\sum_{\ell=1}^{2^{L+1}} \rnnSizeFunctionHid{\rnnOpp^\ell} + 2 \sum_{\ell=1}^{2^{L+1}-1} \rnnSizeFunctionOut{\rnnOpp^\ell} + 5 \left((2^L-1)+ (2^L-1)+1\right).
	\end{align}
	This finalizes the induction step going from $\rnnMultiConcatTree^L$ to $\rnnMultiConcatTree^{L+1}$ and thereby completes the overall proof.
\end{proof}

We finally extend Lemma~\ref{lem:multiconcat_rnn_tree} to the concatenation of an arbitrary number---as opposed to a power of two---RNNs. This will be effected by suitably inserting dummy networks and is formalized as follows.
	{
		\newcommand{\logLayer}{L'}

		\begin{corollary}\label{cor:multiconcat_rnn}
			Fix $D \geq 1 $, $L \in \N$, and let $\rnnOp{}^1, \dots, \rnnOpp{}^{L}$ be RNNs with $\rnnSizeFunctionIn{\rnnOpp^{\ell+1}}=\rnnSizeFunctionOut{\rnnOpp^{\ell}}$. Furthermore, assume that there are constants $D_1, \dots, D_{L}$ and $\hidBound{}\geq \max\{2, \max_{\ell\in\{1,\dots,L\}} D_\ell\}$  such that,
			for all $\ell\in\{1, \dots, L\}$,
			\begin{align}
            \sup_{\inorm{x} \leq D_{\ell-1}, t\in\No } \inorm{(\rnnOpp^{\ell} \spreadInputOp{} x)[t]} & \leq D_{\ell},  \label{teq:cor-assumption_concat3}\\ 
				\sup_{\inorm{x} \leq D_{\ell-1}, t\in\No } \inorm{(\hidOp^{\ell} \spreadInputOp{} x)[t]}  & \leq \hidBound{},   \label{teq:cor-assumption_concat4} 
			\end{align}
			where we set $D_0 = D$.
			Then, there exists a hidden state operator (Definition \ref{def:elman_rnn}) $\hidOpMultiConcatTree{}$
			with 
			\begin{equation}\label{eq:cor-multiconcat_sizes}
				\rnnSizeFunctionHid{\hidOpMultiConcatTree{}} \leq \sum_{\ell=1}^{L} \rnnSizeFunctionHid{\rnnOpp^\ell} + 2 \sum_{\ell=1}^{L} \rnnSizeFunctionOut{\rnnOpp^\ell} + 13L,
			\end{equation}
			such that, for every $\ell \in \{1, \dots, L\}$, there are $A^\ell, b^\ell$, and $\timeOffsetIdx{1}, \dots, \timeOffsetIdx{\ell} \in \{1, \dots,  \ceil{\log(\ell)}\}$, so that
			\begin{equation}\label{eq:cor-multiconcat_hid_outputs}
				A^\ell\left(\hidOpMultiConcatTree{} \spreadInputOp x\right)[2^k - 2 ] + b^\ell =
				\left(
				\multiconcatRNNmap{\ell}{k-\timeOffsetIdx{\ell}} \circ \dots \circ \multiconcatRNNmap{1}{k-\timeOffsetIdx{1}}
				\right)(x),\, \text{for all }
				k \geq \ceil{\log(L)}+2, \;  \text{and all } x\in[-D, D]^{\rnnSizeFunctionIn{\rnnOpp^1}},
			\end{equation}
			with $\multiconcatRNNmap{\ell}{k}$ as defined in \eqref{eq:def_multiconcat_maps}.
		\end{corollary}
		\begin{proof}
			With the goal of applying Lemma~\ref{lem:multiconcat_rnn_tree}, we complete the specified collection $\rnnOp{}^1, \dots, \rnnOpp{}^{L}$ of RNNs by the dummy RNN $\rnnZero_d = \outZero_d \hidZero_d$ for input dimension $d \in \N$ with weights
			\begin{align}\label{eq:zero_net}
				A_h = \begin{pmatrix}
					      0 \\
				      \end{pmatrix}
				\qquad
				A_x =
				\begin{pmatrix}
					0 \Ivec{d}^T
				\end{pmatrix}
				\qquad
				b_h =
				\begin{pmatrix} 0 \end{pmatrix}
				\qquad
				A_o =
				\begin{pmatrix}
					0
				\end{pmatrix}
				\qquad b_o = \begin{pmatrix} 0 \end{pmatrix}.
			\end{align}
			Since $(\rnnZero_d \spreadInputOp x)[t] = 0$ and $(\hidZero_d \spreadInputOp x)[t] = 0$, for all $x\in \R^d$, 
            the conditions \eqref{teq:assumption_concat3} and \eqref{teq:assumption_concat4} in Lemma~\ref{lem:multiconcat_rnn_tree} are trivially satisfied. 
            Next, we let $\logLayer \coloneqq \ceil{\log(L)}$, invoke Lemma \ref{lem:multiconcat_rnn_tree} for 
			\[
				\rnnOp{}^1, \dots, \rnnOpp{}^{L}, \underbrace{\rnnZero_{\rnnSizeFunctionOut{\rnnOpp^L}}, \rnnZero_{1}, \dots,  \rnnZero_{1}}_{2^{\logLayer}-L \textrm{ dummy networks}},
			\]
			and denote the resulting network by $\rnnMultiConcatTree = \outOpMultiConcatTree{} \hidOpMultiConcatTree{}$. Now $\hidOpMultiConcatTree{}$ is the desired hidden state operator since, by \eqref{eq:multiconcat_hid_outputs}, for every $\ell \in \{1, \dots, 2^{\logLayer}\}$, there are $A^\ell, b^\ell$, and $\timeOffsetIdx{1}, \dots, \timeOffsetIdx{\ell} \in \{0, \dots, \ceil{\log(\ell)}\}$ such that
			\begin{equation}
				A^\ell\left(\hidOpMultiConcatTree{} \spreadInputOp x\right)[2^k - 2 ] + b^\ell =
				\left(
				\multiconcatRNNmap{\ell}{k-\timeOffsetIdx{\ell}} \circ \dots \circ \multiconcatRNNmap{1}{k-\timeOffsetIdx{1}}
				\right)(x),\quad \text{for all }
				 k \geq \logLayer+2, \;  \text{and all } x\in[-D, D]^{\rnnSizeFunctionIn{\rnnOpp^1}}.
			\end{equation}
			Restricting to $\ell \in \{1, \dots, L\}$ yields \eqref{eq:cor-multiconcat_hid_outputs}.

It remains to establish that the operator $\hidOpMultiConcatTree{}$ we have identified satisfies \eqref{eq:cor-multiconcat_sizes}.
			When $L=2^{L'}$ the statement follows straight from Lemma \ref{lem:multiconcat_rnn_tree}.
            Else, we have
        \begin{align}
            \rnnSizeFunctionHid{\hidOpMultiConcatTree{}} \overset{\eqref{eq:multiconcat_sizes}} & {=}
            \sum_{\ell=1}^{L} \rnnSizeFunctionHid{\rnnOpp^\ell} + (2^{\logLayer}-L) \rnnSizeFunctionHid{\rnnZero}
            + 2 \left( \sum_{\ell=1}^{L} \rnnSizeFunctionOut{\rnnOpp^\ell} + (2^{\logLayer}-L -1) \rnnSizeFunctionOut{\rnnZero} \right)   \\
            & + 5 (2^{\logLayer}-1)     \\
            \overset{(i)} & {\leq}
            \sum_{\ell=1}^{L} \rnnSizeFunctionHid{\rnnOpp^\ell} + L + 2 \left( \sum_{\ell=1}^{L} \rnnSizeFunctionOut{\rnnOpp^\ell} + (L-1)  \right)  + 5 (2L-1)  \\
            & \leq \sum_{\ell=1}^{L} \rnnSizeFunctionHid{\rnnOpp^\ell}  + 2  \sum_{\ell=1}^{L} \rnnSizeFunctionOut{\rnnOpp^\ell}  + 13L, 
        \end{align}
			where in (i) we used $2^{\logLayer} \leq 2L$.
		\end{proof}
	}

\section{Approximation of monomials}

This section develops the hierarchical constructions used to approximate higher-order
monomials.  In deep feed-forward ReLU networks, such hierarchies are formed by
successive applications of squaring and multiplication.  In our RNN
framework, temporal depth plays the same role: higher powers are obtained through
repeated use of the clocked concatenation mechanism, combined with parallelization and
affine transformations as needed.  This temporal realization of compositional
structure is key to constructing RNNs that efficiently approximate the vector
$(x, x^2, \dots, x^N)$.

\begin{figure}[h]
	\centering
	\includegraphics[width=0.99\textwidth]{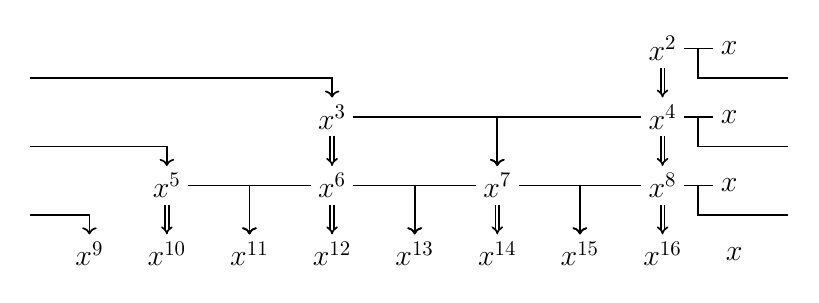}
	\caption{Expressing monomials of degree up to $N=16$ in terms of iterated squaring ($\Rightarrow$) and multiplication ($\rightarrow$).}
	\label{fig:pyramid}
\end{figure}

Our goal is to build a single RNN that, when run for sufficiently many time steps, simultaneously produces accurate approximations of $x^2, x^3, \ldots, x^N$, with $N=2^L,\, L \in \N$. To this end, we combine the previously established RNNs for squaring (Theorem~\ref{thm:square_net}) and multiplication (Theorem~\ref{thm:multiplication_net}) through the concatenation procedure formalized in Corollary~\ref{cor:multiconcat_rnn}. The resulting hierarchical construction, illustrated in Figure~\ref{fig:pyramid}, organizes the powers in a pyramid-like structure, enabling efficient realization of all monomials up to degree $N=2^L$ using $L=\log(N)$ successive concatenations. The restriction to $N=2^{L}$ poses no difficulty, since unused monomials will later
be discarded by assigning them zero coefficients. The $\ell$-th row in Figure \ref{fig:pyramid} corresponds to the application of the  function $\polyMap{\ell}$ as defined next.

\begin{definition}\label{def:mappings-fl}
	We define the mappings $\polyMap{1}: \R \to \R^2$ and, for $\ell \in \N$, with $\ell \geq 2$, $\polyMap{\ell}: \R^{2^{\ell-2}+1} \to \R^{2^{\ell-1}+1}$, as follows
	\begin{equation}\label{eq:def_polymap}
        \begin{array}{rl}        
		\polyMap{1}: x    & \to \begin{pmatrix}
			                        x^2 & x
		                        \end{pmatrix}^T \\
		\polyMap{\ell}(x)_i & = \begin{cases}
			x_{2^{\ell-2}} x_{2^{\ell-2}+1}, & \quad \textrm{ if } i=1                                                         \\
			x_{k}^2,                         & \quad \textrm{ if }\, i=2k,  \textrm{ with } \, k \in \{1, \dots, 2^{\ell-2}\}     \\
			x_{k} x_{k+1},                  & \quad \textrm{ if }\, i=2k+1, \textrm{ with } \, k \in \{1, \dots, 2^{\ell-2}-1\} \\
			x_{2^{\ell-2}+1},               & \quad \textrm{ if }\, i=2^{\ell-1}+1.
		\end{cases}
    \end{array}
    \end{equation}
\end{definition}

The functions $f_\ell$ just defined are designed to generate, through recursive application, all monomials of increasing degree. The next result makes this relationship explicit by characterizing the exact form of the compositions according to $(\polyMap{\ell}\circ \dots \circ \polyMap{1})(x) =
\begin{pmatrix}
		x^{2^{\ell-1}+1} & \dots & x^{2^{\ell}} & x
	\end{pmatrix}^T. 
    $
In particular, it shows that each composition step doubles the range of available powers of $x$ while preserving the input variable itself.

\begin{lemma}\label{lm:polyMap}
	Let $\ell\in\N$ and define $\polyMapConcat{\ell} \coloneqq \polyMap{\ell}\circ \dots \circ \polyMap{1}$. For all $x \in \R$, it holds that
	\begin{align}
		\begin{split}\label{teq:polyMap_induct_base}
			& \polyMapConcat{\ell}(x)_i               =  x^{2^{\ell-1}+i}, \qquad \textrm{ for }\, i\in\{1, \dots, 2^{\ell-1}\}, \quad \textrm{and} \\
			& \polyMapConcat{\ell}(x)_{2^{\ell-1}+1}  =  x.
		\end{split}
	\end{align}
\end{lemma}
\begin{proof}
	Arbitrarily fix $x\in\R$. The proof proceeds by induction. To establish the base case, note that, for $\ell=1$, we have 
	\begin{align}
		\polyMap{1}(x)_1         & = x^2 = x^{2^0 + 1}       \\
		\polyMap{1}(x)_{2^0 + 1} & = \polyMap{1}(x)_{2} = x.
	\end{align}
	For the induction step, assume that \eqref{teq:polyMap_induct_base} holds for some $\ell \in \N$ and set $z \coloneqq (\polyMap{\ell}\circ \dots \circ \polyMap{1})(x) \in \R^{2^{\ell-1} + 1}$. By the induction hypothesis
	\begin{align}
		z_i               = x^{2^{\ell-1}+i}, \qquad \textrm{ for } i\in\{1, \dots, 2^{\ell-1}\}, \quad \textrm{and} \quad z_{2^{\ell-1}+1}  = x.
	\end{align}
	Using \eqref{eq:def_polymap}, we now compute
	\begin{align}
		\left(\polyMap{\ell+1}(z)\right)_1 = z_{2^{\ell-1}} \cdot z_{2^{\ell-1} +1} = x^{2^{\ell-1} + 2^{\ell-1} }\cdot x = x^{2^{\ell} + 1}.
	\end{align}
	For indices of the form $i=2k$, with $k \in \{1, \dots, 2^{\ell-1}\}$, we obtain
	\begin{align}
		\left(\polyMap{\ell+1}(z)\right)_i = (z_k)^2 = \left(x^{2^{\ell-1} + k }\right)^2 = x^{2^{\ell} + i}.
	\end{align}
	Similarly, for $i=2k+1$, with $k \in \{1, \dots, 2^{\ell-1}-1\}$,
	\begin{align}
		\left(\polyMap{\ell+1}(z)\right)_i = z_k \cdot z_{k+1} = x^{2^{\ell-1} + k } x^{2^{\ell-1} + k+1 } = x^{2^{\ell} + i}.
	\end{align}
	Finally,
	\begin{align}
		\left(\polyMap{\ell+1}(z)\right)_{2^{\ell}+1} = z_{2^{\ell-1}+1} = x.
	\end{align}
	Hence, \eqref{teq:polyMap_induct_base} holds for $\ell+1$ as well, completing the induction.
\end{proof}

The following bound, quantifying how the range of the composed maps grows with $\ell$, is an immediate consequence of Lemma~\ref{lm:polyMap}.

\begin{corollary}\label{cor:polyMap_outbound}
	Let $\ell\in\N,D \geq 1$. For all $x \in[-D, D]$, it holds that
	\[
		\inorm{\polyMapConcat{\ell}(x)} \leq D^{2^\ell}.
	\]
\end{corollary}
In addition, we will make use of the following properties of the maps $\polyMap{\ell}$ introduced in Definition~\ref{def:mappings-fl}. 

\begin{lemma}\label{lm:polyMap_lips}
	Let $D \geq 1$. For all $x,y\in [-D, D]$, it holds that
    \begin{align}
        \inorm{\polyMap{1}(x)} \leq D^2
		\qquad \textrm{and}
		\qquad
		\inorm{\polyMap{1}(x) - \polyMap{1}(y)} & \leq 2D |x-y|.
    \end{align}
    For all $\ell \geq 2$ and all $x,y\in [-D^{2^{\ell-1}}, D^{2^{\ell-1}}]^{ 2^{\ell-2} +1}$,
	\begin{align}
		\inorm{\polyMap{\ell}(x)} \leq D^{2^{\ell}}
		\qquad \textrm{and}
		\qquad
		\inorm{\polyMap{\ell}(x) - \polyMap{\ell}(y)} & \leq 2D^{2^{\ell-1}} \inorm{x-y}.
	\end{align}
\end{lemma}
\begin{proof}
	We first consider the case $\ell = 1$. For $x \in [-D,D]$, we have $|x| \leq D \leq D^2$ and $x^2 \leq D^2$, hence $\inorm{\polyMap{1}(x)} \leq D^2$. Moreover, for all $x,y \in [-D,D]$, $|x^2 - y^2| \leq |x||x-y| + |y||x-y| \leq 2D |x-y|$, which implies $\inorm{\polyMap{1}(x) - \polyMap{1}(y)}  \leq 2D |x-y|$. We now turn to the case $\ell \geq 2$.
	From \eqref{eq:def_polymap}, each coordinate of $\polyMap{\ell}(x)$ is either $x_{2^{\ell-2}}x_{2^{\ell-2}+1}$ (the first one), a product  
    $x_i x_j$, with $i, j \in \{1, \dots, {2^{\ell-2}}\}$ (possibly $i=j$), or $x_{2^{\ell-2}+1}$ corresponding to the last component. For the latter the claim is immediate. 
	Now, arbitrarily fix $x,y\in [-D^{2^{\ell-1}}, D^{2^{\ell-1}}]^{2^{\ell-2}+1}$ and $i,j\in\{1, \dots, 2^{\ell-2}+1\}$.
	We have
	\[
		|x_i x_j| = |x_i||x_j| \leq D^{2^{\ell-1}} D^{2^{\ell-1}} = D^{2^{\ell}}
	\]
	and
	\[
		|x_i x_j - y_i y_j| \leq  |x_i x_j -x_i y_j| + | x_i y_j- y_i y_j| = |x_i||x_j - y_j| + |y_j||x_i - y_i| \leq 2D^{2^{\ell-1}} \inorm{x-y}.
	\]
	As $x$, $y$ and $i,j$ were arbitrary, the statement follows.
\end{proof}

Next, we construct RNNs that approximate the mappings $\polyMap{\ell}$. Besides the RNNs for squaring and multiplication, we also require one that realizes the identity map.

\begin{lemma}\label{lm:identity_net}
	There exists an RNN $\rnnId = \outId \hidId$ such that, for all $x\in\R$, and all $t\in\No{}$,
	\begin{align}
		(\rnnId \spreadInputOp x)[t] & = x, \quad \inorm{(\hidId \spreadInputOp x)[t] }  = |x|.
    	\end{align}
	Moreover, $\rnnSizeFunctionHid{\rnnId}=2$.
\end{lemma}
\begin{proof}
	Choose the weights as 
	\begin{align}\label{eq:identity_weights}
		A_h & = \begin{pmatrix}
			        1 & 0 \\
			        0 & 1 \\
		        \end{pmatrix},
		\qquad
		A_x =
		\begin{pmatrix}
			1  \\
			-1 \\
		\end{pmatrix},
		\qquad
		b_h = 0, \qquad
		A_o =
		\begin{pmatrix}
			1 & -1
		\end{pmatrix},
		\qquad b_o = 0.
	\end{align}
	The claim follows by direct verification based on Definition~\ref{def:elman_rnn}, using $x = \relu(x) - \relu(-x)$.
\end{proof}
We start by building an RNN that approximates $\polyMap{1}$.
\begin{lemma}\label{lem:square_and_identity}
	Let $D \geq 1$. There exists an RNN $\rnnPoly{1}{D}$ with $\rnnSizeFunctionOut{\rnnPoly{1}{D}}=2$ such that, for all $x \in [-D, D]$, and all $t \in \No$,
	\begin{equation}
		\left|((\rnnPoly{1}{D} \spreadInputOp x)[t])_1 - x^2 \right|  \leq \frac{D^2}{2}4^{-t},     \quad 
		((\rnnPoly{1}{D}\spreadInputOp x)[t])_2  = x,   
	\end{equation}
	and
	\begin{equation}
		\inorm{ (\hidPoly{1}{D} \spreadInputOp x)[t]} \leq D, \quad \inorm{ (\rnnPoly{1}{D} \spreadInputOp x)[t]} \leq D^2. 
	\end{equation}
	Furthermore, $\rnnSizeFunctionHid{ \rnnPoly{1}{D}} = 9$.
\end{lemma}
\begin{proof}
	Combine the RNN approximating $x^2$ from Theorem \ref{thm:square_net} with the RNN realizing the identity operator from Lemma \ref{lm:identity_net}. 
    By Lemma \ref{lm:parallel_net}, these two networks can be run in parallel, yielding an RNN that exhibits the desired behavior for inputs $\begin{pmatrix} x \\ x \end{pmatrix}$. Finally, noting that $\begin{pmatrix} x \\ x \end{pmatrix} = A x$ with $A \coloneqq \begin{pmatrix} 1 \\ 1 \end{pmatrix}$, we apply Lemma \ref{lm:linear_map_rnn} to incorporate this affine transformation into the input layer.
    This yields $\rnnPoly{1}{D}$ with the desired behavior.
\end{proof}

Next, we construct RNNs that approximate $\polyMap{\ell}$, for arbitrary $\ell\geq 2$.
\begin{lemma}\label{lm:rnn_approx_polymap}
	Let $D \geq 1$. For every $\ell \geq 2$, there exists an RNN $\rnnPoly{\ell}{D}=\outPoly{\ell}{D}\hidPoly{\ell}{D}$ such that, for all $x \in \left[-D^{2^{\ell-1}}, D^{2^{\ell-1}}\right]^{2^{\ell-2}+1}$, and all $t\in\No$, 
	\[
		\inorm{(\rnnPoly{\ell}{D} \spreadInputOp x)[t] - \polyMap{\ell}(x)} \leq \frac{D^{2^{\ell}}}{2} \phantom{\cdot} 4^{-t},
		\quad  \inorm{(\rnnPoly{\ell}{D} \spreadInputOp x)[t]} \leq D^{2^{\ell}},
		\quad  \inorm{(\hidPoly{\ell}{D} \spreadInputOp x)[t]} \leq D.
	\]
	Moreover, $\rnnSizeFunctionHid{ \rnnPoly{\ell}{D}} \leq 10\cdot2^{\ell}$.
\end{lemma}

\begin{proof}
	Arbitrarily fix $\ell\in\N$, $\ell \geq 2,$ and define the selector matrix $A \in \mathbb R^{(3\cdot2^{\ell-2}+1)\times(2^{\ell-2}+1)}$
that rearranges and duplicates the coordinates of 
$x\in\mathbb R^{2^{\ell-2}+1}$ according to the sequence of inputs used in~\eqref{eq:def_polymap}. 
Explicitly,

\newcommand{\undertexflex}[5]{%
  \begin{tikzpicture}[baseline=(C.base)]
    \node[inner sep=0pt, outer sep=0pt] (C) {$#1$};

    \node[anchor=north west, inner sep=0pt, outer sep=0pt] (B)
      at ($(C.south west)+(0,-#3)$) {$\underbrace{\phantom{#1}}{}$};

    \node[anchor=north, inner sep=0pt, outer sep=0pt]
      at ($(B.south)+(#5,-#4)$) {$\small \text{#2}$};
  \end{tikzpicture}%
}

\begin{equation}
Ax =
\Big(
\undertexflex{x_{2^{\ell-2}},\,x_{2^{\ell-2}+1}}{special product}{-7pt}{2pt}{-1.5pt},
\undertexflex{x_1,\dots,x_{2^{\ell-2}}}{squares}{-5.33pt}{4.5pt}{0pt},
\undertexflex{x_{1},x_{2},\,x_{2},x_{3},\,\dots,\,x_{2^{\ell-2}-1},x_{2^{\ell-2}}}{adjacent products}{-7pt}{2pt}{0pt},
\undertexflex{x_{2^{\ell-2}+1}}{identity}{-7pt}{2pt}{0pt}
\Big)^{\!T}.
\label{eq:a-def}
\end{equation}

\noindent Equivalently, the rows of $A$ are arranged as follows:
\begin{enumerate}
\item the first two rows select $x_{2^{\ell-2}}$ and $x_{2^{\ell-2}+1}$;
\item the next $2^{\ell-2}$ rows select $x_1,\dots,x_{2^{\ell-2}}$;
\item for each $k\in\{1,\dots,2^{\ell-2}-1\}$, two successive rows select $x_k$ and $x_{k+1}$;
\item the final row selects $x_{2^{\ell-2}+1}$.
\end{enumerate}
Each row of $A$ contains exactly one nonzero entry equal to~$1$.
For
$\ell=2$, for example, we have $x=(x_1,x_2)$ and 
$$
A =
\begin{pmatrix}
1 & 0\\
0 & 1\\
1 & 0\\
0 & 1
\end{pmatrix}.
$$

	Next, let $\rnnMult$ and $\rnnSquare$ denote the multiplication and squaring RNNs from Theorem \ref{thm:multiplication_net} and Theorem \ref{thm:square_net}, respectively, each instantiated with input bound $D^{2^{\ell-1}}$. Since $(D^{2^{\ell-1}})^2 = D^{2^{\ell}}$, it follows that,
    for all $x, x_1, x_2 \in [-D^{2^{\ell-1}},D^{2^\ell-1}]$, and all $t\in\No$,
	\begin{align}
		\left|\left(\rnnMult{}\spreadInputOp{} \! \begin{pmatrix} x_1 \\ x_2 \end{pmatrix}\right)[t] - (x_1 \cdot x_2) \right| & \leq \frac{D^{2^{\ell}}}{2} \phantom{\cdot} 4^{-t}, \,
		\inorm{\left(\hidMult \spreadInputOp{} \! \begin{pmatrix} x_1 \\ x_2 \end{pmatrix}\right)[t]} \leq 1 \leq D,
		                                                                                                                   \ \,  \left|\left(\rnnMult{} \spreadInputOp{} \!\! \begin{pmatrix} x_1 \\ x_2 \end{pmatrix}\right)[t]\right|  \leq D^{2^{\ell}}, \\
		\left|\left(\rnnSquare{}\spreadInputOp x\right)[t] - x^2 \right|                                                & \leq \frac{D^{2^{\ell}}}{2} \phantom{\cdot} 4^{-t}, \,
		\inorm{\left(\hidSquare \spreadInputOp{} x \right)[t]} \leq 1 \leq D, 
		                                                                                                                   \ \, \left|\left(\rnnSquare{} \spreadInputOp{} x \right)[t]\right|   \leq D^{2^{\ell}}.
	\end{align} 
	To assemble the network approximating $f_\ell$, we use Lemma \ref{lm:parallel_net} to construct the RNN $\rnnTmp$ that runs
    the RNNs
	\begin{equation}
		\rnnMult, \rnnSquare, \rnnMult, \rnnSquare, \dots, \rnnMult, \rnnSquare, \rnnId
	\end{equation}
	in parallel, with $2^{\ell-2}$ copies of both $\rnnMult$ and $\rnnSquare$. The desired RNN $\rnnPoly{\ell}{D}$ is then obtained by applying
    Lemma \ref{lm:linear_map_rnn} to $\rnnTmp$, with $A$ as defined in \eqref{eq:a-def}. Its hidden state size satisfies
    	\[
		\rnnSizeFunctionHid{\rnnPoly{\ell}{D}} = 2^{\ell-2} \left(\rnnSizeFunctionHid{\rnnSquare} + \rnnSizeFunctionHid{\rnnMult} \right) + \rnnSizeFunctionHid{\rnnId}
		= 2^{\ell-2} (7+14) + 2 = 2^{\ell} \frac{21}{4} + 2 \leq
		10 \cdot 2^{\ell}.
	\]
\end{proof}
The next step is to combine the RNNs approximating $\polyMap{\ell}$ into a single RNN that approximates $\polyMapConcat{L}$. Its hidden state operator
is defined as follows.
\begin{definition}\label{def:powers_hid_op}
	Let $L \in \N$ and $D \geq 1$. Invoke Corollary \ref{cor:multiconcat_rnn} with the RNNs $\rnnPoly{1}{D}, \dots, \rnnPoly{L}{D}$ constructed in Lemmata \ref{lem:square_and_identity} and \ref{lm:rnn_approx_polymap}, and denote the resulting hidden state operator by $\hidPowers{D}{L}$.
\end{definition}
The following lemma provides an explicit bound on the hidden-state dimension of $\hidPowers{D}{L}$.
\begin{lemma}\label{lem:powers_size}
	Let $L \in \N$ and $D \geq 1$. The hidden state dimension of $\hidPowers{D}{L}$ according to Definition~\ref{def:powers_hid_op} satisfies 
    $\rnnSizeFunctionHid{\hidPowers{D}{L}}  \leq 40 \cdot 2^L$. 
\end{lemma}
\begin{proof}
	\begin{align}
		\rnnSizeFunctionHid{\hidPowers{D}{L}} \overset{\eqref{eq:cor-multiconcat_sizes}} & {\leq}
		\sum_{\ell=1}^{L} \rnnSizeFunctionHid{\rnnPoly{\ell}{D}} + 2 \sum_{\ell=1}^{L} \rnnSizeFunctionOut{\rnnPoly{\ell}{D}} + 13L                                                 \\
		                                                                           & \leq \sum_{\ell=1}^{L} 10 \cdot 2^{\ell} + 2 \sum_{\ell=1}^{L} (2^{\ell-1} +1) + 13L \\
		                                                                           & = 20 \sum_{\ell=0}^{L-1} 2^{\ell} + 2 \sum_{\ell=0}^{L-1} 2^{\ell} +  15L          \\
		                                                                           & = 20 (2^L-1) + 2 (2^L - 1) +  15L                                                    \\
		                                                                           & \leq 22 \cdot 2^L +  15L                                                             \\
                                                                                  & \leq 40 \cdot 2^L.\qedhere
	\end{align}
\end{proof}

Using the definition of $\multiconcatRNNmap{\ell}{k}$ in \eqref{eq:def_multiconcat_maps}, we can now quantify how well the finite-time output of each network $\mathcal R_D^{\ell}$ approximates the corresponding target function $\polyMap{\ell}$. The following corollary provides uniform bounds on the approximation error and the output magnitude.

\begin{corollary}\label{cor:specific_rnns_for_poly_approx}
Let $L\in\mathbb N$ and $D\ge1$. 
For each $\ell\in\{1,\dots,L\}$ and every $k\ge2$, let
\begin{equation}\label{eq:def-glk}
    \multiconcatRNNmap{\ell}{k}(x) := (\mathcal R_D^{\ell}\mathcal D x)[2^k - 2]. 
\end{equation}
Then, for all $x\in[-D,D]$ if $\ell=1$ and all 
$x\in[-D^{2^{\ell-1}},D^{2^{\ell-1}}]^{\,2^{\ell-2}+1}$ if $\ell\ge2$, it holds that
\[
\|\multiconcatRNNmap{\ell}{k}(x) - \polyMap{\ell}(x)\|_\infty
\le 
8D^{2^{\ell}}\,4^{-2^k},
\qquad
\|\multiconcatRNNmap{\ell}{k}(x)\|_\infty \le D^{2^{\ell}}.
\]
\end{corollary}

\begin{proof}
Fix $L\in\mathbb N$, $D\ge1$, and $k\ge2$.

\medskip
\noindent\emph{Case $\ell=1$.}
By Lemma~\ref{lem:square_and_identity}, for all $x\in[-D,D]$ and all $t\in\mathbb N_0$,
\[
\big|((\mathcal R^{1}_{D}\mathcal D x)[t])_1 - x^2\big| \le \frac{D^2}{2}\,4^{-t},
\qquad
((\mathcal R^{1}_{D}\mathcal D x)[t])_2 = x,
\]
and hence
\[
\big\|(\mathcal R^{1}_{D}\mathcal D x)[t] - \polyMap{1}(x)\big\|_\infty
\le \frac{D^2}{2}\,4^{-t},
\qquad
\big\|(\mathcal R^{1}_{D}\mathcal D x)[t]\big\|_\infty \le D^2.
\]
Evaluating at $t=2^k-2$ and using $\multiconcatRNNmap{1}{k}(x)=(\mathcal R^{1}_{D}\mathcal D x)[2^k-2]$, yields
\[
\|\multiconcatRNNmap{1}{k}(x)-\polyMap{1}(x)\|_\infty
\le \frac{D^2}{2}\,4^{-(2^k-2)}
= 8D^2\,4^{-2^k},
\qquad
\|\multiconcatRNNmap{1}{k}(x)\|_\infty \le D^2,
\]
which matches the claimed bounds for $\ell=1$. 

\medskip
\noindent\emph{Case $\ell\ge2$.}
By Lemma~\ref{lm:rnn_approx_polymap}, for all admissible $x$ and all $t\in\mathbb N_0$,
\[
\big\|(\mathcal R_D^\ell\mathcal D x)[t] - \polyMap{\ell}(x)\big\|_\infty
\le \frac{D^{2^\ell}}{2}\,4^{-t},
\qquad
\big\|(\mathcal R_D^\ell\mathcal D x)[t]\big\|_\infty \le D^{2^\ell}.
\]
Setting $t=2^k-2$ and using $\multiconcatRNNmap{\ell}{k}(x)=(\mathcal R_D^\ell\mathcal D x)[2^k-2]$, this yields
\[
\|\multiconcatRNNmap{\ell}{k}(x) - \polyMap{\ell}(x)\|_\infty
\le \frac{D^{2^\ell}}{2}\,4^{-(2^k-2)}
= 8D^{2^\ell}\,4^{-2^k},
\qquad
\|\multiconcatRNNmap{\ell}{k}(x)\|_\infty \le D^{2^\ell}.
\]

\noindent The two cases together prove the corollary.
\end{proof}

We now demonstrate that each composite function $\polyMapConcat{\ell}$ can be approximated
by applying an affine output map to the hidden-state sequence generated by $\mathcal K^{\pi}_{D,L}$ (see Definition~\ref{def:powers_hid_op}).

\begin{lemma}\label{lem:hidpowers-approx}
Let $L \in \mathbb N$ and $D \ge 1$.
For each $\ell \in \{1,\dots,L\}$, there exist matrices $A^{\ell}$ and vectors $b^{\ell}$ 
such that, for all $x \in [-D,D]$, and all $k \ge \lceil \log(L) \rceil + 2$, 
$$
\big\|
A^{\ell}(\mathcal K^{\pi}_{D,L}\mathcal D x)[2^k - 2] + b^{\ell} - \polyMapConcat{\ell}(x)
\big\|_\infty
\le
8\,\cdot\,2^{\ell} D^{2^{\ell}} 4^{-\tfrac{1}{2\ell} 2^k}
=:\varepsilon_{\ell,k}.
$$
\end{lemma}


\begin{proof}
Let $L\in\mathbb N$ and $D\ge1$, and define $\multiconcatRNNmap{\ell}{k}$ as in~\eqref{eq:def_multiconcat_maps} 
for $\ell\in\{1,\dots,L\}$ and $k\ge2$. 
Fix $\ell\in\{1,\dots,L\}$ arbitrarily. 
By Corollary~\ref{cor:multiconcat_rnn}, there exist matrices $A^\ell$, vectors $b^\ell$, 
and integers $\tilde k_1,\dots,\tilde k_\ell\le\lceil\log\ell\rceil$ such that
\begin{equation}
A^{\ell}(\mathcal K^{\pi}_{D,L}\mathcal D x)[2^k-2] + b^{\ell}
= \big(\multiconcatRNNmap{\ell}{{\,k-\tilde k_\ell}}\circ\cdots\circ \multiconcatRNNmap{1}{{\,k-\tilde k_1}}\big)(x),
\,\, \text{for all }
k \ge \lceil\log(L)\rceil + 2,\; \text{and all }
x\in[-D,D]. \label{eq:AK-representation}
\end{equation}

We now verify that $A^\ell$ and $b^\ell$ satisfy the desired properties. 
To this end, define for each $\ell'\in\{1,\dots,\ell\}$ 
and $k\ge\lceil\log (L)\rceil+2$, the mappings
\[
G_{\ell'}^{k} := \multiconcatRNNmap{{\ell'}}{{\,k-\tilde k_{\ell'}}} \circ \cdots \circ \multiconcatRNNmap{1}{{\,k-\tilde k_1}}.
\]
Fix $x\in[-D,D]$ and $k\ge\lceil\log (L)\rceil+2$. 
We show by induction on $\ell'\in\{1,\dots,\ell\}$ that
\begin{equation}
\|G_{\ell'}^{k}(x)\|_\infty \le D^{2^{\ell'}}, \label{eq:Gnormbound}
\end{equation}
and
\begin{equation}
\|G_{\ell'}^{k}(x) - \polyMapConcat{\ell'}(x)\|_\infty
\le
8D^{2^{\ell'}}\,4^{-2^{k-\ceil{\log(\ell)}}}
\left(\sum_{i=0}^{\ell'-1}2^{i}\right). \label{eq:Gerrorbound}
\end{equation}
The base case, corresponding to $\ell' = 1$, follows directly from Corollary~\ref{cor:specific_rnns_for_poly_approx}. 
Indeed, since $G_1^{k} = \multiconcatRNNmap{1}{{\,k-\tilde k_1}}$, we have for all $x\in[-D,D]$,
\begin{eqnarray}
\|G_1^{k}(x)\|_\infty & = & \|\multiconcatRNNmap{1}{{\,k-\tilde k_1}}(x)\|_\infty \le D^{2},\\
\|G_1^{k}(x) - \polyMapConcat{1}(x)\|_\infty
 & = & \|\multiconcatRNNmap{1}{{\,k-\tilde k_1}}(x) - \polyMap{1}(x)\|_\infty
\le 8D^{2}\,4^{-2^{k-\tilde k_1}}
\le 8D^{2}\,4^{-2^{k-\ceil{\log(\ell)}}}.
\end{eqnarray}
Next, assume that~\eqref{eq:Gnormbound} and~\eqref{eq:Gerrorbound} hold for some 
$\ell' \in \{1,\dots,\ell-1\}$, and consider $\ell'+1$. 
We have
$$
\|G_{\ell'+1}^{k}(x)\|_\infty
= \|\multiconcatRNNmap{{\ell'+1}}{{\,k-\tilde k_{\ell'+1}}}(G_{\ell'}^{k}(x))\|_\infty
\le 
\sup_{\|z\|_\infty \le D^{2^{\ell'}}}
\|\multiconcatRNNmap{{\ell'+1}}{{\,k-\tilde k_{\ell'+1}}}(z)\|_\infty
\le D^{2^{\ell'+1}},
$$
where the last inequality again follows from Corollary~\ref{cor:specific_rnns_for_poly_approx}.
	This shows that \eqref{eq:Gnormbound} holds for $\ell' + 1$ as well. Next, we have
	\begin{align}
		\inorm{\multiconcatRNNmapG{\ell'+1}{k}(x) - \polyMapConcat{\ell'+1}(x) } & =
		\inorm{\multiconcatRNNmap{\ell'+1}{k-\timeOffsetIdx{\ell'+1}}(\multiconcatRNNmapG{\ell'}{k}(x)) - \polyMap{\ell'+1}(\polyMapConcat{\ell'}(x)) }  \\
		                                                                         & \leq \inorm{\multiconcatRNNmap{\ell'+1}{k-\timeOffsetIdx{\ell'+1}}(\multiconcatRNNmapG{\ell'}{k}(x)) - \polyMap{\ell'+1}(\multiconcatRNNmapG{\ell'}{k}(x)) } +
		\inorm{ \polyMap{\ell'+1}(\multiconcatRNNmapG{\ell'}{k}(x)) - \polyMap{\ell'+1}(\polyMapConcat{\ell'}(x)) }                                                                                                                               \\
		\overset{\eqref{eq:Gnormbound}}                                          & {\leq} \sup_{\inorm{z}\leq D^{2^{\ell'}}}\inorm{\multiconcatRNNmap{\ell'+1}{k-\timeOffsetIdx{\ell'+1}}(z) - \polyMap{\ell'+1}(z) } +
		\inorm{ \polyMap{\ell'+1}(\multiconcatRNNmapG{\ell'}{k}(x)) - \polyMap{\ell'+1}(\polyMapConcat{\ell'}(x)) }                                                                                                                              \\
		\overset{\textrm{Cor. } \ref{cor:specific_rnns_for_poly_approx}}         & {\leq}
		8D^{2^{\ell'+1}} 4^{-2^{k-\timeOffsetIdx{\ell'+1}}} +
		\inorm{ \polyMap{\ell'+1}(\multiconcatRNNmapG{\ell'}{k}(x)) - \polyMap{\ell'+1}(\polyMapConcat{\ell'}(x)) }                                                                                                                              \\
		\overset{\textrm{Lem. } \ref{lm:polyMap_lips}}                           & {\leq}
		8D^{2^{\ell'+1}} 4^{-2^{k-\timeOffsetIdx{\ell'+1}}} +
		2D^{2^{\ell'}}\inorm{\multiconcatRNNmapG{\ell'}{k}(x) -\polyMapConcat{\ell'}(x) }                                                                                                                                                        \\
		\overset{\eqref{eq:Gerrorbound}}                                          & {\leq}
		8D^{2^{\ell'+1}} 4^{-2^{k-\timeOffsetIdx{\ell'+1}}} +
		2D^{2^{\ell'}} \left(  8 \cdot D^{2^{\ell'}} 4^{-2^{k-\ceil{\log(\ell)}}} \left( \sum_{i=0}^{\ell'-1} 2^i\right) \right)                                                                                                                  \\
		\overset{(\ast)}& {\leq}
		8D^{2^{\ell'+1}} 4^{-2^{k-\ceil{\log(\ell)}}} +
		8 \cdot D^{2^{\ell'+1}} 4^{-2^{k-\ceil{\log(\ell)}}} \left( \sum_{i=0}^{\ell'-1} 2^{i+1}\right)                                                                                                                                         \\
		                                                                         & =
		8 D^{2^{\ell'+1}} 4^{-2^{k-\ceil{\log(\ell)}}} \left( \sum_{i=0}^{\ell'} 2^{i}\right), 
	\end{align}
	where in $(\ast)$ we used $\timeOffsetIdx{\ell'+1} \leq \ceil{ \log(\ell)}$. This proves that \eqref{eq:Gerrorbound} holds for $\ell' + 1$ as well. In particular, using \eqref{eq:Gerrorbound} with $\ell' = \ell$, we can thus compute
	\begin{align}
		\inorm{A^\ell\left(\hidPowers{D}{L} \spreadInputOp x\right)[2^k - 2 ] + b^\ell - \polyMapConcat{\ell}(x)}
		\overset{\eqref{eq:AK-representation}}       & {=}
		\inorm{\multiconcatRNNmapG{\ell}{k}(x) - \polyMapConcat{\ell}(x)}                                  \\
		\overset{\eqref{eq:Gerrorbound}} & {\leq}
		8 \cdot D^{2^{\ell}} 4^{-2^{k-\ceil{\log(\ell)}}} \left( \sum_{i=0}^{\ell-1} 2^{i}\right)           \\
		                                & = 8 \cdot D^{2^{\ell}} 4^{-2^{k-\ceil{\log(\ell)}}}  (2^\ell - 1) \\
		                                & \leq 8 \cdot 2^\ell \cdot D^{2^{\ell}} 4^{-2^{k-\log(\ell) - 1}}  \\
		                                &  = 8  \cdot 2^\ell \cdot D^{2^{\ell}} 4^{-\frac{1}{2\ell}2^{k}}.
	\end{align}
	Since $\ell\in\{1, \dots, L\}$, $k \geq \ceil{\log(L)}+2$, and $x\in[-D, D]$ were arbitrary, this completes the proof.
\end{proof}

Having constructed the hidden-state operator $\hidPowers{D}{L}$ in Definition~\ref{def:powers_hid_op} and established its approximation properties in Lemma~\ref{lem:hidpowers-approx}, we now specify the affine output map that extracts the desired polynomial terms from its hidden-state sequence.
\begin{lemma}\label{lem:power_output_mapping}
	Let $L \in \N$ and $D \geq 1$. There exist matrices $\Ao{\pi}$ and vectors $\bo{\pi}$, such that the affine 
    output mapping $\outPowers{D}{L}: h \to \Ao{\pi} h + \bo{\pi}$ satisfies, for all $x\in[-D, D]$ and all $k \geq \ceil{\log(L)}+2$, 
	$$
		((\outPowers{D}{L} \hidPowers{D}{L} \spreadInputOp x)[2^k-2])_1  = x,                                                                          
    $$
    and, for every $\ell\in\{1,\dots,L\}$ and $j\in\{1,\dots,2^{\ell-1}\}$,
    $$
		| ((\outPowers{D}{L} \hidPowers{D}{L} x)[2^k-2])_{2^{\ell-1} + j} - x^{2^{\ell-1}+j}| \leq \varepsilon_{\ell, k}.
	$$
    Here, $\varepsilon_{\ell,k}$ denotes the approximation error defined in 
Lemma~\ref{lem:hidpowers-approx}, and $\mathcal K^{\pi}_{D,L}$ is the hidden-state 
operator introduced in Definition~\ref{def:powers_hid_op}. 
\end{lemma}
\begin{proof}
	Fix $L \in \N$ and $D \geq 1$. Let $A^\ell, b^\ell$ be as in Lemma \ref{lem:hidpowers-approx} and note that $A^\ell \in \R^{\left(2^{\ell-1}+1\right) \times \rnnSizeFunctionHid{\hidPowers{D}{L}}}$ and  $b^\ell \in \R^{2^{\ell-1}+1}$, for $\ell \in \{1, \ldots, L\}$. Define the matrices and vectors
	\begin{align}
		\AoDel{1}                                  & \coloneqq
		\begin{pmatrix}
			0 & 1 \\
			1 & 0
		\end{pmatrix} A^1,
		                                           &
		\boDel{1}                                      & \coloneqq
		\begin{pmatrix}
			0 & 1 \\
			1 & 0
		\end{pmatrix} b^1,                                     \\
		\AoDel{\ell}                               & \coloneqq
		\begin{pmatrix}
			\eye{2^{\ell-1}} & 0
		\end{pmatrix} A^{\ell}, &
		\boDel{\ell}                               & \coloneqq
		\begin{pmatrix}
			\eye{2^{\ell-1}} & 0
		\end{pmatrix} b^{\ell}, \quad \textrm{ for } \ell \in\{2, \dots, L\},
	\end{align}
	and the output mapping
	\[
		\outPowers{D}{L}(h) \coloneqq A_o^{\pi } h + b_o^{\pi}, \textrm{ with }
		A_o^{\pi} \coloneqq \begin{pmatrix}
			\AoDel{1} \\ \AoDel{2}  \\ \vdots \\ \AoDel{L}
		\end{pmatrix}
		\textrm{ and }
		b_o^{\pi} \coloneqq \begin{pmatrix}
			\boDel{1} \\ \boDel{2}  \\ \vdots \\ \boDel{L}
		\end{pmatrix}.
	\]
	We now verify that $\outPowers{D}{L}$ satisfies the claimed properties. To this end, fix $x \in [-D, D]$ and $k \geq \ceil{\log(L)}+2$ arbitrarily and note that, by \eqref{eq:AK-representation} in the proof of Lemma \ref{lem:hidpowers-approx}, 
	\[
		A^{1} (\hidPowers{D}{L} \spreadInputOp x)[2^k-2] + b^{1} = \multiconcatRNNmap{1}{k-\timeOffsetIdx{1}}(x)
	\]
	and, by \eqref{eq:def-glk},
	$\multiconcatRNNmap{1}{k-\timeOffsetIdx{1}}(x) = (\rnnOpp^1_{D} \spreadInputOp x)[2^{k-\timeOffsetIdx{1}}-2]$.
	Combining the two identities yields
	\begin{align}
		((\outPowers{D}{L} \hidPowers{D}{L} \spreadInputOp x)[2^k-2])_1 & =
		\left(A^{1} (\hidPowers{D}{L} \spreadInputOp x)[2^k-2] + b^1\right)_2  =
		\left(\left(
		\rnnOpp^1_{D} \spreadInputOp x
		\right)[2^{k-\timeOffsetIdx{1}}-2]\right)_2 \overset{\textrm{Lem. }\ref{lem:square_and_identity}}{=} x.
	\end{align}
	For the second entry we have
	\begin{align}
		|((\outPowers{D}{L} \hidPowers{D}{L}  \spreadInputOp x)[2^k-2])_2 - x^{2^0+1}| & =
		\left|\left(A^{1} (\hidPowers{D}{L} \spreadInputOp x)[2^k-2] + b^{1}\right)_1 - \polyMapConcat{1}(x)_1\right| \overset{\textrm{Lem. \ref{lem:hidpowers-approx}}}{\leq}  \varepsilon_{1, k}.
	\end{align}
For $\ell \in \{2, \dots, L\}$ and $j\in\{1,\dots,2^{\ell-1}\}$, consider the 
$(2^{\ell-1}+j)$-th coordinate of 
$(\mathcal Q^{\pi}_{D,L}\mathcal K^{\pi}_{D,L}\mathcal D x)[2^{k}-2]$. 
Since $\AoDel{1}$ has two rows and, for $i=2,\dots,\ell-1$, each block 
$\AoDel{i}$ has $2^{i-1}$ rows, the total number of preceding rows is
\[
2 + \sum_{i=2}^{\ell-1}2^{i-1} = 2^{\ell-1}.
\]
Consequently, the $(2^{\ell-1}+j)$-th coordinate of the global output corresponds 
to the $j$-th coordinate of 
$A^{\ell}(\mathcal K^{\pi}_{D,L}\mathcal D x)[2^{k}-2]+b^{\ell}$. 
By Lemma~\ref{lem:hidpowers-approx}, we obtain
\[
\big|
(\mathcal Q^{\pi}_{D,L}\mathcal K^{\pi}_{D,L}\mathcal D x)[2^{k}-2]_{2^{\ell-1}+j}
- x^{2^{\ell-1}+j}
\big|
=
\big|
(A^{\ell}(\mathcal K^{\pi}_{D,L}\mathcal D x)[2^{k}-2]+b^{\ell})_{j}
- \polyMapConcat{\ell}(x)_{j}
\big|
\le \varepsilon_{\ell,k}.
\]
This completes the proof.
\end{proof}

We now consolidate the preceding results into a single statement describing the full RNN and its approximation properties.
\begin{theorem}\label{thm:final_powers}
	Let $L\in \N$ and $D \geq 1$. Then, there exists an RNN $\rnnPowers{D}{L} = \outPowers{D}{L} \hidPowers{D}{L}$ such that
	\begin{equation}\label{eq:final_powers_size_bound}
		\rnnSizeFunctionHid{\rnnPowers{D}{L}} \leq 40 \cdot 2^L, 
	\end{equation} and, for all $x\in[-D,D]$ and all $k \geq \ceil{\log{L}}+2$, 
	\begin{equation} \label{eq:final_powers_outputs1}
			((\rnnPowers{D}{L} \spreadInputOp x)[2^k-2])_1  = x,      
            \end{equation}
            and, for every $\ell \in\{1, \dots, L\}$ and $j\in\{1, \dots, 2^{\ell-1}\}$,
            \begin{equation}\label{eq:final_powers_outputs2}
            			| ((\rnnPowers{D}{L} x)[2^k-2])_{2^{\ell-1} + j} - x^{2^{\ell-1}+j}| \leq \varepsilon_{\ell, k}, 
	\end{equation}
	where $\varepsilon_{\ell, k} = 8 \cdot 2^\ell D^{2^{\ell}} 4^{-\frac{1}{2\ell}2^k}$.
\end{theorem}
\begin{proof}
	Fix $L\in \N$ and $D \geq 1$. All results in this section apply for these values. 
    Let $\hidPowers{D}{L}$ be as in Definition~\ref{def:powers_hid_op} and $\outPowers{D}{L}$ as in Lemma~\ref{lem:power_output_mapping}. Now, \eqref{eq:final_powers_size_bound} follows from Lemma \ref{lem:powers_size} and \eqref{eq:final_powers_outputs1} and \eqref{eq:final_powers_outputs2} are by Lemma \ref{lem:power_output_mapping}.
\end{proof}

\section{Approximation of polynomials}
\label{sec:approximation_of_polynomials}

It is now clear that any monomial can be approximated by an RNN according to Theorem \ref{thm:final_powers}. This construction, however, produces
meaningful outputs only at discrete time steps $t=2^k - 2$. The following modification yields an RNN
that generates a continuous sequence of outputs by holding the most recent valid approximation, rather than remaining idle between update times. Specifically, the output of the modified RNN remains constant between successive update times and is clipped to a prescribed range $[-B,B]$ to ensure boundedness; this clipping is operationally irrelevant, since $B>0$ may be chosen arbitrarily large.

\begin{lemma}\label{lem:outSmoothing}
	Fix $B\posConst{}$ and let $\rnnOpp=\outputOp\hidOp$ be an RNN with $\rnnSizeFunctionOut{\rnnOpp}=1$. Then there exists an RNN $\rnnOpp'$ such that, for all $x\in\R$,
	\begin{equation}
		(\rnnOpp' \spreadInputOp{} x)[2^k - 1 + \ell]  = {\cal C}((\rnnOpp \spreadInputOp{} x)[2^k - 2 ], -B, B), \quad \delforall k\geq 2, \, \ell\in\{0, \dots, 2^k-1\},
	\end{equation}
	with the clipping operator
	\begin{equation}
		{\cal C}(y, A, B) \coloneqq \begin{cases}
			A, & \textrm{if } y \leq A \\[2pt]
			B, & \textrm{if } y \geq B \\[2pt]
			y, & \textrm{otherwise.}
		\end{cases}
	\end{equation}
	Furthermore, the hidden state dimension satisfies 
	\begin{equation}
		\rnnSizeFunctionHid{\rnnOpp'} = \rnnSizeFunctionHid{\rnnOpp} + 11.
	\end{equation}
\end{lemma}
\begin{proof}
	Let the weights of $\rnnOpp$ be $\Ax{} \in \R^{\hiddenStateSize{} \times \inputSize{}}, \Ah{} \in \R^{\hiddenStateSize{} \times \hiddenStateSize{}}, \bh{} \in \R^{\hiddenStateSize{}}, \Ao{}\in\R^{1 \times \hiddenStateSize{}}$, and $\bo{} \in \R$. 
	The modified RNN $\rnnOp{}'$ has hidden state dimension $m+11$ and corresponding weights $A_x’\in\mathbb R^{(m+11)\times d}, A_h’\in\mathbb R^{(m+11)\times(m+11)}, b_h’\in\mathbb R^{m+11}, A_o’\in\mathbb R^{1\times(m+11)}, b_o’\in\mathbb R$, which are given by 
	\begin{align}
		A'_h & = \begin{pNiceArray}{c|c|c|c|c}
			        \Ah{} & 0 & 0 & 0  & 0\\
			        \hline
			        \Ao{}& 0  & 0  & 0& {B}\switchNetReadout{}\\
			        -\Ao{}& 0  & 0 &  0 & {B}\switchNetReadout{}\\
			        \hline
			        \Ao{} & 0 & 0 & 0 & 0\\
			        -\Ao{} & 0 & 0 & 0 & 0\\
			        \hline
			        0 &  \Imat{2} & -\Imat{2} & \Imat{2} & -B \Ivec{2} \switchNetReadout{}\\
			        \hline
			        0 &  0 & 0 &  0 &  \switchNetA{}\\
		        \end{pNiceArray},    \qquad
		b'_h = \begin{pNiceArray}{c}
			      \bh{} \\
			      \hline
			      \bo{}-B \\
			      -\bo{}-B \\
			      \hline
			      \bo{}-B\\
			      -\bo{}-B\\
			      \hline
			      0 \\
			      \hline
			      \switchNetB{}
		      \end{pNiceArray},
		\qquad
		A'_x = \begin{pNiceArray}{c}
			      \Ax{} \\
			      \hline
			      0 \\
			      0 \\
			      \hline
			      0 \\
			      0 \\
			      \hline
			      0 \\
			      \hline
			      0
		      \end{pNiceArray},
		\\
		A'_o & = \begin{pNiceArray}{c|cc|cc|cc|c}
			        0 &  1 & -1  & -1 & 1 & 1 & -1 & 0
		        \end{pNiceArray},
		\qquad
		b'_o = 0,
\end{align}
where $\switchNetReadout \coloneqq
		\begin{pmatrix}1 & 0 & 0 & 0& 0\end{pmatrix}$ 
	and $\switchNetA$ and $\switchNetB$ are as in \eqref{eq:switchnet_weights}. All zero blocks are of the dimensions required for blockwise compatibility.
	Fix an arbitrary input $x\in\R^{\inputSize{}}$ and consider the hidden state sequence $h'[\cdot]$ of $\rnnOp{}'$ generated by the corresponding input sequence $(\spreadInputOp x)[\cdot]$. We partition $h'[\cdot]$ according to the block structure of $A'_h$ as follows
	\begin{equation}
		h'[t] = \begin{pNiceArray}{c}h'_1[t] \\\hline  h'_2[t]  \\\hline  h'_3[t] \\\hline h'_4[t] \\\hline h'_5[t]
		\end{pNiceArray}  = \relu \left( A'_h h'[t-1] + A'_x( (Dx)[t] ) + b'_h  \right), \, t \in \No,
	\end{equation}
	and analyze the dynamics of each block individually. Observe that $h'_5[\cdot]$ evolves according to
	\begin{equation}
		h'_5[t] = \relu(\switchNetA h'_5[t-1] + \switchNetB{}), \, t \in \N_0, \text{ with } h'_5[-1] = 0.
	\end{equation}
	By Lemma \ref{lem:switch_net}, this implies
	\begin{equation} \label{teq:50}\switchNetReadout h'_5[t] = \switch[t+2], \quad \text{for all } t \in \N_0.
	\end{equation}
	The first block $h'_1[t]$ follows the recursion
	\begin{equation}
		h'_1[t] = \relu\left( \Ah{} h'_1[t-1] + \Ax{} ((\spreadInputOp{} x)[t]) + \bh{}\right), \, t \in \N_0, \quad \textrm{with } h'_1[-1] = 0,
	\end{equation}
	and hence coincides with the hidden state of the base network,
    $h'_1[t] = (\hidOp \spreadInputOp{} x)[t]$.
	Next, the second block $h'_2[t]$ evolves according to
	\begin{align}
		h'_2[t]
		 & = \relu
		\begin{pmatrix}
			\Ao{} h'_1[t-1] + {B}\switchNetReadout h'_5[t-1] + \bo{} - {B}{}  \\
			-\Ao{} h'_1[t-1] + {B}\switchNetReadout h'_5[t-1] - \bo{} - {B}{} \\
		\end{pmatrix}                          \\
		 & = \relu
		\begin{pmatrix}
			(\Ao{}  (\hidOp \spreadInputOp{} x)[t-1]+ \bo{}) - {B}(1 - \switchNetReadout h'_5[t-1])   \\
			-(\Ao{}  (\hidOp \spreadInputOp{} x)[t-1] + \bo{}) - {B}(1 - \switchNetReadout h'_5[t-1]) \\
		\end{pmatrix} \\
		& = \relu
		\begin{pmatrix}
			(\rnnOpp \spreadInputOp{} x)[t-1] - {B}(1 - \switch[t+1])    \\
			-(\rnnOpp   \spreadInputOp{} x)[t-1] - {B}(1 - \switch[t+1])\\
		\end{pmatrix}.
	\end{align}
	Similarly, we get
	\begin{align}
		h'_3[t]
		 & = \relu
		\begin{pmatrix}
			\Ao{} h'_1[t-1] + \bo{} - {B}{}   \\
			-\Ao{} h'_1[t-1]  - \bo{} - {B}{} \\
		\end{pmatrix}                                                                                                                                   \\
		 & = \relu
		\begin{pmatrix}
			(\rnnOpp \spreadInputOp{} x)[t-1] - {B}{}  \\
			-(\rnnOpp \spreadInputOp{} x)[t-1] - {B}{} \\
		\end{pmatrix}\label{teq:abc1}                                                                                                                         \\
		 & =
		\begin{pmatrix}
			\left((\rnnOpp \spreadInputOp{} x)[t-1] - {B}{}\right)\ind{(\rnnOpp \spreadInputOp{} x)[t-1] \, \geq \, B}   \\
			\left(-(\rnnOpp \spreadInputOp{} x)[t-1] - {B}{}\right)\ind{-(\rnnOpp \spreadInputOp{} x)[t-1] \, \geq \, B} \\
		\end{pmatrix}                   \\
		 & =
		\begin{pmatrix}
			\left(\relu\left((\rnnOpp \spreadInputOp{} x)[t-1]\right) - {B}{}\right)\ind{(\rnnOpp \spreadInputOp{} x)[t-1] \, \geq \, B}   \\
			\left(\relu\left(-(\rnnOpp \spreadInputOp{} x)[t-1]\right) - {B}{}\right)\ind{-(\rnnOpp \spreadInputOp{} x)[t-1] \, \geq \, B} \\
		\end{pmatrix}\label{teq:abc2}.\\
	\end{align}
	Next, we show that
	\begin{equation}\label{teq:bounding_term}
		 h'_2[t] -  h'_3[t] =  \begin{pmatrix}
			{\cal C}((\rnnOpp \spreadInputOp{} x)[t-1], 0, B)  \\
			{\cal C}(-(\rnnOpp \spreadInputOp{} x)[t-1], 0, B) \\
		\end{pmatrix} \switch{[t+1]}.
	\end{equation}
	This follows by a case distinction. If $\switch{[t+1]}=0$, then by \eqref{teq:abc1},
	\begin{align}
		 h'_2[t] -  h'_3[t] & =
		\relu\begin{pmatrix}
			     (\rnnOpp \spreadInputOp{} x)[t-1] - {B}  \\
			     -(\rnnOpp \spreadInputOp{} x)[t-1] - {B} \\
		     \end{pmatrix} -
		\relu\begin{pmatrix}
			     (\rnnOpp \spreadInputOp{} x)[t-1] - {B}{}  \\
			     -(\rnnOpp \spreadInputOp{} x)[t-1] - {B}{} \\
		     \end{pmatrix} = 0.
	\end{align}
	For $\switch{[t+1]}=1$, we obtain from \eqref{teq:abc2}, 
	\begin{align}
		 h'_2[t] - h'_3[t] & =
		\relu\begin{pmatrix}
			     (\rnnOpp \spreadInputOp{} x)[t-1]  \\
			     -(\rnnOpp \spreadInputOp{} x)[t-1] \\
		     \end{pmatrix} -
		\begin{pmatrix}
			\left(\relu\left((\rnnOpp \spreadInputOp{} x)[t-1]\right) - {B}{}\right)\ind{(\rnnOpp \spreadInputOp{} x)[t-1] \, \geq \, B}   \\
			\left(\relu\left(-(\rnnOpp \spreadInputOp{} x)[t-1]\right) - {B}{}\right)\ind{-(\rnnOpp \spreadInputOp{} x)[t-1] \, \geq \, B} \\
		\end{pmatrix}
		\\
		                                  & =
		\begin{pmatrix}
			{\cal C}((\rnnOpp \spreadInputOp{} x)[t-1], 0, B)  \\
			{\cal C}(-(\rnnOpp \spreadInputOp{} x)[t-1], 0, B) \\
		\end{pmatrix},
	\end{align}
	thereby establishing \eqref{teq:bounding_term}.

	Hence, $h'_4[t], t \in \N_0$, satisfies the recurrence 
	\begin{align}
		h'_4[t]
		 & = \relu\left(
		h'_2[t-1] - h'_3[t-1] + h'_4[t-1] - B \Ivec{2}\switchNetReadout h'_5[t-1]
		\right)          \\
		 \overset{\eqref{teq:bounding_term}, \eqref{teq:50}}&{=} \relu\left(
		\begin{pmatrix}
			{\cal C}((\rnnOpp \spreadInputOp{} x)[t-2], 0, B)  \\
			{\cal C}(-(\rnnOpp \spreadInputOp{} x)[t-2], 0, B) \\
		\end{pmatrix}\switch[t] +  h'_4[t-1] - B \Ivec{2}\switch[t+1]
		\right), \text{with } h'_4[-1]=0. \label{eq:recurrenceh4}
	\end{align}
	Note that $h'_4[t] = 0$, for $t\in\{0,1,2, 3\}$.
	We now prove by nested induction that
	\begin{align}
		h'_4[2^k-1]      & = 0,             & \text{for } k                                                   & \in\N,\; k\geq 2, \quad \label{teq:outSmoothing_induction_3a} \\
		\text{and}\quad
		h'_4[2^k + \ell] & =
		\begin{pmatrix}
			{\cal C}((\rnnOpp \spreadInputOp{} x)[2^k-2], 0, B)  \\
			{\cal C}(-(\rnnOpp \spreadInputOp{} x)[2^k-2], 0, B) \\
		\end{pmatrix},
		                & \text{for } \ell & \in \{0, \dots, 2^k - 2\}.\label{teq:outSmoothing_induction_3b}
	\end{align}
	We have already concluded that \eqref{teq:outSmoothing_induction_3a} holds for $k=2$. Assume now that \eqref{teq:outSmoothing_induction_3a} is valid for some $k\geq 2$, and compute
	\begin{align}
		h'_4[2^k ]
		 & = \relu\left(
		\begin{pmatrix}
			{\cal C}((\rnnOpp \spreadInputOp{} x)[2^k-2], 0, B)  \\
			{\cal C}(-(\rnnOpp \spreadInputOp{} x)[2^k-2], 0, B) \\
		\end{pmatrix}\switch[2^k] + h'_4[2^k-1] - B \Ivec{2}\switch[2^k+1]
		\right)                                                \\
		 & = \begin{pmatrix}
			     {\cal C}((\rnnOpp \spreadInputOp{} x)[2^k-2], 0, B)  \\
			     {\cal C}(-(\rnnOpp \spreadInputOp{} x)[2^k-2], 0, B) \\
		     \end{pmatrix}.
	\end{align}
	This verifies \eqref{teq:outSmoothing_induction_3b} for $\ell=0$.
	Next, assume that \eqref{teq:outSmoothing_induction_3b} holds for some $\ell\in\{0, \dots, 2^k-3\}$, and compute
	\begin{align}
		h'_4[2^k + \ell +1 ]
		 & = \relu\left(
		\begin{pmatrix}
			{\cal C}((\rnnOpp \spreadInputOp{} x)[2^k+\ell-1], 0, B)  \\
			{\cal C}(-(\rnnOpp \spreadInputOp{} x)[2^k+\ell-1], 0, B) \\
		\end{pmatrix} \switch[2^k+\ell+1] + h'_4[2^k+\ell] - B \Ivec{2}\switch[2^k+\ell+2]
		\right)                                          \\
		 & = \relu( h'_4[2^k+\ell])
		= \begin{pmatrix}
			  {\cal C}((\rnnOpp \spreadInputOp{} x)[2^k-2], 0, B)  \\
			  {\cal C}(-(\rnnOpp \spreadInputOp{} x)[2^k-2], 0, B) \\
		  \end{pmatrix}.
	\end{align}
	This completes the induction over $\ell$ and establishes \eqref{teq:outSmoothing_induction_3b}. To complete the induction step for
    \eqref{teq:outSmoothing_induction_3a}, we use \eqref{eq:recurrenceh4} to get
	\begin{align}
		h'_4[2^{k+1} -1]
		 & =
		\relu\left(
		\begin{pmatrix}
			{\cal C}((\rnnOpp \spreadInputOp{} x)[2^{k+1}-3], 0, B)  \\
			{\cal C}(-(\rnnOpp \spreadInputOp{} x)[2^{k+1}-3], 0, B) \\
		\end{pmatrix} \switch[2^{k+1} -1 ] + h'_4[2^{k+1} -2] - B \Ivec{2}\switch[2^{k+1} ]
		\right)                                                           \\
		 & =
		\relu\left(
		h'_4[2^{k}  + 2^k -2] - B \Ivec{2}
		\right)                                                           \\
		 & \overset{\eqref{teq:outSmoothing_induction_3b}}{=} \relu\left(
		\begin{pmatrix}
			{\cal C}((\rnnOpp \spreadInputOp{} x)[2^k-2], 0, B)  \\
			{\cal C}(-(\rnnOpp \spreadInputOp{} x)[2^k-2], 0, B) \\
		\end{pmatrix}
		- B \Ivec{2}
		\right) = 0.
	\end{align}
	This establishes \eqref{teq:outSmoothing_induction_3a} for $k+1$ and thereby completes the overall nested induction. The network output is given by
	\begin{align}
		(\rnnOpp' \spreadInputOp x)[t] & =
		\begin{pmatrix} 1 & -1 \end{pmatrix} h'_2[t] + \begin{pmatrix} -1 & 1 \end{pmatrix} h'_3[t]+
		\begin{pmatrix} 1 & -1 \end{pmatrix}h'_4[t]                              \\
		                               & =
		\begin{pmatrix} 1 & -1 \end{pmatrix}\left(h'_2[t] - h'_3[t]+
		h'_4[t]\right)                                                  \\
		                               & \overset{\eqref{teq:bounding_term}}{=}
		\begin{pmatrix} 1 & -1 \end{pmatrix} \left( \begin{pmatrix}
			                                            {\cal C}((\rnnOpp \spreadInputOp{} x)[t-1], 0, B)  \\
			                                            {\cal C}(-(\rnnOpp \spreadInputOp{} x)[t-1], 0, B) \\
		                                            \end{pmatrix} \switch{[t+1]} + h'_4[t]\right).
	\end{align}
	By \eqref{teq:outSmoothing_induction_3a}, \eqref{teq:outSmoothing_induction_3b}, and the fact that $\switch[t+1] = 1$, for $t=2^k-1$ with $k\geq2$, and $\switch[t+1]=0$ otherwise, we find
	\begin{align}
		(\rnnOpp' \spreadInputOp x)[2^k-1+\ell] & =
		\begin{pmatrix} 1 & -1 \end{pmatrix} \begin{pmatrix}
			                                     {\cal C}((\rnnOpp \spreadInputOp{} x)[2^k-2], 0, B)  \\
			                                     {\cal C}(-(\rnnOpp \spreadInputOp{} x)[2^k-2], 0, B) \\
		                                     \end{pmatrix} \\
		                                        & =
		{\cal C}((\rnnOpp \spreadInputOp{} x)[2^k-2], -B, B),
		\qquad \delforall \ell \in \{0, \dots, 2^k-1\},
	\end{align}
	as claimed.
\end{proof}
To describe the RNN behavior uniformly across all time indices,
we next introduce a convenient re-indexing of the dyadic intervals arising in Lemma~\ref{lem:outSmoothing}.
\begin{lemma}\label{lem:write_t_as_power}
	For $t\in\N$, let $\widetilde{k}(t) \coloneqq \floor{\log(t+1)}$ and $\widetilde{\ell}(t)\coloneqq t+1 - 2^{\widetilde{k}(t)}$. Then, 
	\[
		t = 2^{\widetilde{k}(t)} -1 + \widetilde{\ell}(t) \quad \textrm{and} \quad \widetilde{\ell}(t) \in \{0, \dots, 2^{\widetilde{k}(t)}-1\}.
	\]
\end{lemma}
\begin{proof}
	Arbitrarily fix $t\in\N$ and let $\widetilde{k} \coloneqq \widetilde{k}(t)$ and $\widetilde{\ell} \coloneqq \widetilde{\ell}(t)$.
	Then,
	\[
		2^{\widetilde{k}} -1 + \widetilde{\ell} = 2^{\widetilde{k}} -1 + t+1 - 2^{\widetilde{k}} = t.
	\]
	Moreover, we have 
	\[
		\widetilde{\ell} = t+1 - 2^{\floor{\log(t+1)}} \geq t+1 - 2^{\log(t+1)} = t+1 - t-1 = 0.
	\]
	Furthermore, 
	\begin{align}
		\widetilde{\ell} & = t+1 - 2^{\floor{\log(t+1)}}                          \\
		                 & = 2\cdot 2^{\log(t+1)-1} - 2^{\floor{\log(t+1)}}       \\
		                 & < 2\cdot 2^{\floor{\log(t+1)}} - 2^{\floor{\log(t+1)}} \\
		                 & = 2^{\floor{\log(t+1)}} = 2^{\widetilde{k}},
	\end{align}
so $\tilde{\ell} \in \{0,...,2^{\tilde{k}}-1\}$.
\end{proof}

We now combine the constructions from the previous sections to approximate arbitrary polynomials by a single RNN. The following theorem summarizes the result.
\begin{theorem}\label{tm:main_theorem_with_proof}
	Let $N\in\N$, $D \geq 1$, and $a_0, \dots, a_N \in \R$. There exists an RNN $\rnnPolynom$ such that, for all $x\in [-D, D]$ and all $t \geq 16 \log(N)$, 
	\begin{equation}
		\left|(\rnnPolynom \spreadInputOp x)[t] - \sum_{i=0}^{N} a_i x^i\right| \leq
		\norm{a}_1 \constMultiplicative 4^{- \constInPower t},
	\end{equation}
	with $\constMultiplicative = 16N D^{2N},\constInPower = \frac{1}{4\ceil{\log(N)}}$, and $\|a\|_{1}=\sum_{i=0}^{N}|a_i|$. Furthermore, 
    $\rnnSizeFunctionHid{\rnnPolynom} \leq 80N + 11$.
\end{theorem}
\begin{proof}
	We may assume without loss of generality that $N\geq 2$. Indeed, if $N \leq 1$, then by Lemma~\ref{lm:identity_net} the identity RNN $\rnnId$ satisfies
    $(\rnnId \spreadInputOp x)[t]  = x$, for all $t\in \N_0$. Applying Lemma~\ref{lm:linear_map_rnn_output} with the affine readout $h \mapsto a_1 h + a_0$ (with $a_1=0$ when $N=0$) yields an RNN that realizes the polynomial $a_1 x +a_0$ exactly.
	Let $L \coloneqq \ceil{\log(N)}\in\N$ and extend the coefficient sequence by setting $a_{N+1}= \dots = a_{2^L}=0$.
	We now modify the RNN $\rnnPowers{D}{L}$ from Theorem \ref{thm:final_powers} using Lemma \ref{lm:linear_map_rnn_output} with $A= \begin{pmatrix}
			a_1, \dots, a_{2^L}
		\end{pmatrix}$ and $b = a_0$ to obtain the RNN $\widetilde{\rnnPolynom}$ realizing
	\[
		(\widetilde{\rnnPolynom} \spreadInputOp x)[t] =  a_0 + \sum_{i=1}^{2^L} a_i \left((\rnnPowers{D}{L} \spreadInputOp x)[t]\right)_i.
	\]
	We bound, for $k\geq \ceil{\log(L)}+2$,
	\begin{align}
		\begin{split}\label{teq:34}
			\left|(\widetilde{\rnnPolynom} \spreadInputOp x)[2^k-2] -  \sum_{i=0}^{N} a_i x^i \right|
			 & \leq \sum_{i=2}^{2^L} |a_i| \left|\left( (\rnnPowers{D}{L} \spreadInputOp x)[2^k-2]\right)_i - x^i  \right|                                                               \\
			 & = \sum_{\ell=1}^{L} \sum_{j=1}^{2^{\ell-1}} |a_{2^{\ell-1} + j}| \left| \left((\rnnPowers{D}{L} \spreadInputOp x)[2^k-2]\right)_{2^{\ell-1}+j} - x^{2^{\ell-1}+j} \right| \\
			 \overset{\textrm{Thm.~\ref{thm:final_powers}}}&{\leq}  \sum_{\ell=1}^{L}  \varepsilon_{\ell, k} \sum_{j=1}^{2^{\ell-1}} |a_{2^{\ell-1} + j}|                          \\
			 & \leq   \varepsilon_{L, k} \sum_{\ell=1}^{L}  \sum_{j=1}^{2^{\ell-1}} |a_{2^{\ell-1} + j}|                                                                           \\
			 & = \norm{a}_1  8 \cdot  2^{L}  D^{2^L}4^{- \frac{1}{2L}2^{k}},
		\end{split}
	\end{align}
    where $\varepsilon_{\ell,k}$ was defined in Lemma~\ref{lem:hidpowers-approx}.
	Next, set $B = \max_{x\in[-D, D]} \left| \sum_{i=0}^{N} a_i x^i\right|$ and apply Lemma \ref{lem:outSmoothing} with this $B$ and $\rnnOpp=\widetilde{\rnnPolynom}$ to obtain an RNN $\rnnPolynom$ satisfying, for $k\geq 2$,
	\begin{equation}
		\label{teq:35}
		(\rnnPolynom \spreadInputOp x)[2^k-1 + \ell] = {\cal C}((\widetilde{\rnnPolynom} \spreadInputOp x)[2^k-2], -B, B),\qquad \delforall \ell \in \{0, \dots, 2^k-1\}.
	\end{equation}
	Combining this with Lemma~\ref{lem:write_t_as_power}, we compute 
	\begin{align}
		\left|(\rnnPolynom \spreadInputOp x)[t] -  \sum_{i=0}^{N} a_i x^i \right|
		& = \left| (\rnnPolynom \spreadInputOp x)[2^{\widetilde{k}(t)} - 1 + \widetilde{\ell}(t)]  -  \sum_{i=0}^{N} a_i x^i \right| \label{teq:write_t_as_powers} \\
		 \overset{\eqref{teq:35}}
         & = 
		\left| {\cal C}((\widetilde{\rnnPolynom} \spreadInputOp x)[2^{\widetilde{k}(t)} - 2], -B, B)  -  \sum_{i=0}^{N} a_i x^i \right| \label{teq:37} \\ 
         & \leq
		\left| (\widetilde{\rnnPolynom} \spreadInputOp x)[2^{\widetilde{k}(t)} - 2]  -  \sum_{i=0}^{N} a_i x^i \right|.                                     
        \end{align}
Since $ t \geq 16 \log(N)$, we have 
\begin{align}
		\widetilde{k}(t) & = \floor{\log(t+1)} \geq \floor{\log( 16 \log(N))} = \floor{\log(2\log(N))} + 3                      \\
		                 & \geq  \ceil{\log(\log(N)+ \log(N))} + 2 \geq  \ceil{\log(\ceil{\log(N)})} + 2 =   \ceil{\log(L)} + 2,
	\end{align}
and therefore \eqref{teq:34} applies. Hence, 
\begin{equation}
		\left|(\rnnPolynom \spreadInputOp x)[t] -  \sum_{i=0}^{N} a_i x^i \right|  \leq 
	\norm{a}_1  8 \cdot  2^{L}  D^{2^L}4^{- \frac{1}{2L}2^{\widetilde{k}(t)}}. \label{teq:38}                                                       \\
	\end{equation}
	We further upper bound the RHS of \eqref{teq:38} according to 
	\begin{align}
		& \norm{a}_1  8 \cdot  2^{L}  D^{2^L}4^{- \frac{1}{2L}2^{\widetilde{k}(t)}}    \\
		 & = \norm{a}_1  8 \cdot  2^{\ceil{\log(N)}}
		D^{2^{\ceil{ \log(N)}}}4^{- \frac{1}{2\ceil{ \log(N)}}2^{\widetilde{k}(t)}}            \\
		 & \leq \norm{a}_1  8 \cdot  2N
		D^{2N}4^{- \frac{1}{2\ceil{ \log(N)}}2^{\widetilde{k}(t)}}                            \\
		 & = \norm{a}_1  16 N     D^{2N}4^{- \frac{1}{2\ceil{ \log(N)}}2^{\floor{\log(t+1)}}} \\
		 & \leq \norm{a}_1  16 N     D^{2N}4^{- \frac{1}{2\ceil{ \log(N)}}2^{\log(t+1)-1}}    \\
		 & \leq \norm{a}_1  16 N     D^{2N}4^{- \frac{t}{4\ceil{ \log(N)}}}.
	\end{align}
	The proof is finalized by noting that 
	\begin{equation}
		\rnnSizeFunctionHid{\rnnPolynom} \overset{\textrm{Lem. \ref{lem:outSmoothing}}}{=} \rnnSizeFunctionHid{\widetilde{\rnnPolynom}}+11
		= \rnnSizeFunctionHid{\rnnPowers{D}{L}} + 11
		\overset{\textrm{Lem. \ref{lem:powers_size}}}{\leq} 40 \cdot 2^L + 11 \leq 80 N + 11. 
	\end{equation}
\end{proof}

\section{Conclusion and Outlook}

This paper introduced a new recurrent neural network approximation paradigm based on a 
reversal of the classical quantifier structure. Specifically, for a fixed target function, a single 
RNN with fixed topology and fixed weights achieves any prescribed error tolerance by 
running for sufficiently many time steps. Applied to univariate 
polynomials, this approach yields RNNs whose size is independent of the 
error tolerance and whose hidden-state dimension grows linearly with the polynomial 
degree.

Several directions for further work follow from this new approximation paradigm. An
extension to multivariate polynomials is a natural next step and appears readily
achievable. Extensions to more general smooth or continuous functions appear to be more involved,
yet the methods developed here provide a reasonable starting point. It would also be interesting to examine 
whether the new paradigm can lead to metric-entropy optimality in the approximation of function classes.
Finally, the analysis of RNNs with quantized or
reduced-precision weights constitutes another interesting research direction. 

\section*{Acknowledgments}

The authors gratefully acknowledge the assistance of OpenAI’s ChatGPT, which was used 
for language polishing, consistency verification, and iterative refinement of the 
manuscript’s exposition. All mathematical developments and results are solely the work 
of the authors.

H.~Bölcskei and V.~Abadie gratefully acknowledge support by the Lagrange Mathematics and Computing Research Center, Paris, France.


%% file: src/polyrnn/polyrnn_notation.tex
\DeclarePairedDelimiter{\ceil}{\lceil}{\rceil}
\DeclarePairedDelimiter{\floor}{\lfloor}{\rfloor}
\DeclarePairedDelimiter{\norm}{\lVert}{\rVert}

\renewcommand{\O}{\mathcal{O}}
\newcommand{\Z}{\mathbb{Z}}
\newcommand{\R}{\mathbb{R}}
\newcommand{\N}{\mathbb{N}}
\newcommand{\No}{\mathbb{N}_0} 
\newcommand{\C}{\mathbb{C}}
\newcommand{\E}{\mathbb{E}}
\newcommand{\ZZ}[0]{\Z} 

\newcommand{\shiftOp}[1]{\mathbf{T}_{#1}}
\newcommand{\timeRestrictOp}[2]{R_{#1}^{#2}}

\newcommand{\seqBasisSet}[0]{\mathcal{G}}
\newcommand{\sigSpace}[0]{S}
\newcommand{\projLeftSide}[0]{\mathcal{P}}

\newcommand{\xhat}[0]{\widehat{x}} 
\newcommand{\xtilde}[0]{\widetilde{x}}
\newcommand{\dt}[1]{\delta_{#1}}
\newcommand{\cover}[0]{\mathcal{C}}

\newcommand{\maxNet}[0]{\Phi_\wedge}
\newcommand{\minNet}[0]{\Phi_\vee}

\newcommand{\yhat}[0]{\widehat{y}}

\newcommand{\bound}[0]{D} 

\newcommand{\normIII}[1]{{\left\vert\kern-0.25ex\left\vert\kern-0.25ex\left\vert #1 
    \right\vert\kern-0.25ex\right\vert\kern-0.25ex\right\vert}}

\renewcommand{\vec}[1]{\mathbf{#1}}

\newcommand{\h}[1]{\vec{h}[{#1}]}
\newcommand{\ind}[1]{\mathbbm{1}_{\{#1\}}}
\newcommand{\eye}[1]{\mathbbm{I}_{#1}}
\newcommand{\iZ}[1]{\mathcal{Z}^{-1}\left\{ #1\right\}}
\newcommand{\Imat}[1]{\mathbb{I}_{#1}}
\newcommand{\Omat}[0]{\mathbb{O}}
\newcommand{\Ovec}[1]{\vec{0}_{#1}}
\newcommand{\Ivec}[1]{\vec{1}_{#1}} 

\newcommand{\quantF}[1]{S_{\delta} (#1)}
\newcommand{\quantSet}[0]{\mathbb{S}_{\delta}}

\newcommand{\mnorm}[1]{\norm{#1}_{max}}
\newcommand{\inorm}[1]{\norm*{#1}_{\infty}}

\newcommand{\sysOp}[0]{\mathcal{L}}
\newcommand{\seqSpace}[1]{(\R^{#1})^{\No}} 

\newcommand{\rnnOp}[1]{\mathcal{R}_{#1}}
\newcommand{\rnnOpp}{\mathcal{R}} 
\newcommand{\quantRnnOp}[1]{\widetilde{\rnnOp{#1}}}
\newcommand{\quantRnnImp}[0]{\widetilde{k}}

\newcommand{\mEnt}{\mathcal{E}}
\newcommand{\entSet}{\mathcal{C}}
\newcommand{\sysSubset}{\entSet_{C, a}}
\newcommand{\opMetric}{\rho_{*}}
\newcommand{\hardyRho}{\rho_{H}}

\newcommand{\impResp}{k}
\newcommand{\transF}{K}
\newcommand{\relu}[0]{\rho}

\newcommand{\metric}[0]{\rho}

\newcommand{\stateDim}[0]{m} 
\newcommand{\hidLayerDim}[0]{n}
\newcommand{\rnnOut}[0]{y} 

\newcommand{\nonLinSpace}[0]{\mathcal{G}}
\newcommand{\packing}[0]{M}
\newcommand{\covering}[0]{N}

\newcommand{\sigBall}[0]{\mathcal{S}}

\newcommand{\approxFunctional}{g_{\alpha_1, \dots, \alpha_{M-1}}}

\DeclarePairedDelimiter{\ceilManual}{\lceil}{\rceil}
\DeclarePairedDelimiter{\floorManual}{\lfloor}{\rfloor}

\newcommand{\basisVec}[1]{\mathbf{1}_{#1}}

\newcommand{\ghat}[0]{\hat{g}}

\let\mod\relax

\newcommand{\outputOp}[0]{\mathcal{Q}}
\newcommand{\hidOp}[0]{\mathcal{K}}
\newcommand{\linOp}[0]{\mathcal{L}}

\newcommand{\inputDim}{d}
\newcommand{\outputDim}{d'}
\newcommand{\ho}{h^o}

\newcommand{\actFinterval}{\mathcal{D}}

\newcommand{\stretchOp}{\mathcal{Q}}
\newcommand{\downsampleOp}{\mathcal{Q}^{-1}}

\newcommand{\hidAct}{c}
\newcommand{\inpSeq}{\widetilde{x}} 

\newcommand{\otherHid}{\widetilde{h}}
\newcommand{\otherX}{\widetilde{x}}
\newcommand{\otherY}{\widetilde{y}}

\newcommand{\ytilde}{\widetilde{y}}
\newcommand{\ztilde}{\widetilde{z}}

\newcommand{\spreadInputOp}{\mathcal{D}}

\newcommand{\unrolledRnn}[1]{\Phi_{#1}}
\newcommand{\expansionOp}{\mathrm{E}_{\tiny\beta}}

\newcommand{\timeSwitch}{\tau} 

\newcommand{\switchNetA}{\widehat{A}}
\newcommand{\switchNetB}{\widehat{b}}

\newcommand{\nn}{\mathcal{N}}
\newcommand{\rnnIntro}{\mathcal{R}}

\newcommand{\switch}{\widehat{\delta}} 

\newcommand{\rnnPart}[1]{\rnnOp{}^{#1}}
\newcommand{\rnnf}{\rnnPart{f}} 
\newcommand{\rnng}{\rnnPart{g}}
\newcommand{\WeightMat}[2]{A_{#1}^{#2}} 
\newcommand{\biasVec}[2]{b_{#1}^{#2}} 
\newcommand{\Ax}[1]{\WeightMat{x}{#1}}
\newcommand{\Ao}[1]{\WeightMat{o}{#1}}
\newcommand{\Ah}[1]{\WeightMat{h}{#1}}
\newcommand{\bx}[1]{\biasVec{x}{#1}}
\newcommand{\bo}[1]{\biasVec{o}{#1}}
\newcommand{\bh}[1]{\biasVec{h}{#1}}

\newcommand{\hidBound}{D_h}
\newcommand{\outputBound}{D_o}
\newcommand{\switchNetReadout}{\widehat{A}_o}
\newcommand{\boundOutF}{D_1} 
\newcommand{\boundHidG}{\hidBound} 
\newcommand{\boundHidF}{\hidBound} 
\newcommand{\boundOutG}{D_2} 
\newcommand{\inputBound}{D}
\newcommand{\subMatOut}{W}
\newcommand{\subMatIn}{M}
\newcommand{\subMatOutMap}{\mathring{\mathcal{\subMatOut}}}
\newcommand{\subMatInMap}{\mathring{\mathcal{\subMatIn}}}

\newcommand{\rnnSizeFunctionIn}[1]{\mathcal{M}_{\mathrm{in}}(#1)}
\newcommand{\rnnSizeFunctionOut}[1]{\mathcal{M}_{\mathrm{out}}(#1)}
\newcommand{\rnnSizeFunctionHid}[1]{\mathcal{M}_{\mathrm{hid}}(#1)}

\newcommand{\hiddenStateSize}[1]{m_{#1}}
\newcommand{\inputSize}[1]{d_{#1}}
\newcommand{\outputSize}[1]{d'_{#1}}

\newcommand{\rnnConcatMap}[2]{\mathring{\Psi}_{{#1}}{({#2})}} 

\newcommand{\rnnMultiConcat}[1]{\widehat{\rnnOpp}^{#1}}
\newcommand{\hidMultiConcat}[1]{\widehat{\hidOp}^{#1}}
\newcommand{\outMultiConcat}[1]{\widehat{\outputOp}^{#1}}
\newcommand{\multiTmp}[1]{y_{#1}}

\newcommand{\rnnId}{\rnnOpp^{\textrm{Id}}}
\newcommand{\hidId}{\hidOp^{\textrm{Id}}}
\newcommand{\outId}{\outputOp^{\textrm{Id}}}
\newcommand{\rnnSquare}{\rnnOpp^{\textrm{Sq}}}
\newcommand{\outSquare}{\outputOp^{\textrm{Sq}}}
\newcommand{\hidSquare}{\hidOp^{\textrm{Sq}}}
\newcommand{\rnnZero}{\rnnOpp^{\textrm{o}}}
\newcommand{\outZero}{\outputOp^{\textrm{o}}}
\newcommand{\hidZero}{\hidOp^{\textrm{o}}}
\newcommand{\rnnMult}{\rnnOpp^{\times}}
\newcommand{\hidMult}{\hidOp^{\times}}
\newcommand{\outMult}{\outputOp^{\times}}
\newcommand{\rnnPowers}[2]{\rnnOpp^{\pi}_{#1,#2}}
\newcommand{\hidPowers}[2]{\hidOp^{\pi}_{#1,#2}}
\newcommand{\outPowers}[2]{\outputOp^{\pi}_{#1,#2}}
\newcommand{\rnnPolynom}{\rnnOpp^{\pi}_a}
\newcommand{\hidPolynom}{\hidOp^{\pi}_a}
\newcommand{\outPolynom}{\outputOp^{\pi}_a}

\newcommand{\rnnPoly}[2]{\rnnOpp^{#1}_{#2}}
\newcommand{\hidPoly}[2]{\hidOp^{#1}_{#2}}
\newcommand{\outPoly}[2]{\outputOp^{#1}_{#2}}
\newcommand{\polyMap}[1]{f^{#1}}
\newcommand{\polyMapConcat}[1]{F^{#1}}
\newcommand{\rnnTmp}{\widetilde{\rnnOpp}}
\newcommand{\rnnPolyMap}[1]{g^{#1}}
\newcommand{\rnnPolyMapConcat}[1]{G^{#1}}

\newcommand{\rnnPolyConcat}{\widehat{\rnnOpp}}
\newcommand{\hidPolyConcat}{\widehat{\hidOp}}

\newcommand{\approxError}[1]{\epsilon_{#1}}

\newcommand{\AoDel}[1]{\widetilde{\Ao{#1}}}
\newcommand{\boDel}[1]{\widetilde{\bo{#1}}}

\newcommand{\rnnMultiConcatTree}{\widehat{\mathcal{R}}}
\newcommand{\hidOpMultiConcatTree}{\widehat{\hidOp{}}}
\newcommand{\outOpMultiConcatTree}{\widehat{\outputOp{}}}

\newcommand{\multiconcatRNNmap}[2]{g_{#1}^{#2}} 
\newcommand{\multiconcatRNNmapG}[2]{G_{#1}^{#2}} 
\newcommand{\timeOffsetIdx}[1]{\widetilde{k}_{#1}}

\newcommand{\constMultiplicative}{C_1}
\newcommand{\constInPower}{C_2}
\newcommand{\clip}{\mathrm{clip}} 

\newcommand{\vecleq}{{\preccurlyeq}}

\newcommand{\squareFunc}{\textrm{Sq}}
\newcommand{\posConst}{> 0} 

\newcommand{\iter}[1]{{\color{blue}#1}}
\newcommand{\iterDel}[1]{{\color{red}#1}}
\newcommand{\iterComment}[1]{{\color{green} --- H: #1 --- }}

\newcommand{\delforall}{} 